%% file: main.tex
\pdfoutput=1

\documentclass[11pt]{article}

\usepackage{emnlp2023}

\usepackage{times}
\usepackage{latexsym}

\usepackage[T1]{fontenc}

\usepackage[utf8]{inputenc}

\usepackage{microtype}

\usepackage{booktabs} 
\usepackage{fancyhdr}  

\newcommand{\methodcomp}{MosaiCLIP}
\newcommand{\methodcompbold}{\textbf{MosaiCLIP}}
\newcommand{\methodcompshort}{MC}
\newcommand{\methodcompNoCurric}{MosaiCLIP\textsubscript{NoCurric}}
\newcommand{\methodcompNoCurricbold}{\textbf{MosaiCLIP\textsubscript{NoCurric}}}
\newcommand{\methodcompwiseft}{MosaiCLIP\textsubscript{WiSE-FT}}
\newcommand{\methodcompwiseftbold}{\textbf{MosaiCLIP\textsubscript{WiSE-FT}}}

\newcommand{\clip}{CLIP}
\newcommand{\negclip}{NegCLIP}

\usepackage{graphicx}
\usepackage{array}
\usepackage{multirow}
\usepackage{amsmath}
\usepackage{amssymb}
\usepackage{makecell}
\usepackage{subcaption}
\usepackage{placeins}
\usepackage{xcolor}
\usepackage{colortbl}
\usepackage{float}
\usepackage{makecell}
\usepackage{tabularx}
\usepackage{enumitem}
\usepackage[normalem]{ulem}

    \makeatletter
\def\@fnsymbol#1{\ensuremath{\ifcase#1\or \dagger\or *\or
   \mathsection\or \mathparagraph\or \|\or **\or \dagger\dagger
   \or \ddagger\ddagger \else\@ctrerr\fi}}
    \makeatother

\title{Coarse-to-Fine Contrastive Learning in Image-Text-Graph Space for Improved Vision-Language Compositionality}

\author{
    Harman Singh$^{1}$\thanks{\hspace{0.3em} Work done while at Meta.}\hspace{0.3em},  
    \textbf{Pengchuan Zhang}$^{1}$,
    \textbf{Qifan Wang}$^{1}$\textbf{,} \\
    \textbf{Mengjiao Wang}$^{1}$\textbf{,}
    \textbf{Wenhan Xiong}$^{1}$\textbf{,}
    \textbf{Jingfei Du}$^{1\dagger}$\textbf{,}
    \textbf{Yu Chen}$^{2\dagger}$
    \\
    $^{1}$Meta AI \quad
    $^{2}$Anytime.AI\\
    \texttt{harmansingh.iitd@gmail.com} \\
    \texttt{\{pengchuanzhang, wqfcr, mengjiaow, xwhan, jingfeidu\}@meta.com} \\
    \texttt{ychen@anytime-ai.com}
}

\begin{document}
\maketitle

\newcommand{\Lcal}{\mathcal{L}}
\newcommand{\Bcal}{\mathcal{B}}
\newcommand{\Pcal}{\mathcal{P}}
\newcommand{\Ucal}{\mathcal{U}}
\newcommand{\Vcal}{\mathcal{V}}
\newcommand{\Ecal}{\mathcal{E}}
\newcommand{\Dcal}{\mathcal{D}}
\newcommand{\Ical}{\mathcal{I}}
\newcommand{\Tcal}{\mathcal{T}}
\newcommand{\Xcal}{\mathcal{X}}
\newcommand{\Gcal}{\mathcal{G}}
\newcommand{\uv}{\boldsymbol{u}}
\newcommand{\vv}{\boldsymbol{v}}
\newcommand{\xv}{\boldsymbol{x}}
\newcommand{\tv}{\boldsymbol{t}}
\newcommand{\gv}{\boldsymbol{g}}

\input{./sections/abstract}
\input{./sections/intro}
\input{./sections/related}
\input{./sections/model}
\input{./sections/expt}
\input{./sections/conclusion}
\input{./sections/limitations}
\input{./sections/ack}

\FloatBarrier
\bibliography{references_custom}
\bibliographystyle{acl_natbib}

\appendix
\input{./sections/appendix}

\end{document}

%% file: sections/abstract.tex
\begin{abstract}
Contrastively trained vision-language models have achieved remarkable progress in vision and language representation learning. However, recent research has highlighted severe limitations of these models in their ability to perform compositional reasoning over objects, attributes, and relations. Scene graphs have emerged as an effective way to understand images compositionally. These are graph-structured semantic representations of images that contain objects, their attributes, and relations with other objects in a scene. In this work, we consider the scene graph parsed from text as a proxy for the image scene graph and propose a graph decomposition and augmentation framework along with a coarse-to-fine contrastive learning objective between images and text that aligns sentences of various complexities to the same image. We also introduce novel negative mining techniques in the scene graph space for improving attribute binding and relation understanding. Through extensive experiments, we demonstrate the effectiveness of our approach that significantly improves attribute binding, relation understanding, systematic generalization, and productivity on multiple recently proposed benchmarks (For example, improvements 
up to $\mathbf{18\%}$ for systematic generalization, $\mathbf{16.5\%}$ for relation understanding over a strong baseline), while achieving similar or better performance than CLIP on various general multimodal tasks.

\end{abstract}

%% file: sections/intro.tex
\section{Introduction}
\begin{figure}[h!]
    \centering
    \begin{minipage}{0.53\columnwidth}
        \centering
        \includegraphics[width=\columnwidth]{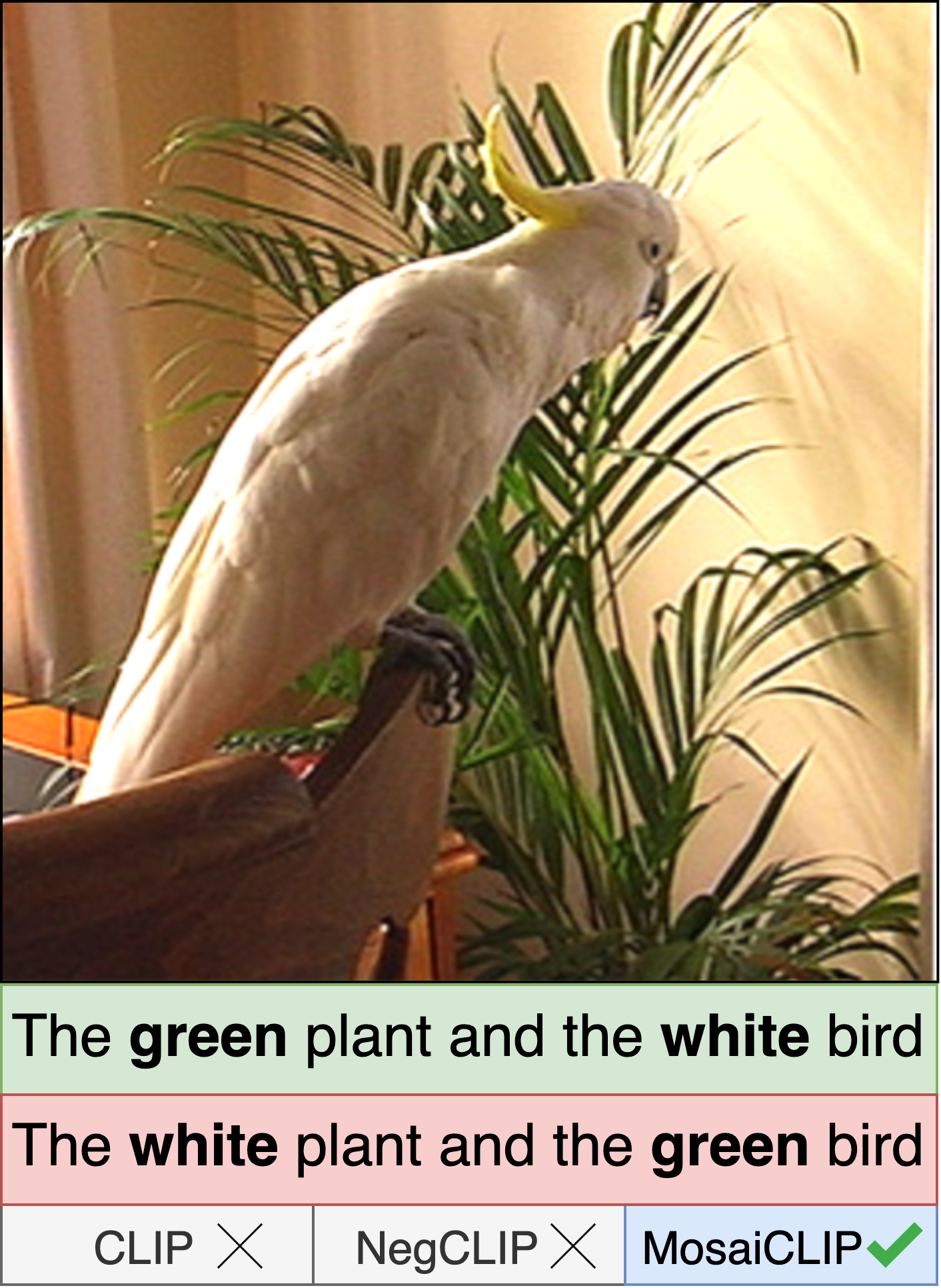}
        \subcaption{}
    \end{minipage}
    \begin{minipage}{0.44\columnwidth}
        \centering
        \includegraphics[width=\columnwidth]{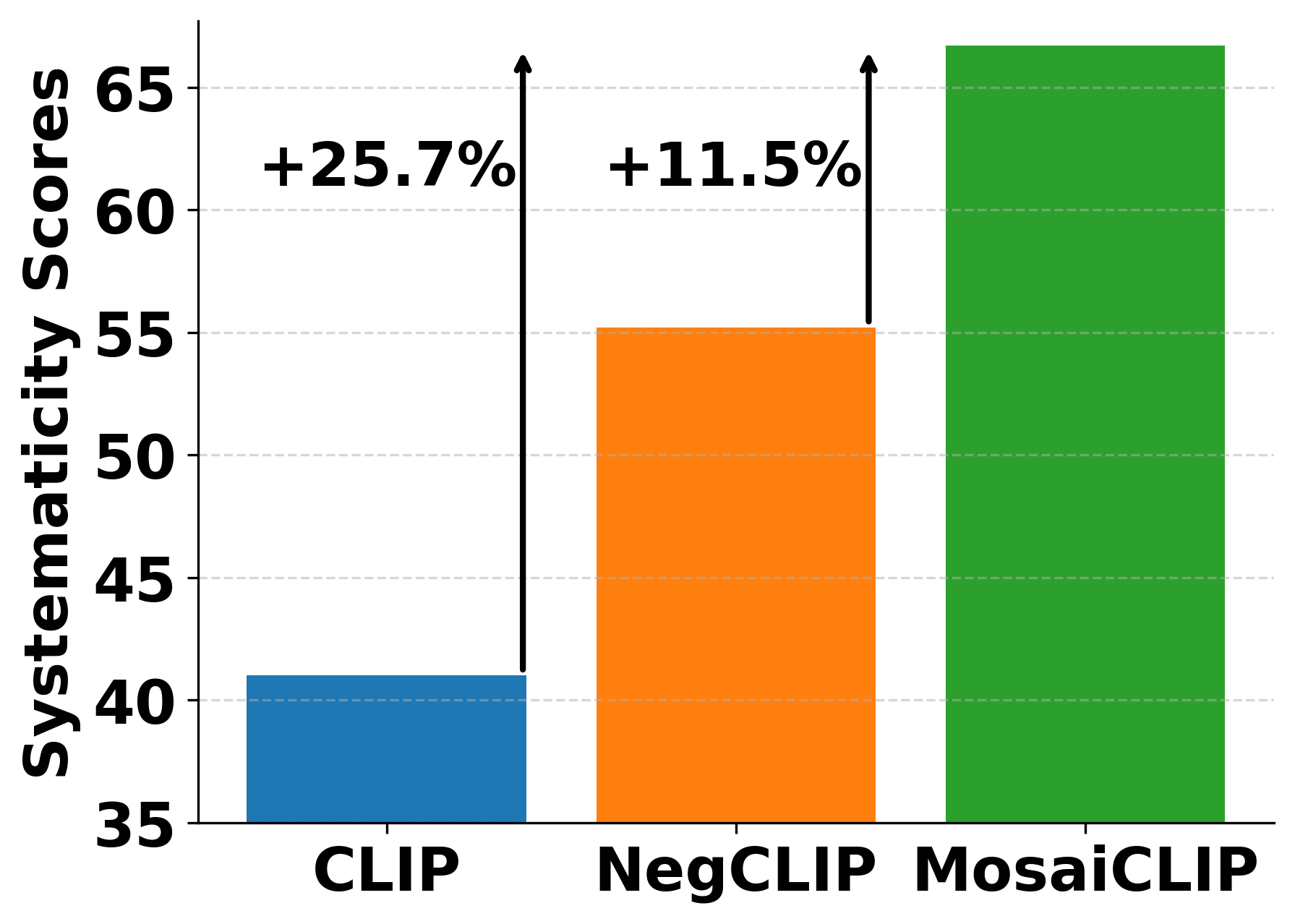}
        \subcaption{}
        \includegraphics[width=\columnwidth]{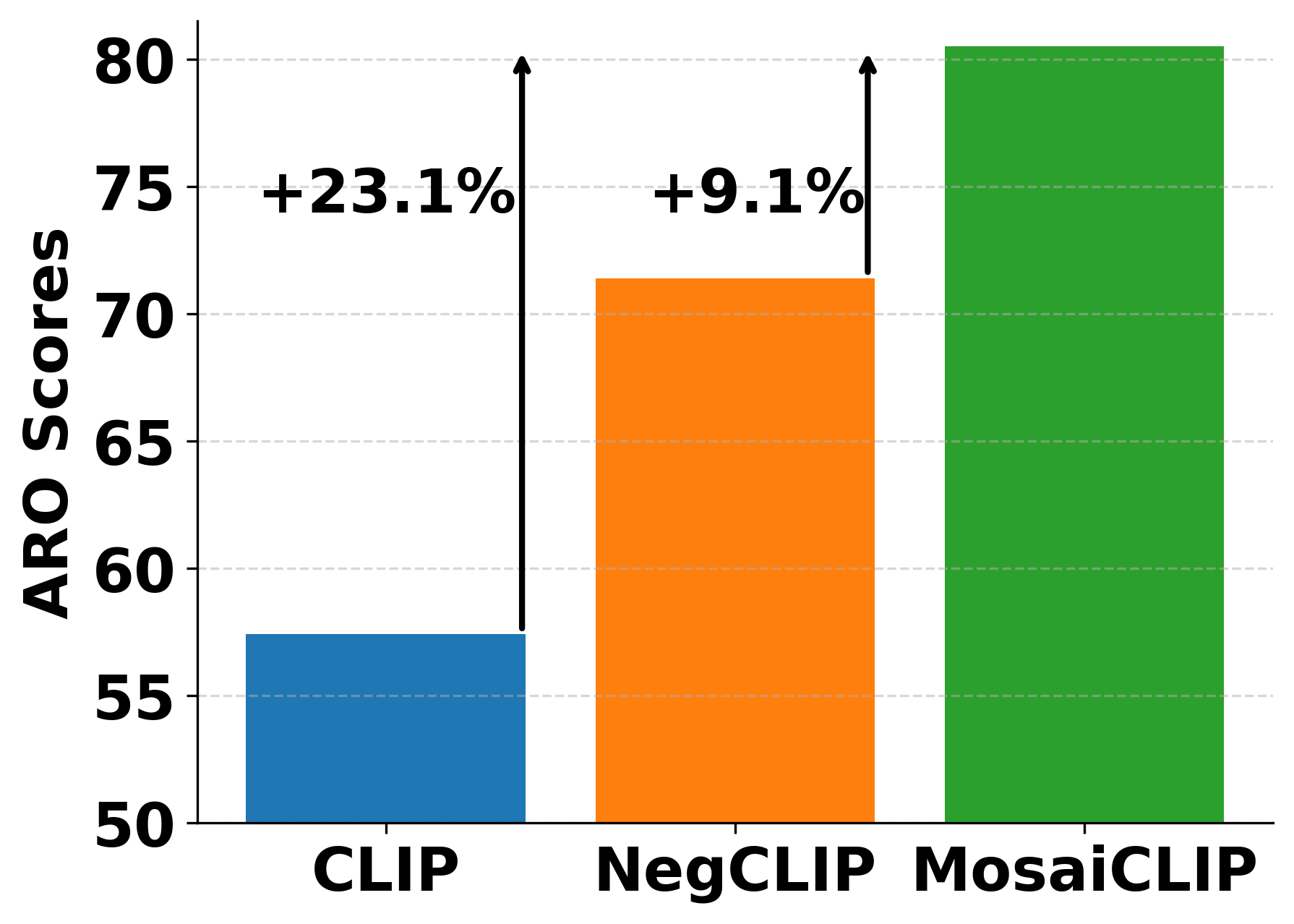}
        \subcaption{}
    \end{minipage}
    \caption{\textbf{(Left)} a) A typical example from the ARO benchmark for testing attribute understanding of VLMs. VLMs struggle with matching the image to the correct caption (in green). \textbf{(Right)} Average scores of \methodcomp{} (our method) compared with \negclip{} and \clip{} on prominent compositionality benchmarks for measuring b) Systematic Generalization c) Attribute, Relation, and Word Order understanding.}
    \label{fig:overview_results}
    \vspace{-0.5cm}
\end{figure}

Recent progress in contrastive learning using large-scale image-text data for joint image-text representation learning has led to Vision-Language models (VLMs) like CLIP \citep{radford2021learning} and ALIGN \citep{jia2021scaling} that show remarkable zero-shot classification and retrieval capabilities. However, recent works have shown that these models struggle at compositional reasoning \cite{yuksekgonul2022and, thrush2022winoground, ma2022crepe}. In particular, they struggle with binding correct attributes to the correct objects, understanding relations between objects, generalizing systematically to unseen combinations of concepts and to larger and more complex sentences.

Some works have made progress on this problem. \citet{yuksekgonul2022and} show that hard negative mining of images and text during fine-tuning is a promising first step to improving compositionality. However, performance gains are highly dependent on how clean the training data is, and generalizing to unseen combinations of concepts remains a challenge. \citet{doveh2023teaching} use LLMs for hard negative mining and \citet{cascantebonilla2023going} explore using synthetic datasets to improve compositional understanding in VLMs. Synthetic datasets lead to a domain gap compared to natural datasets. We aim to develop a general-purpose approach for improving compositionality of all such contrastively trained VLMs.

In this paper, we consider a scene graph representation of the image and text. We observe that multiple sub-graphs of the text scene graph with different semantic complexities can be matched with the same image. Performing this matching improves fine-grained and hierarchical understanding of text and thereby, of images. 
We achieve this by developing a scene graph-based text decomposition strategy that creates a scene graph for any given text, decomposing it into sub-graphs, and matching an image to multiple sentences derived from these sub-graphs (See Fig. \ref{fig:method_overview} for an overview). Each sub-graph represents a distinct part of the image, aligning well with CLIP's original image-text matching objective. Focused on improving {attribute binding} and {relation understanding}, we develop novel hard negative graph creation strategies which helps VL contrastive learning.
We provide a novel Image-to-Multi-Text contrastive loss for matching individual images to multiple sentences. Our approach of matching texts of different complexity (from coarse-grained to fine-grained) to the image leads to fine-grained and hierarchical text understanding. Our resulting model is \methodcomp{}.

Our approach leads to significant improvements across compositionality benchmarks. For example, Figure \ref{fig:overview_results} b) and c) shows that MosaiCLIP improves performance by $11.5\%$ and $9.1\%$ on CREPE and ARO dataset over a strong baseline and by $>20\%$ over CLIP. \textbf{Our contributions encompass:}
\begin{itemize}
    \item A novel graph-based text decomposition and augmentation framework and a coarse-to-fine contrastive learning objective for matching images to text sub-graphs of varying complexity.
    \item Hard-negative mining techniques using graph transformations of the text scene graphs, that are seamlessly coupled with our text decomposition strategy, and applied over any text. 
    \item A thorough \textit{analysis} for understanding why \methodcomp{} improves vision-language compositionality, disentangling the effect of image and text encoders and providing a novel tree-score based analysis showing that \methodcomp{} exhibits improved hierarchical text understanding.
    \item Extensive experiments over three model architectures, two pre-training datasets, three fine-tuning datasets and test over four compositionality benchmarks (11 datasets) to prove the efficacy of \methodcomp{} for improving compositionality.
\end{itemize}

%% file: sections/related.tex
\section{Related Work}
\label{sec_related}
\paragraph{Contrastive Vision-Language Pre-training:}
Large-scale contrastive learning for Vision and Language is utilized to create models like CLIP \citep{radford2021learning} and ALIGN \citep{jia2021scaling}. These models showcase impressive performance on a variety of tasks, including image classification, text and image retrieval, image captioning \citep{mokady2021clipcap}, object detection \citep{zhong2022regionclip, li2022grounded} etc.
\paragraph{Visio-Linguistic Compositionality:}
Various studies have introduced benchmarks for assessing the compositional reasoning abilities of vision-language foundation models (VLMs). For instance, Winoground \citep{thrush2022winoground} is a handpicked collection of 400 test cases, each comprising two images and two sentences. Sentences have the same word content and differ in word-order. \citet{diwan-etal-2022-winoground} show that the Winoground dataset tests additional challenges along with compositionality, including handling ambiguous image-text pairs and unusual examples. \citet{yuksekgonul2022and} proposed the ARO benchmark for probing VLMs ability to understand Attribute, Relations, and Word-Order. \citet{ma2022crepe} proposed CREPE for measuring two aspects of compositionality: systematic generalization and productivity. All benchmarks suggest that contrastively trained VLMs have severe difficulty in compositional reasoning. As a remedy, NegCLIP \citep{yuksekgonul2022and} and Teaching SVLC \citep{doveh2023teaching} create targeted rule-based and LLM-guided hard negative sentences, SyViC \citep{cascantebonilla2023going} fine-tunes CLIP with million scale synthetic images-text pairs, for improving relational and attribute understanding. We observe that previous methods are either highly dependent on how clean the training data is, use expensive LLM's for data augmentation or use synthetic datasets that require special solutions to resolve the synthetic-to-real domain gap. We hence develop a coarse-to-fine contrastive learning framework that matches images with texts of multiple complexities, which serves as a general-purpose solution to improve fine-grained and hierarchical text understanding, thereby improving compositionality.
\paragraph{Scene Graphs} are structured representations of visual scenes, consisting of objects, their attributes, and relationships between objects. Scene graphs are beneficial for a range of tasks including image retrieval \citep{wu2019unified, johnson2015image}, image captioning \citep{yang2019auto}, and image generation \citep{johnson2018image} among others.

%% file: sections/model.tex
\begin{figure*}[h]
    \centering
    \subcaptionbox{Text scene graph creation and decomposition (Sec. \ref{subsec_sg_decomposition}).}{\includegraphics[width=\textwidth]{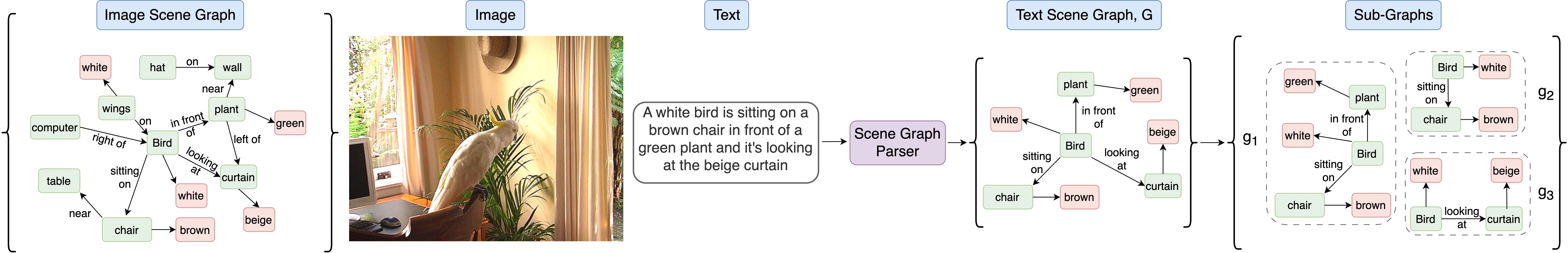}}
    \hfill
    \subcaptionbox{Hard negative sub-graph creation from positive sub-graphs (Sec. \ref{subsec_neg_graph_creation}).}{\includegraphics[width=0.33\textwidth]{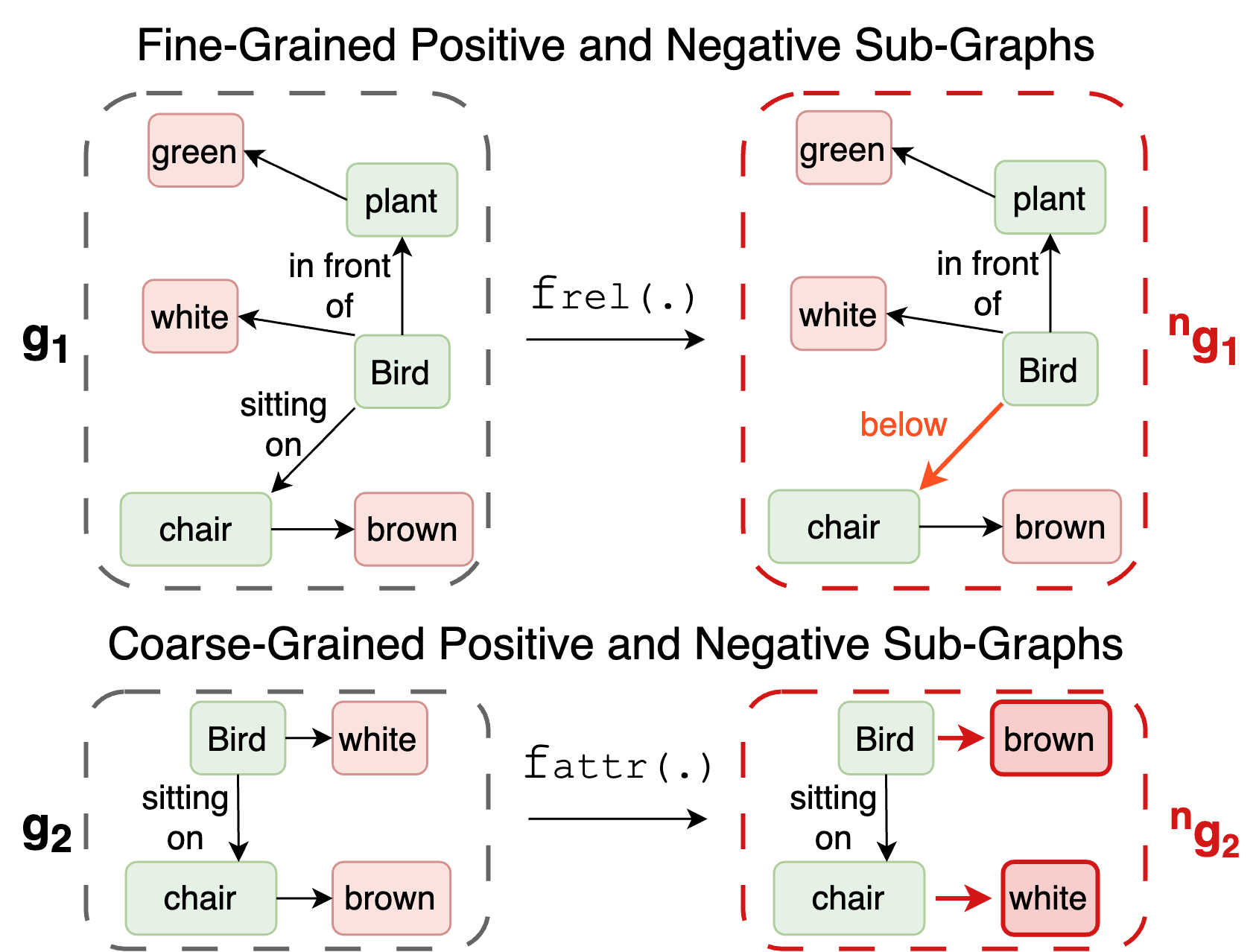}}
    \subcaptionbox{Coarse-to-fine contrastive learning (Sec. \ref{subsec_coarsetofine_loss}).}{\includegraphics[width=0.62\textwidth]{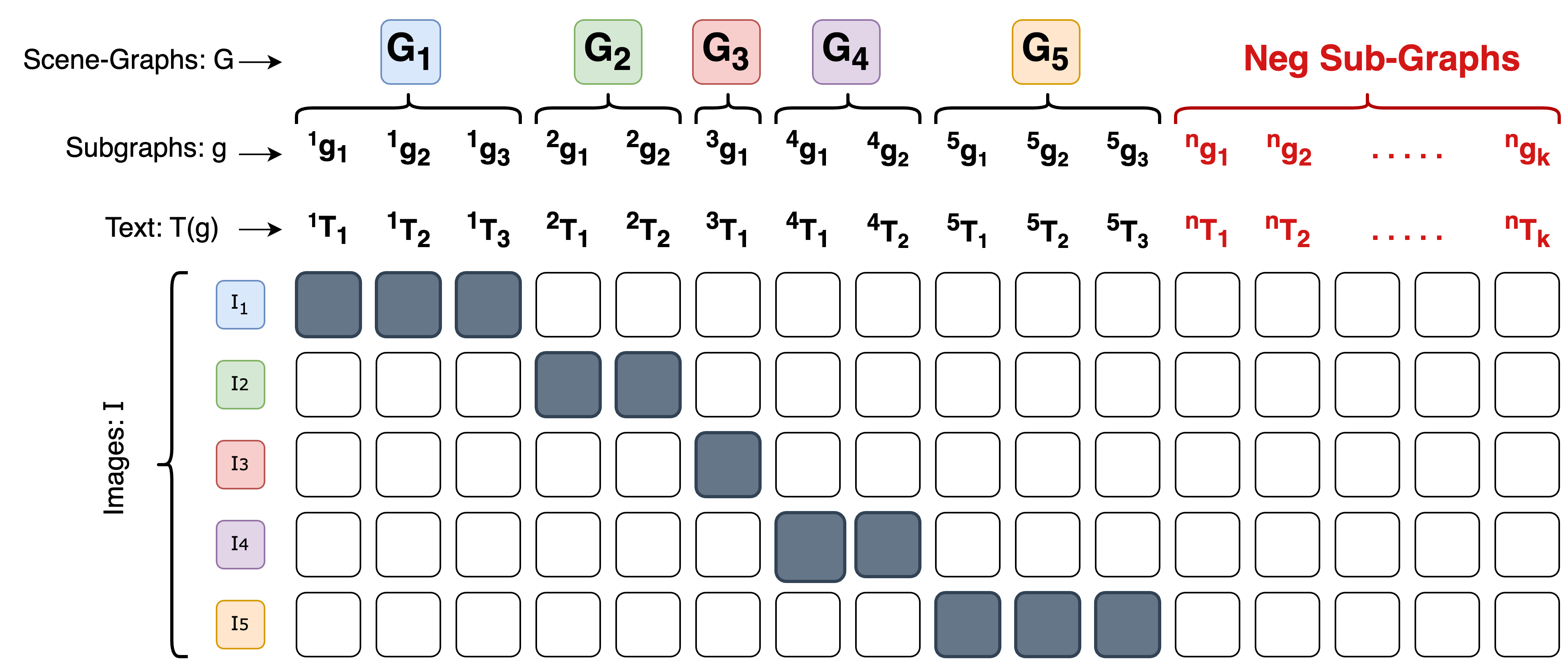}}
    \caption{Overview of our approach. \textbf{a)} Depiction of the scene graph of an image (hypothetical) and a scene graph parsed from text. The text scene graph is a sub-graph of the image scene graph. The text scene graph is decomposed into sub-graphs from which \textbf{b)} minimally perturbed hard-negative sub-graphs are created. \textbf{c)} The Ground truth similarity matrix used for a batch of data during contrastive learning. Solid boxes represent a match between the image and the corresponding text. Different from CLIP, each image can be matched to multiple texts in our method.}
    
    \label{fig:method_overview}
    \vspace{-0.3cm}
\end{figure*}
\section{Methodology}
\label{sec_method}
\subsection{Overview}
\label{subsec_method_overview}
Here we present the key high-level ideas of our approach. We first present a graph-centric view of the standard image-text matching objective in CLIP, which serves as a motivation to develop our approach (Sec. \ref{subsec_itg_alignment}). We create scene graphs derived from the text, decompose them into multiple sub-graphs (Sec. \ref{subsec_sg_decomposition}) and apply augmentations on these sub-graphs to create negative sub-graphs (Sec. \ref{subsec_neg_graph_creation}) which are used as hard negatives in a batch. Sec. \ref{subsec_coarsetofine_loss} formally defines the Image-to-Multi-Text and Text-to-Image losses used for a batch of V-L inputs which is key for learning from multiple positive and negative texts derived from sub-graphs. Matching images with coarse-to-fine sub-graphs results in improved fine-grained and hierarchical understanding of text. Sec. \ref{subsec_curriculum_training} provides a two-stage curriculum learning strategy for improved fine-tuning performance.

\subsection{Image-Text-Graph Alignment}
\label{subsec_itg_alignment}
Our approach builds on the idea that the standard image-text contrastive learning in CLIP can be viewed as a matching between an image scene graph and its sub-graph. Formally, given an image-text pair $(I, T)$, the image can be viewed by its scene graph, $\Gcal_I = (\Vcal_I, \Ecal_I)$. The text scene graph is given by $\Gcal_T = (\Vcal_T, \Ecal_T)$. Then $\Gcal_T \subset \Gcal_I$. According to this assumption, during contrastive learning in CLIP, we implicitly bring the representation of the image scene graph close to \textit{one} of its sub-graph (the text scene graph).
Now, let $S_{\Gcal} = \{g | g \subset \Gcal\}$ represent the set of sub-graphs of a graph $\Gcal$. According to the assumption above, $g \in S_{\Gcal_T} \Rightarrow g \in S_{\Gcal_I}$. Hence $\forall g \in S_{\Gcal_T}$, $(g, \Gcal_I)$ becomes a correct matching pair during contrastive learning.
We match multiple sub-graphs of the text scene graph to the same image, while also including hard negative sub-graphs in the batch. Matching between graphs is an implicit concept, and all graphs are first converted to text via templates, converted to embeddings using transformer-based (text) encoders, and matched to image embeddings.

\subsection{Scene Graph Guided Text Decomposition}
\label{subsec_sg_decomposition}
Scene graphs are succinct representations of images. However, an image scene graph generator used for generating a scene graph for any given input image is expensive to train since it requires supervised scene graph annotations for training \citep{li2017scene, xu2017scene, zhang2019graphical}, and also leads to issues like low coverage or biased generations against the long tail nature of objects and relationship annotations. We instead use the text scene graph created using an off-the-shelf text scene graph parser\footnote{https://github.com/vacancy/SceneGraphParser} \citep{wu2019unified}. This serves as a proxy for the scene graph of (part of) the image and is assumed to be a sub-graph of the image scene graph, as also depicted by Figure \ref{fig:method_overview}.

Let the text scene graph obtained be $G_T = (V_T, E_T)$, where $V_T$ represent the nodes of the graph, which are either objects or their attributes. $E_T$ are the edges of the graph that represent relations between objects. See Fig. \ref{fig:method_overview} for an example of a text scene graph. As shown in the figure, we decompose this scene graph into multiple \textit{positive} sub-graphs $P_g = \{g_1, g_2, g_3, \cdots , g_k\}$, $k \leq M$, where $M$ is the max number of decomposed sub-graphs and is a hyperparameter. Each sub-graph is a representation of a part of the image. We then convert sub-graphs to sentences so that they can be easily processed by transformer-based (text) encoders commonly used to train CLIP. For this, we use a simple template-based approach. For e.g., we create templates of the form "$\{N_1\}$ $\{R\}$ $\{N_2\}$" if we need to convert a graph having two nodes ($N_1$, $N_2$) and a relation $R$, into a sentence format. Corresponding to each sub-graph, we obtain one positive text for the image, creating a positive text set $P_t = \{t_1, t_2, t_3, \cdots , t_k\}$.

\subsection{Negative Sub-Graph Creation}
\label{subsec_neg_graph_creation}
Corresponding to sub-graphs in $P_g$, we create negative sub-graphs $N_g = \{{^{n}g_1}, {^{n}g_2}, {^{n}g_3}, \cdots \}$. Sub-graphs in $N_g$ are a minimally perturbed versions of the positive sub-graphs in $P_g$. Similar to positive sub-graphs, we convert sub-graphs in $N_g$ to text using the same template-based approach, and obtain $N_t = \{{^{n}t_1}, {^{n}t_2}, {^{n}t_3}, \cdots \}$. Texts in $N_t$ serve as hard negative texts in a given batch, see Fig. \ref{fig:method_overview}. We focus on creating negative sub-graphs that improve the attribute binding and relation understanding capabilities of the model, for which we use the following strategies for negative graph creation:
We first consider an external set of objects ($\mathcal{N}$), attributes ($\mathcal{A}$), and relations ($\mathcal{R}$). \\
\textbf{1) Node Swapping and Replacement}: We \textit{swap} nodes in sub-graphs, these can be swaps of nodes which are attributes or objects. We also \textit{replace} nodes with external nodes from $\mathcal{N}$, $\mathcal{A}$ based on their type. \textbf{2) Edge Replacement}: We \textit{replace} edges with randomly sampled edges from the external relations set, $\mathcal{R}$. \textbf{3) Connecting Sub-graphs}: Here we \textit{join} two sub-graphs. For this, we use one sub-graph from $P_g$, and another random graph created using nodes and edges sampled from external sets $\mathcal{N}, \mathcal{A}, \mathcal{R}$. This creates an overall hard negative graph. Sub-graphs are joined by simply joining nodes from both graphs through a randomly sampled edge from $\mathcal{R}$. 
These strategies result in minimally perturbed hard negative sub-graphs for improving attribute and relation understanding. We define multiple graph transformations $\{f_g: \mathbb{G} \longrightarrow P(\mathbb{G})\}$ -- $f_{rel}, f_{attr}, f_{obj}$ using the above techniques and create hard negative sub-graphs. See Appendix Sec. \ref{sg_decomp} for more details regarding negative sub-graph creation.

\subsection{Coarse-to-Fine Contrastive Learning in Image-Text-Graph Space}
\label{subsec_coarsetofine_loss}
Given an image-text batch during training $\Bcal = \{(\xv_i, \tv_i)\}_{i=1}^{n}$, consider separately the batch of images $\Bcal_I = \{\xv_i\}_{i=1}^{n}$ and a batch of texts $\Bcal_T = \{\tv_i\}_{i=1}^{n}$.
The sentences in the text batch are first converted to scene graphs to obtain a batch of scene graphs $\Bcal_G = \{\Gcal_{i}\}_{i=1}^{n}$, followed by decomposition to sub-graphs to obtain the positive sub-graph batch $\Bcal_g^{pos} = \{\gv_{i}\}_{i=1}^{m}$, $m>n$. $r$ negative sub-graphs are sampled and added to the batch to obtain $\Bcal_g = \{\gv_{i}\}_{i=1}^{m+r}$.
We convert these sub-graphs to text to obtain the final text batch $\Bcal_t = \{\tv_{i}^{g}\}_{i=1}^{m+r}$.

Consider an image encoder model $f_\theta$ parameterized by $\theta$, a text encoder $f_\phi$ parameterized by $\phi$. For any image $\xv$, text $\tv$, $\Tilde{\uv} = f_\theta(\xv)$ is the unnormalized image feature, and $\Tilde{\vv} = f_\phi(\tv)$ is the unnormalized text feature. As common practice, the features are normalized to obtain $\uv = {\Tilde{\uv}}/{\|\Tilde{\uv}\|}$ and $\vv = {\Tilde{\vv}}/{\|\Tilde{\vv}\|}$. 
\noindent The \textbf{Image-to-Multi-Text} contrastive loss is given by:
{
\begin{align*} 
\Lcal_{i2t}^{\text{\tiny \methodcompshort{}}} = & - \sum_{ i=1 }^{|\Bcal_I|} \frac{1}{ |\Pcal(i)|}  \sum_{ k \in \Pcal(i) }
\hspace{-0.2cm} \log \frac{ \exp(\tau \uv_{i}^T \vv_k)  }{\sum_{ j=1}^{|\Bcal_t|}  \exp(\tau \uv_{i}^T \vv_{j}) }
\end{align*}
}%
where $\Pcal(i) = \{ k | k \in [1, |\Bcal_t^{pos}|], \gv_k \subseteq \Gcal_i\}$. \\
The \textbf{Text-to-Image} contrastive loss is only calculated for the positive texts. It is given by:
{
\begin{align*} 
\Lcal_{t2i}^{\text{\tiny \methodcompshort{}}}	= & - \sum_{j=1}^{|\Bcal_t^{pos}|}
\log \frac{ \exp(\tau \uv_{p(j)}^T \vv_j )  }{\sum_{i=1}^{|\Bcal_I|}  \exp(\tau \uv_{i}^T \vv_{j} )  }
\end{align*}
}%
where $\gv_{p(j)} \subseteq \Gcal_j$.
$\Bcal_t = [\Bcal_t^{pos}; \Bcal_t^{neg}]$, in which $\Bcal_t^{pos}, \Bcal_t^{neg}$ represent the texts in $\Bcal_t$, obtained from positive and negative sub-graphs respectively. The overall loss is $\Lcal_{\text{\tiny \methodcomp{}}} = (\Lcal_{t2i}^{\text{\tiny \methodcompshort{}}} + \Lcal_{i2t}^{\text{\tiny \methodcompshort{}}})/2$. 

\subsection{Curriculum and Robust Fine-tuning}
\label{subsec_curriculum_training}
\noindent For fine-tuning experiments, we develop a two-stage curriculum learning strategy motivated by recent work \citep{goyal2022finetune, wortsman2022robust, kumar2022fine} that show how fine-tuning can distort pre-trained features and closely mimicking the contrastive pre-training objective while fine-tuning CLIP can help mitigate this problem \citep{goyal2022finetune}. However, our coarse-to-fine contrastive learning objective naturally deviates from pre-training in two ways. $a)$ Existence of hard negative texts in the batch, and $b)$ Having multiple positive and negative texts for an image. This can lead to a \textit{gap} in pre-training vs fine-tuning objective, and a lower than optimal performance after fine-tuning. To solve this, our two-stage curriculum learning strategy first fine-tunes the model while sampling (at max) a single positive and negative sub-graph per image, followed by fine-tuning it with multiple positive and negative sub-graphs. The hardness of data in this curriculum learning setup is defined by the amount of difference the fine-tuning setup has as compared to the pre-training setup. According to this intuition, it is easier for the model to first learn to handle hard negatives in a batch and then learn to handle multiple positive and hard negative sentences at once. We see consistent improvements using this strategy compared to a direct one-step fine-tuning, which we term as \methodcompNoCurric{} in our ablations. For better performance on non-compositonal tasks, we use the robust fine-tuning approach \citep{wortsman2022robust} of weight space ensembling of the vision encoder, before and after fine-tuning. This model is called \methodcompwiseft{}

%% file: sections/expt.tex
\section{Experiments}
\label{sec_experiments}
\noindent \textbf{Evaluation Datasets:}
We test \methodcomp{} and baselines on large scale benchmarks that require compositional reasoning: \underline{CREPE-Systematicity} \cite{ma2022crepe} measures systematic generalization, \underline{ARO} \cite{yuksekgonul2022and} measures attribute, relation and word-order understanding, \underline{SVO} \cite{hendricks-nematzadeh-2021-probing} measures verb (relation) understanding, \underline{VL-Checklist} \cite{zhao2022vlchecklist} measures relation, attribute and object understanding. We use \underline{CREPE-Productivity} \cite{ma2022crepe} for measuring model's ability to productively generalize to more complex and long sentences.
Methods for improving compositionality should be tested on general downstream tasks used to evaluate the quality of learned representations of language and vision. For this, we utilize the popular ELEVATER benchmark \cite{li2022elevater} consisting of 20 datasets and ImageNet \cite{deng2009ImageNet} following prior work \citep{doveh2023teaching}.\\
\noindent \textbf{Baselines:} We compare with all recent techniques used for improving compositionality of CLIP style models including NegCLIP \citep{yuksekgonul2022and}, Teaching SVLC \citep{doveh2023teaching} and Syn-CLIP \citep{cascantebonilla2023going} along with CLIP \citep{radford2021learning} as well as CLIP-FT (fine-tuned) on datasets we use. See Appendix Sec. \ref{baselines} for more details. 

\begin{table*}[h!]
\small
  \fontsize{7.7}{10pt}\selectfont
      \centering
      \setlength{\tabcolsep}{2.5pt}
      {
      \begin{tabular}{lccccc|ccccccc|ccccccc|c}
        \toprule
        \multicolumn{1}{l}{FineTun. data $\rightarrow$} & \multicolumn{5}{c|}{COCO} & \multicolumn{7}{c|}{CC-FT} & \multicolumn{7}{c|}{YFCC-FT} &\\
        \cmidrule(lr){2-6} \cmidrule(lr){7-13} \cmidrule(lr){14-20}
        \multicolumn{1}{l}{Benchmark} $\rightarrow$ & \multicolumn{3}{c}{\textbf{ARO}} & \multicolumn{1}{c}{\textbf{VLC}} & \multicolumn{1}{c|}{\textbf{SVO}} & \multicolumn{3}{c}{\textbf{ARO}} & \multicolumn{2}{c}{\textbf{CREPE}} & \multicolumn{1}{c}{\textbf{VLC}} & \multicolumn{1}{c|}{\textbf{SVO}} & \multicolumn{3}{c}{\textbf{ARO}} & \multicolumn{2}{c}{\textbf{CREPE}} & \multicolumn{1}{c}{\textbf{VLC}} & \multicolumn{1}{c|}{\textbf{SVO}} & Meta\\
        \cmidrule(lr){2-4} \cmidrule(lr){5-5} \cmidrule(lr){6-6} \cmidrule(lr){7-9} \cmidrule(lr){10-11} \cmidrule(lr){12-12} \cmidrule(lr){13-13} \cmidrule(lr){14-16} \cmidrule(lr){17-18} \cmidrule(lr){19-19} \cmidrule(lr){20-20}
        Method $\downarrow$ & Rel. & Attr. & Ord. & Avg. & Avg. & Rel. & Attr. & Ord. & CU & AU & Avg. & Avg. & Rel. & Attr. & Ord. & CU & AU & Avg. & Avg. & Avg.\\
        \midrule
          Random                         & 50.0 & 50.0 & 20.0 & 50.0 & 50.00 & 50.0 & 50.0 & 20.0 & 14.3 & 20.0 & 50.0 & 50.00 & 50.0 & 50.0 & 20.0 & 14.3 & 20.0 & 50.0 & 50.00 & 38.35\\[1pt]
          \midrule
          \clip{}                         & 59.8 & 63.2 & 53.3 & 70.8 & 83.58 & 59.8 & 63.2 & 53.3 & 45.1 & 35.0 & 70.8 & 83.58 & 59.8 & 63.2 & 53.3 & 39.8 & 39.5 & 70.8 & 83.58 & 60.60\\[1pt]
          \clip{}-FT                      & 58.9 & 65.3 & 38.4 & 71.3 & 90.15 & 58.1 & 63.3 & 42.7 & 45.8 & 35.6 & 70.1 & 88.56 & 51.4 & 63.1 & 25.3 & 36.4 & 38.3 & 68.9 & 85.27 & 57.73\\[1pt]
          \negclip{}                      & 81.7 & 72.7 & 85.7 & 75.6 & 90.20 & 71.5 & 65.4 & 84.5 & 53.1 & 37.5 & 72.4 & 88.36 & 57.8 & 63.1 & 52.1 & 38.8 & 39.0 & 70.4 & 83.90 & 67.57\\[1pt]
          \midrule
          \rowcolor{cyan!12}
          \methodcompbold{}             & \textbf{82.6} & \textbf{78.0} & \textbf{87.1} & \textbf{81.4} & \textbf{90.67} & \textbf{80.4} & \textbf{69.8} & \textbf{85.5} & \textbf{72.4} & \textbf{40.9} & \textbf{77.6} & \textbf{88.73} & \textbf{74.3} & \textbf{66.9} & \textbf{84.4} & \textbf{48.8} & \textbf{41.5} & \textbf{75.1} & \textbf{85.36} & \textbf{74.29}\\[1pt]
        \bottomrule
      \end{tabular}
      }
      
      \caption{Fine-tuning results on the {\color{blue} ARO}, {\color{blue} CREPE - Systematicity},  {\color{blue} VL-Checklist (VLC)} and {\color{blue} SVO} benchmark (total 10 datasets). Abbreviations -- Rel.:= VG-Relation, Attr.:= VG-Attribution, Ord:=Average of ARO-Flickr and ARO-COCO Order results, CU: HN-Comp-Unseen, AU: HN-Atom-Unseen. See Sec. \ref{results} for more details.}
      \label{clip_fine-tune_all}
      
  \end{table*}

\begin{table}[h]
\fontsize{9.4}{8pt}\selectfont
  \centering
  \setlength{\tabcolsep}{3.5pt}
  \begin{tabular}{lccc|ccc}
    \toprule
    FineTun. data $\rightarrow$ & \multicolumn{6}{c}{CC3M} \\
    \cmidrule(lr){2-7}
    Benchmark $\rightarrow$ & \multicolumn{3}{c|}{\textbf{VL-Checklist}} & \multicolumn{3}{c}{\textbf{ARO}} \\
    \cmidrule(lr){2-4} \cmidrule(lr){5-7}
    \multirow{1}{*}{Method} & Obj. & Attr. & Rel. & Rel. & Attr. & Ord. \\
    \midrule
    \clip{} & 81.6	& 67.6	& 63.1 & 59.9 & 63.6 & 53.3 \\[1pt]
    \clip{}-FT & 79.0 & 64.7 & 54.3 & 41.7 & 59.3 & 25.2 \\[1pt]
    Syn-CLIP\textsuperscript{$\dagger$}\textsuperscript{1} & - - & 70.4 & 69.4 & 71.4 & 66.9 & 65.1\\[1pt]
    Teaching SVLC\textsuperscript{$\ddagger$}\textsuperscript{2} & 85.0 & 72.0 & 69.0 & - - & - - & - -\\[1pt]
    \midrule
    \rowcolor{cyan!12}
    \methodcompbold{} & \textbf{86.4} & \textbf{73.7} & \textbf{71.9} & \textbf{83.7} & \textbf{78.0} & \textbf{79.4} \\
    \bottomrule
  \end{tabular}
  \textsuperscript{1}\citep{cascantebonilla2023going}\textsuperscript{2}\citep{doveh2023teaching}
  
  \caption{\methodcomp{} vs contemporary works that use \textsuperscript{$\dagger$}synthetic data or \textsuperscript{$\ddagger$}LLM's. See Appx. \ref{comparison_other_baselines} for details.}
  \label{tab:comparison_other_baselines_main}
  \vspace{-0.5cm}
  
\end{table}

  \begin{table*}[h!]
  \fontsize{9.35}{11pt}\selectfont
      \centering
      \setlength{\tabcolsep}{3.5pt}
      \begin{tabular}{llccccccc|ccccccc|c}
          \toprule
          \multicolumn{2}{r}{Pre-training data $\rightarrow$} & \multicolumn{7}{c|}{CC-12M} & \multicolumn{7}{c|}{YFCC-15M} \\
          \cmidrule(lr){3-9} \cmidrule(lr){10-16}
          & \multicolumn{1}{l}{Benchmark $\rightarrow$} & \multicolumn{3}{c}{\textbf{ARO}} & \multicolumn{2}{c}{\textbf{CREPE}} & \multicolumn{1}{c}{\textbf{VLC}} & \multicolumn{1}{c|}{\textbf{SVO}} & \multicolumn{3}{c}{\textbf{ARO}} & \multicolumn{2}{c}{\textbf{CREPE}} & \multicolumn{1}{c}{\textbf{VLC}} & \multicolumn{1}{c|}{\textbf{SVO}} & Meta \\
          \cmidrule(lr){3-5} \cmidrule(lr){6-7} \cmidrule(lr){8-8} \cmidrule(lr){9-9} \cmidrule(lr){10-12} \cmidrule(lr){13-14} \cmidrule(lr){15-15} \cmidrule(lr){16-16}
          \multirow{-3}{*}{\rotatebox[origin=c]{90}{Arch.}} & Method $\downarrow$ & Rel. & Attr. & Ord. & CU & AU & Avg. & Avg. & Rel. & Attr. & Ord. & CU & AU & Avg. & Avg. & Avg. \\
          \midrule
          & Random                         & 50.0 & 50.0 & 20.0 & 14.3 & 20.0 & 50.0 & 50.00 & 50.0 & 50.0 & 20.0 & 14.3 & 20.0 & 50.0 & 50.00 & 36.33\\[1pt]
          \midrule
          & \clip{} & 51.0 & 56.6 & 25.5 & 44.1 & 37.3 & 65.6 & 82.21 & 53.8 & 56.2 & 18.4 & 39.6 & 41.7 & 66.2 & 76.27 & 51.03\\[1pt]
          & \negclip{} & 82.4 & 66.8 & \textbf{59.7} & 80.3 & 39.6 & 70.0 & 82.04 & 73.6 & 58.9 & 35.5 & 47.1 & 41.5 & 66.0 & 76.10 & 62.82\\[1pt]
          \rowcolor{cyan!12}
          \cellcolor{white} \multirow{-3}{*}{\rotatebox[origin=c]{90}{Swin-T}} & \methodcompbold{} & \textbf{84.3} & \textbf{76.8} & 55.5 & \textbf{92.1} & \textbf{44.5} & \textbf{72.4} & \textbf{85.62} & \textbf{74.7} & \textbf{66.1} & \textbf{35.8} & \textbf{89.6} & \textbf{45.3} & \textbf{71.8} & \textbf{77.87}  & \textbf{69.46}\\[1pt]

          \midrule
          & \clip{} & 52.9 &	59.7 & 22.6 & 42.9 & 36.7 & 66.2 & 82.13 & 57.8 &	55.1 & 18.3 & 38.9 & 38.9 & 64.8 & 75.60  & 50.90\\[1pt]
          & \negclip{} & 80.5 & 66.5 & \textbf{60.5} & 82.0 & 41.4 & 69.5 & 82.03 & 68.0 &	58.5 & 37.1 & 67.2 & 41.5 & 66.1 & 75.18  & 64.00\\[1pt]
          \rowcolor{cyan!12}
          \cellcolor{white} \multirow{-3}{*}{\rotatebox[origin=c]{90}{RN-50}} & \methodcompbold{} & \textbf{82.0} & \textbf{78.5} & 55.4 & \textbf{92.6} & \textbf{44.4} & \textbf{72.6} & \textbf{83.86} & \textbf{76.3} & \textbf{68.9} & \textbf{38.2} & \textbf{90.2} & \textbf{45.0} & \textbf{72.3} & \textbf{77.42}  & \textbf{69.83}\\[1pt]
          \bottomrule
      \end{tabular}
      
      \caption{Pre-training results on all compositionality benchmarks (4 benchmarks, 10 datasets) over four expt. settings (two pre-training datasets, two backbones). See Table \ref{clip_fine-tune_all} for abbreviations and Sec. \ref{results} for more details.}
      \label{pre-training_results_all}
  \end{table*}
  
\noindent \textbf{Training and Evaluation Details:} \\
\noindent \underline{Fine-tuning}: NegCLIP \cite{yuksekgonul2022and} was developed by fine-tuning CLIP on the COCO dataset \citep{lin2014microsoft}, however, COCO images might overlap with benchmarks like CREPE and ARO which may lead to confounding of results. Hence we consider 2 additional similar sized fine-tuning datasets randomly sampled from CC-12M \citep{sharma-etal-2018-conceptual, changpinyo2021cc12m} and YFCC-15M \citep{thomee2016yfcc100m} and call them CC-FT, YFCC-FT. We also use CC3M \citep{sharma-etal-2018-conceptual} for comparing with recent baselines. We fine-tune the commonly used OpenAI CLIP-ViT-B32 model and report results on all datasets, except for CREPE dataset which tests the \textit{systematic generalization} for which we used OpenCLIP \cite{Ilharco_OpenCLIP_2021} models pre-trained on \{CC-12M, YFCC-15M\}, fine-tune them on \{CC-FT, YFCC-FT\}, and report results on \{CC-12M,YFCC-15M\} splits of CREPE. See Appendix \ref{eval_data_details} for more information on evaluation datasets.

\noindent \underline{Pre-training}: We pre-train \methodcomp{}, \negclip{} and \clip{} on two prominent large-scale pre-training datasets, CC-12M and YFCC-15M, and use two different backbones (ResNet-50 and Swin-Tiny) following prior work \citep{yang2022unified} and report zero-shot performance on all test datasets. See Appendix \ref{hyperparams} for hyperparameters details.

\subsection{Results}
\label{results}
In this section we provide experimental results in both pre-training and fine-tuning settings to show the efficacy of our approach. These are as follows:\\
\newline
\noindent \textbf{Fine-tuning:} Main fine-tuning results are shown in Table \ref{clip_fine-tune_all} and \ref{tab:comparison_other_baselines_main}, where we fine-tune CLIP models using our method and compare it to baselines. Notably, we see that the generalization performance on unseen compounds and atoms as measured by the CREPE dataset is up to {$18\%$} higher than NegCLIP. Additionally \methodcomp{} shows upto {$16.5\%$, $5.3\%$, ${32.3}\%$} of improvement over \negclip{} in understanding relations, attributes and word order respectively. \methodcomp{} also shows consistent improvements in the verb understanding task as measured by the SVO dataset. \textbf{Additional Comparisons}: We also compare with latest contemporary works in Table \ref{tab:comparison_other_baselines_main} and Appendix Sec. \ref{comparison_other_baselines}. We find significant improvements (upto $14\%$ on ARO) over models that use LLMs or synthetic data for making CLIP more compositonal.\\
    
\begin{figure}[h!]
    \centering
    \includegraphics[width=\linewidth]{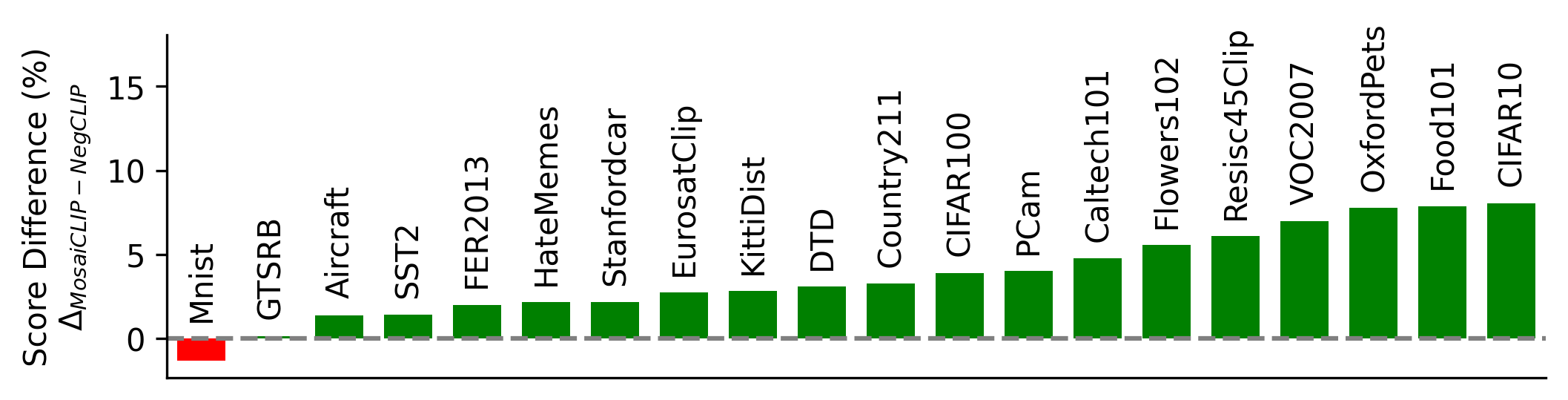}
    \caption{\methodcomp{}'s average score difference with \negclip{} on 20 datasets from {\color{blue} ELEVATER} benchmark.}
    \label{fig:20_datasets_avg_all_pretraining}
    \vspace{-0.5cm}
\end{figure}
    
\begin{figure*}[h!]
    \centering
    \begin{minipage}[b]{0.19\textwidth}
        \centering
        \subcaptionbox{CREPE-Productivity}{\includegraphics[width=\textwidth]{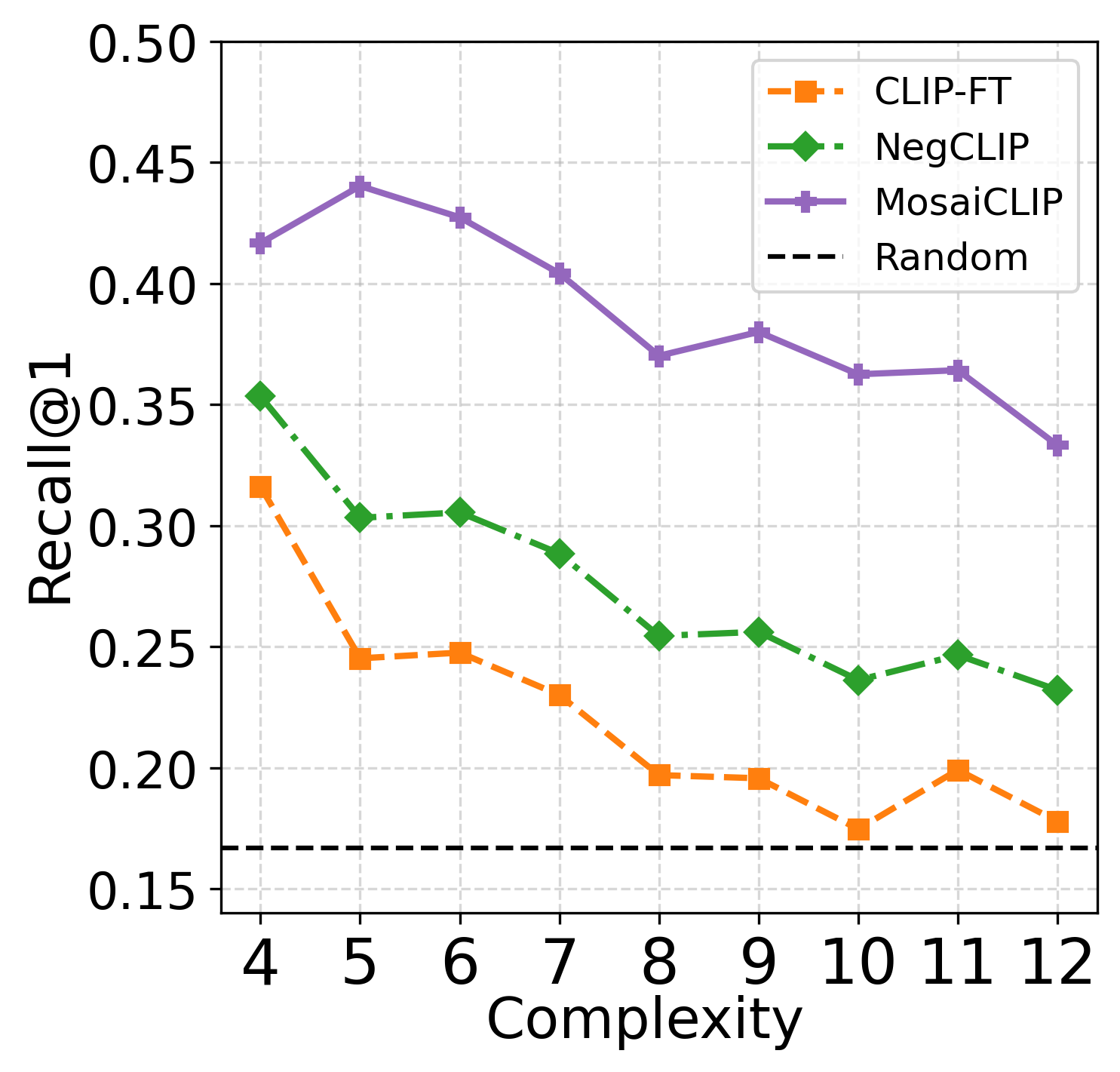}
        }
    \end{minipage}
    \hfill
    \vline
    \hfill
    \begin{minipage}[b]{0.24\textwidth}
        \centering
        \subcaptionbox{Tree Scores}{\includegraphics[width=\textwidth]{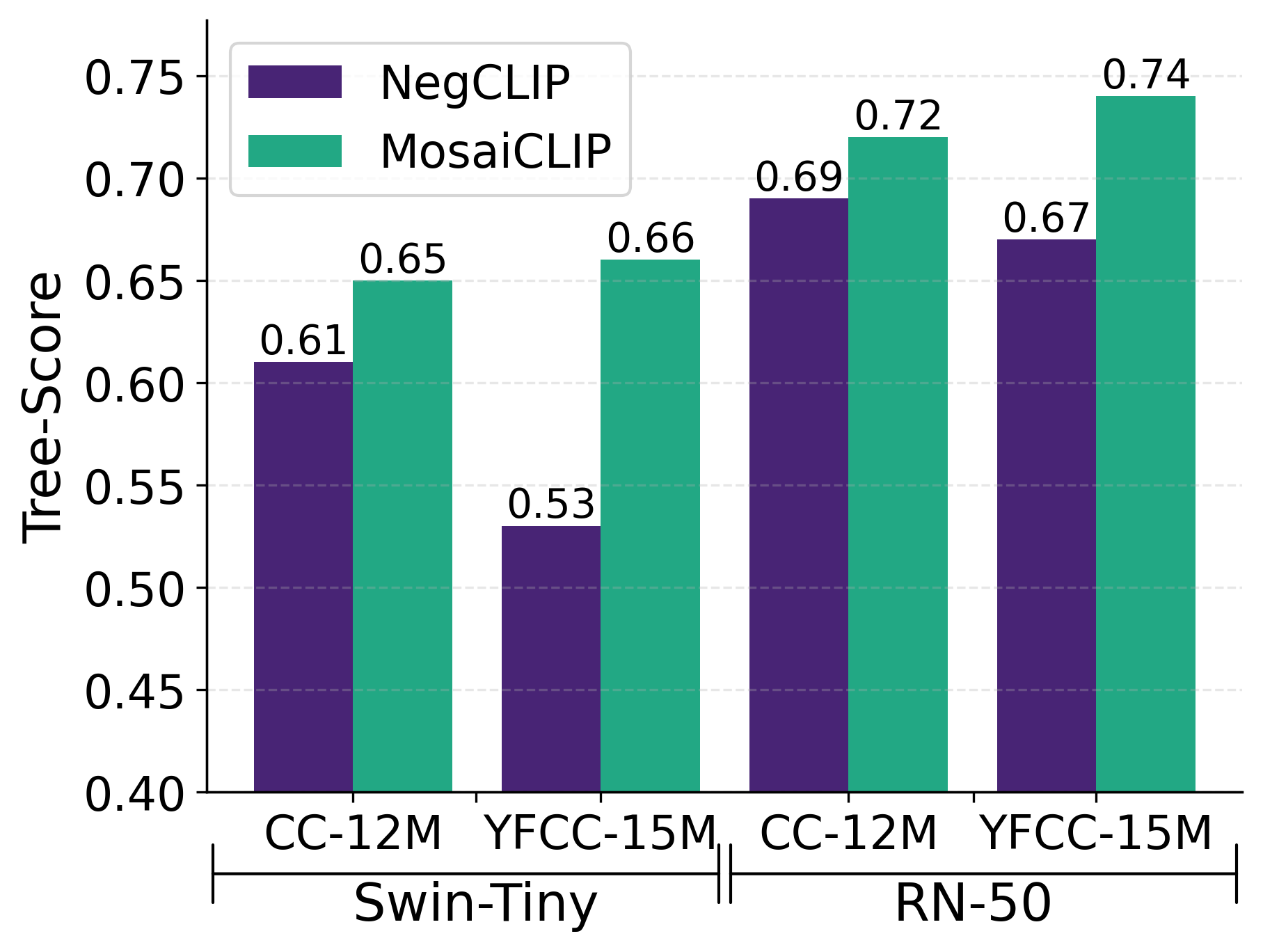}}
    \end{minipage}
    \hfill
    \vline
    \hfill
    \begin{minipage}[b]{0.23\textwidth}
        \centering
        \subcaptionbox{ARO-Relation}{\includegraphics[width=\textwidth]{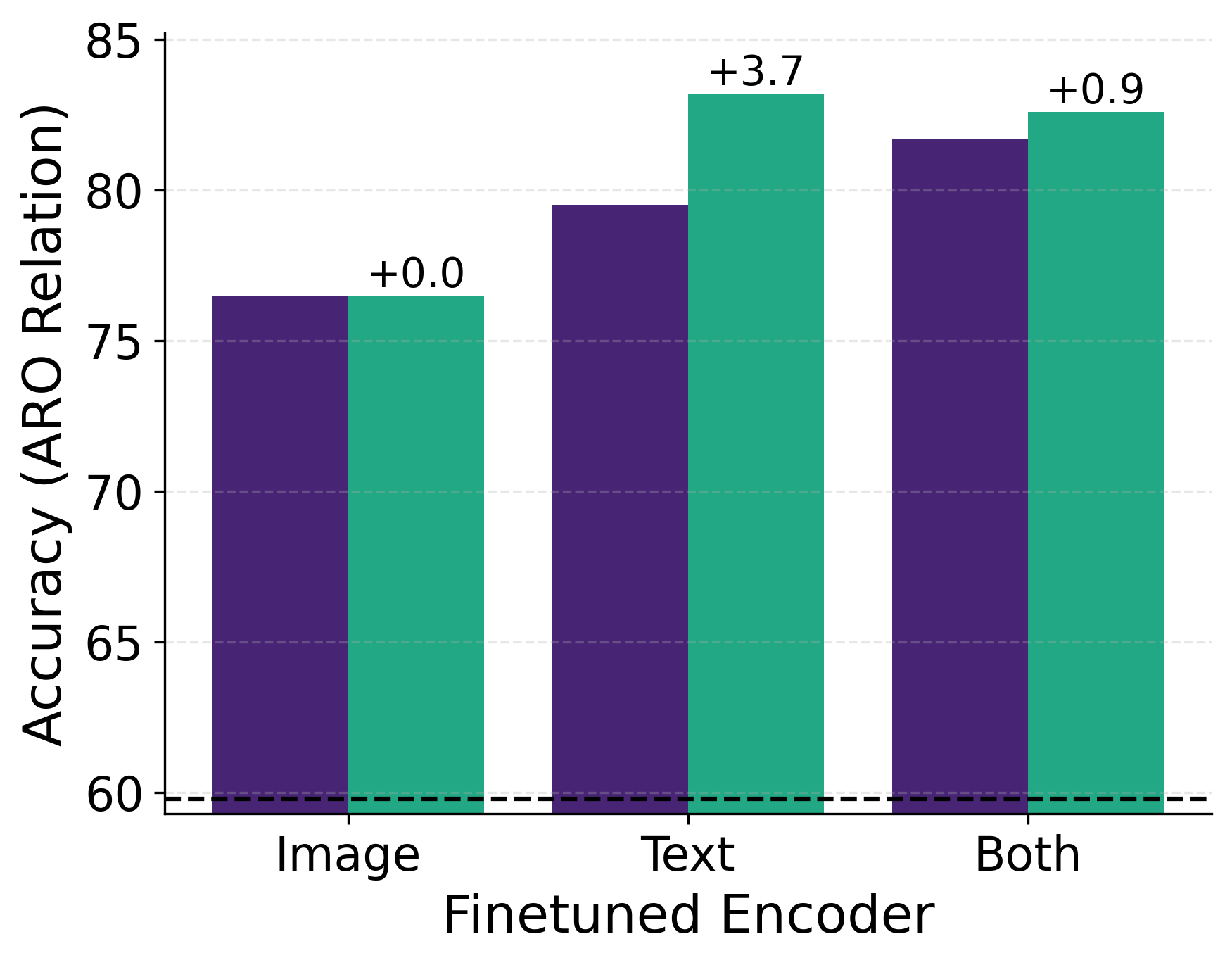}}
    \end{minipage}
    \hfill
    \begin{minipage}[b]{0.23\textwidth}
        \centering
        \subcaptionbox{ARO-Attribute}{\includegraphics[width=\textwidth]{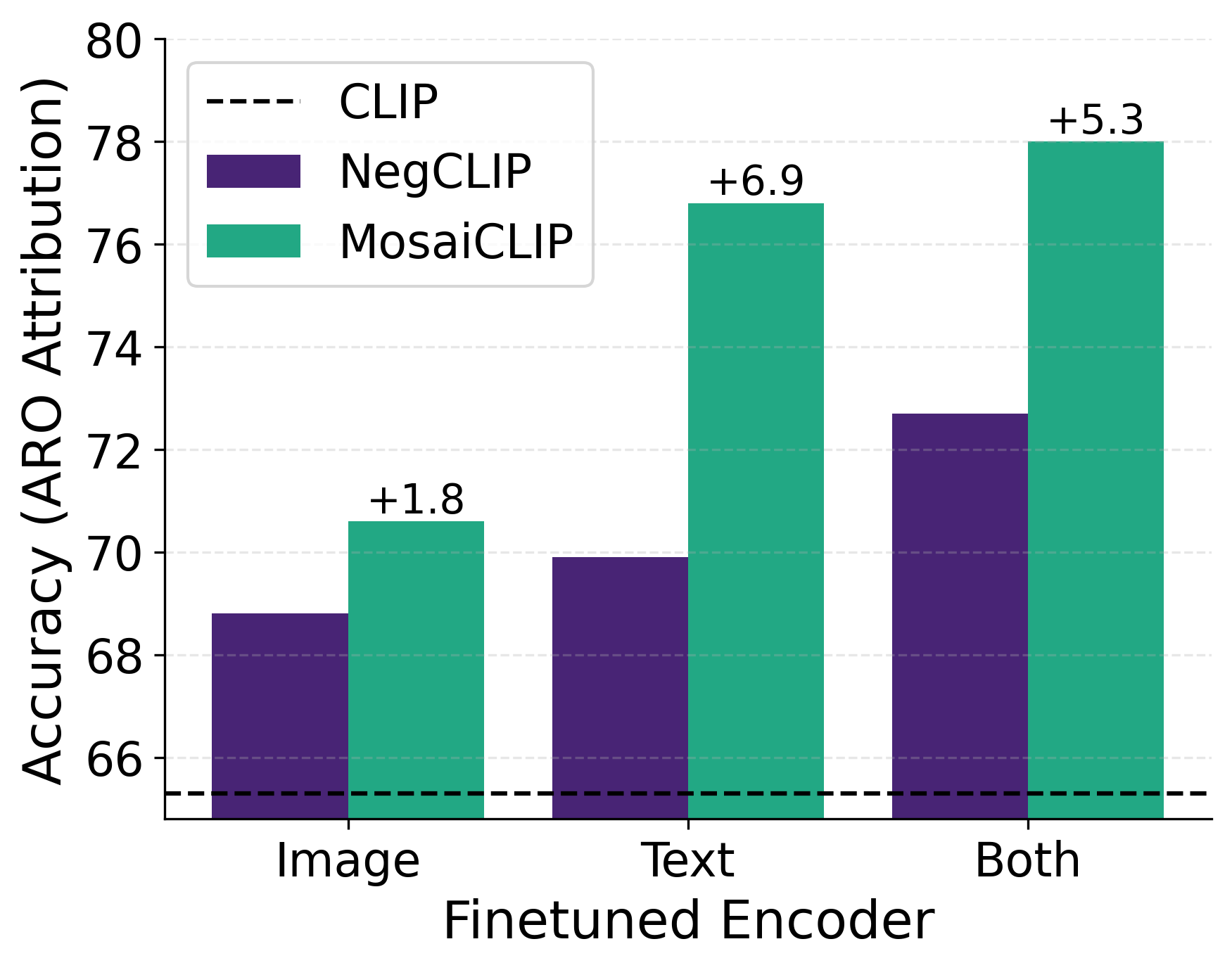}}
    \end{minipage}
    \hfill
    \caption{\textbf{a)} Results on {\color{blue} CREPE-Productivity} dataset \textbf{b)} Tree-score comparison of \methodcomp{} with \negclip{}: \methodcomp{} shows improved hierarchical understanding of language. \textbf{c) and d)} Selectively fine-tuning of image, text encoders and measure performance on different datasets. Also see similar results for SVO in Figure \ref{fig:freezing_expts__SVO}.}
    \label{fig:freezing_expts_ourmethod_and_tree_score_clip_pretr}
    \vspace{-0.5cm}
\end{figure*}

\noindent \textbf{Pre-training:} Table \ref{pre-training_results_all} shows pre-training results over all benchmarks. CREPE results show a significant gain in ability to systematically generalize to unseen combinations of concepts. Across pre-training settings, \methodcomp{} improves over \negclip{} by up to {$42.5\%, 4.9\%$} when evaluated against HN-Comp (CU), HN-Atom (AU) hard negatives respectively. Significant improvements are observed in attribute and relation understanding, giving gains of up to {$8.3\%, 12.0\%$} respectively across pretraining settings. We also note that order understanding of \methodcomp{} is worse than that of \negclip{} for the CC-12M pre-training dataset, while better than \negclip{} for the YFCC-15M dataset. Notably, there is a large variance in \negclip{}'s performance across pre-training datasets as seen in Table \ref{pre-training_results_all}, and it also performs poorly when the pre-training dataset has higher noise (e.g. YFCC-15M). \methodcomp{} is fairly consistent and more robust to the change in the pre-training dataset. In Appendix \ref{data_eff} we find that \methodcomp{} can provide improvements over \negclip{} while using as low as \textbf{0.3x} of the total pre-training or fine-tuning data.\\
\newline
\noindent \textbf{Results on classification and retrieval}:
\begin{table}[h]
\fontsize{10.5}{13pt}\selectfont
    \centering
    \begin{tabular}{lcccc|c}
    \toprule
    {Model} & \multicolumn{2}{c}{\textbf{COCO}} &  \multicolumn{2}{c}{\textbf{Flickr30K}} & {AVG.}\\
    \cmidrule(lr){2-3} \cmidrule(lr){4-5}
    & {I2T} & {T2I} & {I2T} & {T2I} & \\
    \midrule
    CLIP & 20.7 & 13.1 & 36.2 & 24.1 & 23.5 \\
    NegCLIP & 20.1 & 12.9 & 38.6 & 23.3 & 23.7 \\
    \rowcolor{cyan!12}
    MosaiCLIP & \textbf{25.9} & \textbf{16.5} & \textbf{44.5} & \textbf{29.5} & \textbf{29.1} \\
    \bottomrule
    \end{tabular}
    \caption{Comparison of Recall@1 scores of \methodcomp{} with \negclip{} and \clip{}. All models are pre-traind on YFCC-15M with swin-Tiny backbone}
    \label{classification-results}
    \vspace{-1.5em}
\end{table}
On average, \methodcomp{} achieves $+3.3\%, +6.3\%$ better performance on the ELEVATER classification benchmark compared to \negclip{} and \clip{} while pre-training and maintains similar accuracy as \clip{} while fine-tuning. We also try using our method along with the robust fine-tuning technique (WiSE-FT) so that performance degradation during fine-tuning is minimal, as shown in Appendix Table \ref{ft_zs_21_table}. See Fig. \ref{fig:20_datasets_avg_all_pretraining} for average results on ELEVATER over four training settings and Table \ref{classification-results} for results on retrieval benchmarks where we see a $+5.4$ point improvement over \negclip{}. We use the popular Karpathy splits having a 5K and 1K sized test set for obtaining the COCO and Flickr30k retrieval scores respectively.
Hence \methodcomp{}'s training strategy improves or maintains the quality of learned representations while improving compositonality. Figures \ref{fig:20_datasets_swin_CC}-\ref{fig:20_datasets_rn50_YFCC} show detailed results on ELEVATER.\\
\newline
\noindent \textbf{Productivity}:
As defined by \citet{ma2022crepe}, a productive VL model can handle arbitrarily long and complex sentences and is an important aspect of compositionality. Although we do not explicitly train our models for generalization to longer sentences, the improved hierarchical language understanding using our methods lead to an emergent behavior such that \methodcomp{} generalizes better than \negclip{} and \clip{} to more complex sentences. We can see this effect in Fig. \ref{fig:freezing_expts_ourmethod_and_tree_score_clip_pretr} a) and Appendix Fig. \ref{fig:Productivity_clip_ft_cc_coo} and \ref{fig:Productivity_clip_pretr}. We report the average of retrieval over swap and atom splits and find \methodcomp{} significantly improves over \negclip{} by upto $15\%$ across different text complexities (4-12). \\
\newline
\noindent \textbf{Application to more advanced VLMs:}
While our focus in this work has been on CLIP style, dual encoder models due to their various benefits, we believe our methods are model agnostic and aimed at improving contrastive learning through our coarse-to-fine learning framework and negative mining techniques. In this section we test our model on an advanced VLM, BLIP.
We modified BLIP’s original image-text contrastive learning objective and create two variants, one called BLIP+NegCLIP where we use NegCLIP style hard negatives and the other BLIP+MosaiCLIP which uses our methods of scene graph guided text decomposition and negative sub-graph creation. We fine-tune BLIP model taken from the official BLIP repository and use the “BLIP w/ ViT-B and CapFilt-L model (pre-trained on 129M examples)” as our base model. Results for fine-tuning experiment using COCO dataset is shown in Table \ref{blip-expts}. We use the hyperparameters used by the official codebase (for the task of fine-tuning on COCO dataset for image-text retrieval). For each setting, we report performance of four models, namely BLIP (before fine-tuned version), BLIP-FT (vanilla fine-tuned version), BLIP+NegCLIP, BLIP+MosaiCLIP. The model are evaluated on the ARO dataset to measure attribute, relation and word-order understanding, using the evaluation scripts provided by the authors of the dataset \citep{yuksekgonul2022and}.
\begin{table}[h]
    \centering
    \begin{tabular}{lcccccc}
    \toprule
    \textbf{Model} & \textbf{Rel} & \textbf{Attr} & \textbf{Ord} & \textbf{Avg} \\
    \midrule
    BLIP & 53.5 & 91.0 & 53.5 & 66.0 \\
    BLIP-FT & 58.9 & 88.4 & 58.9 & 68.7 \\
    BLIP+NegCLIP & 63.6 & 90.7 & 63.6 & 72.6 \\
    \rowcolor{cyan!12}
    BLIP+MosaiCLIP & \textbf{69.9} & \textbf{91.1} & \textbf{69.9} & \textbf{77.0} \\
    \bottomrule
    \end{tabular}
    \caption{Comparison of BLIP \citep{li2022blip} and fine-tuned version of BLIP with BLIP models that have integrated \negclip{} and \methodcomp{} methodology while training. Fine-tuning has been performed on COCO.}
    \label{blip-expts}
\end{table}
We find that compared to vanilla fine-tuning, both NegCLIP and MosaiCLIP methodologies bring improvements to relation and word order understanding, while maintaining or improving performance on attribute understanding. The MosaiCLIP methodology significantly improves relational reasoning performance and word-order understanding compared to the NegCLIP methodology, up to 6.3\%. Attribute understanding performance remains nearly the same as the baseline BLIP performance, with the MosaiCLIP methodology bringing in slight gains over NegCLIP’s methodology. On average MosaiCLIP’s methodology brings more improvements to BLIP than NegCLIP or vanilla fine-tuning.

\subsection{Analysis}
\label{sec_analysis}
We provide a detailed analysis of our models and baselines, across different dimensions as follows:\\
\newline
\noindent\textbf{Disentangling \methodcomp{} improvements:}
We quantify the relative importance of the vision and language side by freezing the language and vision encoder individually while fine-tuning all models. See Fig. \ref{fig:freezing_expts_ourmethod_and_tree_score_clip_pretr} c,d for the results. 
Notably, we find that \textbf{1)} \uline{Language encoder has significant scope for improvement over \negclip{}'s language encoder}, and \methodcomp{} is able to successfully exploit this potential and deliver an enhanced compositional understanding of language, which is evident by performance increase of $+3.7, +6.9\%$ over \negclip{} when only the language encoder is fine-tuned, as shown in Fig. \ref{fig:freezing_expts_ourmethod_and_tree_score_clip_pretr} c,d.
\textbf{2)} \uline{Improvements brought by \methodcomp{} over \negclip{} in the text encoder are always higher than improvements in the image encoder}. This is evident from Fig. \ref{fig:freezing_expts_ourmethod_and_tree_score_clip_pretr} c,d where the performance increase over \negclip{} when only the language encoder is fine-tuned is always higher as compared to when only the image encoder is fine-tuned; for example, $3.7\% > 0.0\%$, $6.9\% > 1.8\%$ for ARO-Relation, ARO-Attribution.
\textbf{3)} \uline{\methodcomp{} brings significant improvements on the image encoder side} (higher than \negclip{}) \textit{without} using any image negative mining, unlike \negclip{}.
\begin{figure*}[h!]
    \centering
    {\includegraphics[width=\textwidth]{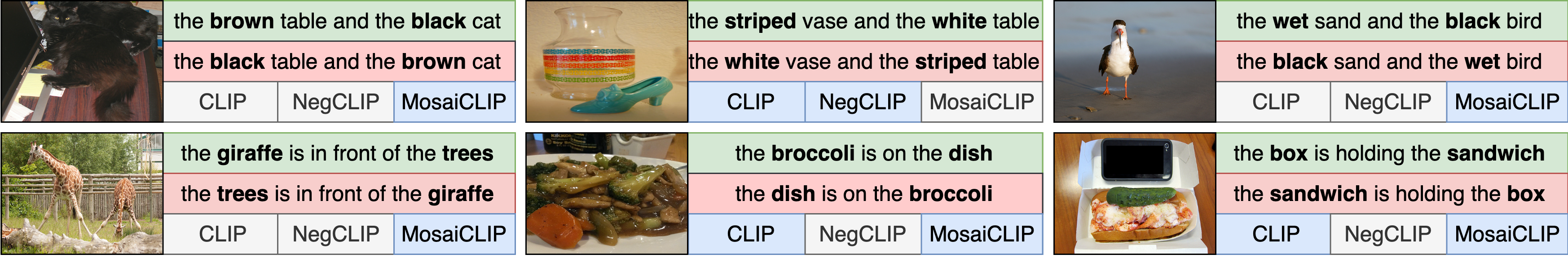}}
    \caption{Qualitative analysis on ARO dataset (Top:ARO-Attribution, Bottom: ARO-Relation). Models highlighted in blue match the image to the \colorbox{lime}{correct sentence} (in green) while the models in white match the image to the \colorbox{pink}{incorrect sentence} (in red). Here, models are taken from our fine-tuning experiments on COCO from Table \ref{clip_fine-tune_all}.}
    \label{fig:quali_aro}
    \vspace{-1em}
    
\end{figure*}

\noindent\textbf{\methodcomp{} improves hierarchical text understanding:}
For further understanding \methodcomp{}'s improved compositional understanding, we provide a novel analysis by considering the recently proposed Tree-Score \citep{murty2022characterizing} that measures the degree to which a transformer (text) encoder processes text in a hierarchical manner. We hypothesize that having tree-like hierarchical computation over language can be one leading factor for explaining the compositionality (or lack thereof) of CLIP-like models. Along with this, we have previously shown that the language encoder has the most prominent effect in improving compositionality in the case of \methodcomp{} . These two reasons motivate the use of tree-score to compare the language encoder's hierarchical understanding capability. Fig. \ref{fig:freezing_expts_ourmethod_and_tree_score_clip_pretr} a) shows that \uline{\methodcomp{}'s language encoder has higher tree-scores than \negclip{}'s language encoder}, suggesting that \methodcomp{} performs more tree-like computations. This explains the improved language compositionality of \methodcomp{} since a hierarchical tree-structured computation allows the language encoder to better understand input text compositionally, thereby improving vision-language compositionality. This is in line with the hypothesis that human's semantic understanding of sentences involves a hierarchical (tree-structured) computation which has significant evidence \citep{crain,hale-etal-2018-finding,pallier2011cortical} and this leads to their compositional generalization capability.\\
\newline
\noindent \textbf{\methodcomp{} is Robust:}
Noisy texts often have meaningful sub-texts which can be exploted by \methodcomp{}, hence \methodcomp{} often achieves consistent performance increase regardless of noise in the pre-training or fine-tuning dataset. For example, \negclip{} achieves significantly low performance on ARO when fine-tuned with YFCC-FT (having more noise in text) as compared CC-FT or COCO as shown in Table \ref{clip_fine-tune_all}. \negclip{} takes a $>10\%$ hit in performance across various ARO datasets when the fine-tuning dataset is changed from COCO to YFCC, whereas, \methodcomp{} achieves similar performance using both datasets. Appendix Sec. \ref{robustness_detailed_results} shows that pre-trained MosaiCLIP is robust to natural distributon shifts.\\
\newline
\noindent \textbf{Qualitative Analysis:}
We take \methodcomp{}, \negclip{} and \clip{} fine-tuned on COCO and filter out examples from the ARO dataset where \methodcomp{} and \negclip{}'s disagree. Some notable examples in Fig. \ref{fig:quali_aro} include cases where \negclip{} and \clip{} often struggle to understand simple concepts like understanding the color of the cat and table ({top-left} Fig. \ref{fig:quali_aro} or understanding the "is holding" relation b/w sandwich and the box in {bottom-right} Fig. \ref{fig:quali_aro}.
\vspace{-0.3em}
\subsection{Ablations}
\label{abl_main_paper}
Table \ref{ablations_cc_ft} and Appendix Tables \ref{clip_fine-tune_all_curric_rob_abl},\ref{ft_zs_21_table} show the effect of curriculum learning and robust fine-tunining where we find that curriculum learning can bring consistent improvements of up to $1.2\%$ on average and robust-finetuning (WiSE-FT) technique performs the best on zero-shot tasks (i.e. minimal forgetting while fine-tuning), while still improving over \negclip{} by about $5\%$ on compositional reasoning tasks. Table \ref{ablations:different_kind_subgraphs} shows the effects of different kinds of sub-graphs sampled during training. 
More details including the effect of sampling larger number of sub-graphs are presented in Appendix Sec. \ref{sec_ablations}.

\label{main_paper_ablations}
\begin{table}[h!]
\small
  \fontsize{7.5}{10pt}\selectfont
      \centering
      \setlength{\tabcolsep}{2.5pt}
      {
      \begin{tabular}{lccccccc|c}
        \toprule
        \multicolumn{1}{l}{Benchmark $\rightarrow$} & \multicolumn{3}{c}{\textbf{ARO}} & \multicolumn{2}{c}{\textbf{CREPE}} & \multicolumn{1}{c}{\textbf{VLC}} & \multicolumn{1}{c|}{\textbf{SVO}} & Meta \\
        \cmidrule(lr){2-4} \cmidrule(lr){5-6} \cmidrule(lr){7-7} \cmidrule(lr){8-8}
        Method $\downarrow$ & Rel. & Attr. & Ord. & CU & AU & Avg. & Avg. & Avg. \\
        \midrule
          \methodcompbold{}             & \textbf{80.4} & \textbf{69.8} & \textbf{85.5} & \textbf{72.4} & 40.9 & 77.6 & 88.73 & \textbf{73.6}\\[1pt]
          \methodcompNoCurricbold{}     & 79.0 & 69.6 & 80.6 & 71.1 & 40.2 & \textbf{77.7} & \textbf{88.91} & 72.4\\[1pt]
          \methodcompwiseftbold{}       & 78.8 & 69.4 & 82.6 & 67.5 & \textbf{41.2} & 76.4 & 88.08 & 72.0\\[1pt]
        \bottomrule
      \end{tabular}
      }
      
      \caption{Effect of Curriculum learning and Robust Fine-tuning (\methodcompwiseftbold{}) using CC-FT data.}
      \label{ablations_cc_ft}
      \vspace{-1em}
\end{table}

\begin{table}[h!]
\scriptsize
  \fontsize{8.5}{10pt}\selectfont
      \centering
      \setlength{\tabcolsep}{4.2pt}
      {
    \begin{tabular}{lrr|rr|rr}
      \toprule
      Fine-tuning data $\rightarrow$ & \multicolumn{2}{c}{COCO} & \multicolumn{2}{c}{CC-FT} & \multicolumn{2}{c}{YFCC-FT} \\
      Method $\downarrow$ & Rel. & Attr. & Rel. & Attr. & Rel. & Attr. \\
      \midrule 
      \methodcomp & \textbf{82.6} & \textbf{78.0} & \textbf{80.4} & \textbf{69.8} & \textbf{74.3} & \textbf{66.9} \\[1pt]
      \midrule 
      without $f_{rel}$ & 81.7 & 76.6 & 78.8 & 68.7 & 73.5 & 66.2\\[1pt]
      without $f_{attr}$ & 77.7 & 73.2 & 70.5 & 68.2 & 69.0 & 65.9\\[1pt]
      without $f_{rel}$, $f_{attr}$ & 79.0 & 70.4 & 68.8 & 64.9 & 57.4 & 63.6\\[1pt]
      \bottomrule
    \end{tabular}
    }
    
    \caption{Effect of different positive-negative sub-graph types sampled while training. Results are presented on the {\color{blue} ARO} benchmark.
    }
    \vspace{-1em}
    \label{ablations:different_kind_subgraphs}
\end{table}

%% file: sections/conclusion.tex
\vspace{-0.3em}
\section{Conclusion}
\label{sec_conclusion}
We present a method to improve the compositional reasoning capabilities of contrastively trained large vision-language models. In particular, we provide a coarse-to-fine contrastive learning framework and a scene graph-based text decomposition strategy for matching subgraphs of the text scene graph having varying complexity to an image during contrastive learning. We also develop hard negative graph creation strategies focused on improving attribute binding and relation understanding capabilities. Our techniques leads to significant improvements in compositional reasoning capabilities. We investigate the reasons for improved compositionality and present a novel finding based on language encoder tree-scores, suggesting that our models learn improved fine-grained and hierarchical text understanding, which is likely the key reason for improved vision and language compositionality of \methodcomp{} as compared to baselines.

%% file: sections/limitations.tex
\section{Limitations}
\label{sec_limitations}
\underline{Computational Cost:} Although \methodcomp{} leads to significant performance increase on several benchmarks that test compositional reasoining, it requires a higher per-batch computational cost while training. For this we give a detailed analysis on the computational cost in Appendix \ref{computational_cost} and show that simply providing more compute to prior methods in the form of larger batch sizes does not improve compositional reasoning. We also show ways to tackle this computational cost, by using less data in Appendix \ref{data_eff}, since \methodcomp{} is data efficient and can provide improvements over baselines with as low as 0.3x of the total data. This along with our ablations in Appendix \ref{more_subgraphs_ablation} gives some control to any practitioner to vary either the training dataset size or the number of sub-graphs in our method, and obtain a clean tradeoff between accuracy and compute. As future work we would like to develop a coarse-to-fine grained objective requiring minimal extra computation cost per batch. Future work should also look at decreasing the extra computational cost incurred by contemporary methods like Syn-CLIP \citep{cascantebonilla2023going} and Teaching SVLC \citep{doveh2023teaching}.\\

\noindent \underline{Other Vision Language Models:} In our current work we primarily aim to improve the compositionality of CLIP-Style, dual-tower models trained using large scale \textit{contrastive learning}, since they severely lacked compostional reasoning capabilities as shown by \citep{yuksekgonul2022and}. Many other VLMs exist such as those that undergo cross modal interactions between vision and language such as BLIP \citep{li2022blip}, X-VLM \citep{zeng2021multi}, LXMERT \citep{tan-bansal-2019-lxmert}. Although our methods show promise in improving more advanced VLMs like BLIP as shown in Section \ref{sec_experiments} and Table \ref{blip-expts}, a more thorough analysis will be beneficial to study the extent to which our methods can improve vision-language contrastive learning for these models.\\

\noindent \underline{Sentence Templates:} For simplicity, we currently use manually curated templates to convert sub-graphs to sentences, however, this can lead to similar looking and synthetic sentences. Large language models like GPT-4 \citep{openai2023gpt4}, BLOOM \citep{bloom} should be looked into for developing sentences from scene-graphs, by directly giving the LLM a scene-graph as input and requiring it to generate a sentence. This approach might be effective but may also lead to higher computational cost while training.

%% file: sections/ack.tex
\section*{Acknowledgements}
We thank anonymous reviewers for their insightful suggestions that helped in greatly improving our paper. We also thank Aditi Khandelwal, Animesh Sinha, Abhishek Kadian for their helpful comments and suggestions on this work.

%% file: sections/appendix.tex
\clearpage
\hspace{-0.4cm}\textbf{\LARGE Appendix}

\section{Background}
\label{sec_background}
\textbf{Contrastive Language-Image pre-training \citep{radford2021learning}} (CLIP) aims to learn general-purpose representations of vision and language using paired image-text data. This is achieved using contrastive learning in the image-text space. In particular consider a pre-training dataset of size $n$, $\Dcal \subset \Xcal \times \Tcal$, $\Dcal = \{\xv_i, \tv_i\}_{i=1}^{n}$. Here $\Xcal$ and $\Tcal$ are the space of images and text, respectively, and $\xv_i, \tv_i$ are images and text in the dataset. Also, consider access to image and text encoders, that we represent by $f_{\theta}: \Xcal \rightarrow \mathbb{R}^d$ and $f_{\phi}: \Tcal \rightarrow \mathbb{R}^d$ respectively. To learn distributed representations for images and text, the following contrastive losses are used:
{
\begin{align}\label{eq:t2i_clip}
\Lcal_{t2i}	= & - \frac{1}{|\Bcal|} \sum_{j=1}^{|\Bcal|}
\log \frac{ \exp(\tau \uv_i^T \vv_j )  }{\sum_{i=1}^{|\Bcal|}  \exp(\tau \uv_{i}^T \vv_{j} )  }
\end{align}
}%
{
\begin{align}\label{eq:i2t_clip}
\Lcal_{i2t}	= & - \frac{1}{|\Bcal|} \sum_{i=1}^{|\Bcal|}
\log \frac{ \exp(\tau \uv_j^T \vv_j )  }{\sum_{j=1}^{|\Bcal|}  \exp(\tau \uv_{i}^T \vv_{j} )}
\end{align}
}%
Where $\Bcal$ represents the batch during one iteration of training. $\uv_i, \vv_i$ are the $\ell_2$ normalized embeddings of $\Tilde{\uv_i}, \Tilde{\vv_i}$, where $\Tilde{\uv_i} = f_\theta(\xv_{i})$, $\Tilde{\vv_i} = f_\phi(\tv_{i})$. $\tau$ is the temperature parameter and is trainable. The overall loss is $\Lcal_{clip} = (\Lcal_{t2i} + \Lcal_{i2t})/2$.

\section{Scene Graph Decomposition}
\label{sg_decomp}
Here we provide additional details for text scene graph decomposition. Denote the text scene graph obtained from the scene graph parser by $G_T = (V_T, E_T)$, where $V_T$ represent the nodes of the graph, which are either objects or their attributes. $E_T$ are the edges of the graph that represent relations between objects. Let $\mathbb{G}$ denote the set of all possible scene graphs. We first consider an external set of objects ($\mathcal{N}$), attributes ($\mathcal{A}$), and relations ($\mathcal{R}$) that we use for creating negative sub-graphs. In practice, we create this set from Visual Genome (VG) dataset \citep{krishnavisualgenome}. Following \citet{Zhang_2021_CVPR}, we sample a total of 1594 entities that have 30 instances of them in the VG dataset. The attribute and Relation list contains 524, and 50 unique instances, respectively. Hence $|\mathcal{N}| = 1594$, $|\mathcal{A}| = 524$, $|\mathcal{R}| = 50$. We first sample all possible sub-graphs having \textit{one} or \textit{two} objects in them, and these can have multiple attributes for the objects. We develop and use scene graph transformations that take a sub-graph as input and return a (set of) modified versions of the graph (minimally-perturbed negative sub-graphs for the image). For this, we define three graph transformations as follows:
\begin{itemize}
    \item $f_{obj}: \mathbb{G} \longrightarrow P(\mathbb{G})$ takes input a single object scene graph, where the object has attributes $A_o$. For each attribute, $a\in{A_o}$, a random attribute $a'$ is sampled uniformly at random from $\mathcal{A}$. We finally obtain a set of sub-graphs $G_{obj} \in P(\mathbb{G})$ where $P(.)$ denotes the power set. Each $g \in G_{obj}$ contains one object node connected with an attribute node which is sampled from $\mathcal{A}$.
    \item $f_{rel}: \mathbb{G} \longrightarrow P(\mathbb{G})$ takes input sub-graphs having one relation edge and gives output a set of sub-graphs $G_{rel} \in P(\mathbb{G})$ where each $g \in G_{rel}$ has either object nodes shuffled,  replaced by an external object node $n'$ sampled uniformly at random from $\mathcal{N}$, and/or relation replaced by external relation $r'$ sampled uniformly at random from $\mathcal{R}$. Along with this, we also \textit{join} the input positive sub-graph with a random sub-graph created by sampling random nodes and edges from $\mathcal{N}$, $\mathcal{A}$, $\mathcal{R}$.
    \item $f_{attr}: \mathbb{G} \longrightarrow P(\mathbb{G})$ takes input sub-graphs having one relation edge and gives output a set of sub-graphs $G_{attr} \in P(\mathbb{G})$ where each $g \in G_{attr}$ has attribute nodes shuffled, and/or replaced by an external attribute node $a'$ sampled uniformly at random from $\mathcal{A}$.
\end{itemize}
$f_{obj}, f_{attr}$ broadly aims at improving the model's attribute understanding, while $f_{rel}$ broadly targets improved relation understanding.
For each positive sub-graph, we sample all possible negative subgraphs using $f_{obj}, f_{rel}, f_{attr}$ and make positive-negative sub-graph pairs $(g_{pos_i}, \{g_{neg_i}\})$. These pairs can be classified into three categories $C = \{c_{obj}, c_{rel}, c_{attr}\}$ according to the transformation that created the negative sub-graphs. We sample sub-graph pairs from these categories according to probabilities $p_i$, $i\in$ $\{1,2,3\}$ corresponding to the three categories respectively, and $\sum p_i = 1$. These probabilities are hyperparameters; see Appendix Section \ref{hyperparams} for more details. Multiple sub-graph pairs can have common positive or negative sub-graphs, and sampling these pairs would result in duplication, hence for each image, we make sure to deduplicate sub-graphs so that all sub-graphs, and therefore the text made from them are unique for a given image in a batch. After sampling, all sub-graphs are transformed to text using simple templates, as explained in Section \ref{subsec_sg_decomposition}.

\section{Ablations and Model Analysis}
\label{sec_ablations}

\subsection{Sampling more subgraphs}
\label{more_subgraphs_ablation}
\begin{figure}[h!]
    \centering
    \includegraphics[width=\linewidth]{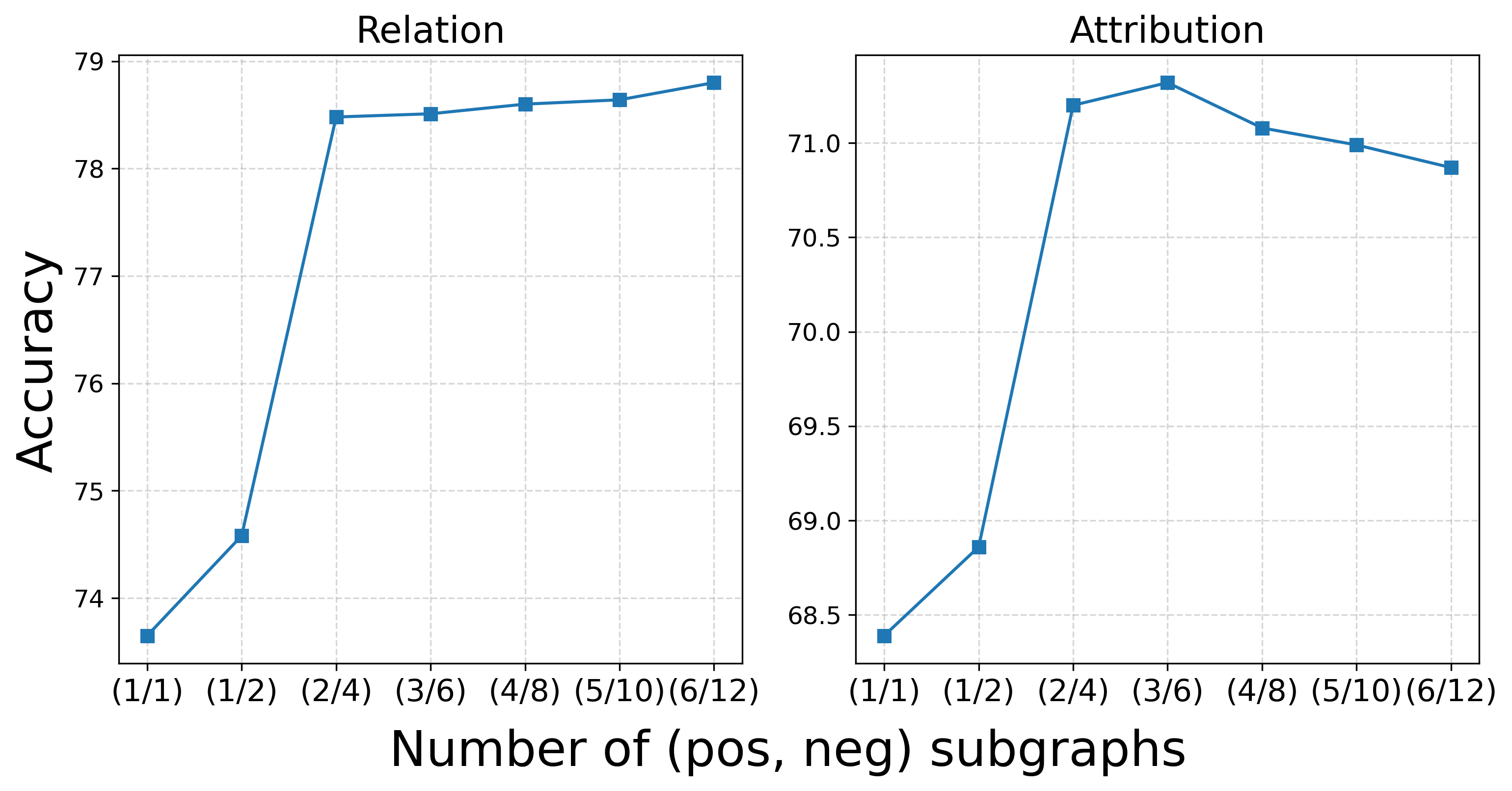}
    \caption{Effect of increasing the number of positive and negative subgraphs on {\color{blue} ARO} benchmark when fine-tuning \methodcomp{}. Results are averaged over 3 fine-tuning datasets considered in this work}
    \label{fig:abalation_num_posneg_aro}
\end{figure}

\begin{figure}[h!]
    \centering
    \includegraphics[width=\linewidth]{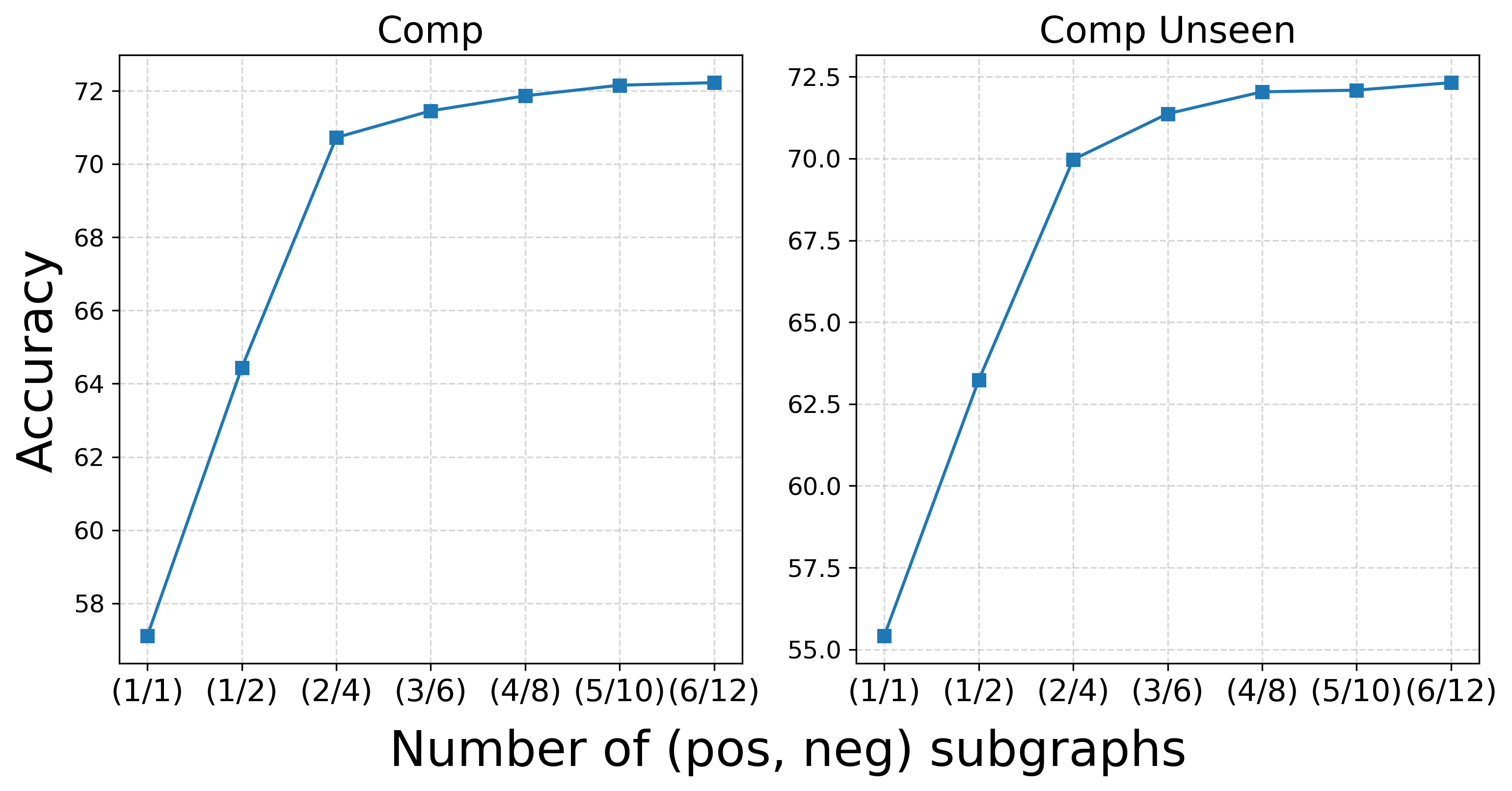}
    \caption{Effect of increasing the number of positive and negative subgraphs on {\color{blue} CREPE - Systematicity} benchmark, when fine-tuning \methodcomp{} (Here we use OpenCLIP RN-50 model pre-trained on CC-12M and fine-tune it on CC-FT).}
    \label{fig:abalation_num_posneg_crepe}
\end{figure}

We analyze the effect of increasing the maximum number of sub-graphs sampled for any given image in a batch of data during training. See Figures \ref{fig:abalation_num_posneg_aro} and \ref{fig:abalation_num_posneg_crepe}, in which we test the performance on ARO and CREPE benchmarks (averaged over three fine-tuning datasets considered in this work), as we increase the max positive and negative sub-graphs per image. We find that as we increase both positive and negative sub-graphs for an image, the performance steadily increases up to a point for \textit{all} datasets, after which the performance can either flatten out, increase, or even decrease in some of the datasets. This is intuitive since a larger number of positive and negative sub-graphs per image leads to a gap w.r.t the pre-training stage as described in Sec. \ref{subsec_curriculum_training}. Also, different compositional splits require different reasoning skills, and as we keep sampling positive and negative sub-graphs for an image, it is natural for certain types of positive and negative sub-graphs to be more pronounced, depending on the dataset statistics, and this can have varied effects on different datasets.

\subsection{Effect of different sub-graph types}
Here we analyze the effect of sampling different kinds of sub-graphs from the original scene graph of the text. In particular, we measure the effect of graph transformations that we define in Appendix Sec. \ref{sg_decomp}. Results are presented in Table \ref{ablations:different_kind_subgraphs}. We observe that both $f_{rel}$ and $f_{attr}$ as described in Appendix Sec. \ref{sg_decomp}, are useful for improving relation and attribute understanding (as measured on the ARO benchmark), across fine-tuning datasets.\\

\subsection{Effect of curriculum training}
\label{curric_ft_effect}
As shown in Table \ref{clip_fine-tune_all_curric_rob_abl}, in all fine-tuning results, we can see consistent improvements when using our curriculum learning strategy, such as upto $2\%$ on systematic generalization, and sometimes more than $6\%$ as seen for ARO-Order results when the fine-tuning dataset is YFCC-FT.

\begin{table*}[h!]
\small
  \fontsize{7.5}{10pt}\selectfont
      \centering
      \setlength{\tabcolsep}{2.5pt}
      {
      \begin{tabular}{lccccc|ccccccc|ccccccc|c}
        \toprule
        \multicolumn{1}{l}{FineTun. data $\rightarrow$} & \multicolumn{5}{c|}{COCO} & \multicolumn{7}{c|}{CC-FT} & \multicolumn{7}{c|}{YFCC-FT} &\\
        \cmidrule(lr){2-6} \cmidrule(lr){7-13} \cmidrule(lr){14-20}
        \multicolumn{1}{l}{Benchmark} $\rightarrow$ & \multicolumn{3}{c}{\textbf{ARO}} & \multicolumn{1}{c}{\textbf{VLC}} & \multicolumn{1}{c|}{\textbf{SVO}} & \multicolumn{3}{c}{\textbf{ARO}} & \multicolumn{2}{c}{\textbf{CREPE}} & \multicolumn{1}{c}{\textbf{VLC}} & \multicolumn{1}{c|}{\textbf{SVO}} & \multicolumn{3}{c}{\textbf{ARO}} & \multicolumn{2}{c}{\textbf{CREPE}} & \multicolumn{1}{c}{\textbf{VLC}} & \multicolumn{1}{c|}{\textbf{SVO}} & Meta\\
        \cmidrule(lr){2-4} \cmidrule(lr){5-5} \cmidrule(lr){6-6} \cmidrule(lr){7-9} \cmidrule(lr){10-11} \cmidrule(lr){12-12} \cmidrule(lr){13-13} \cmidrule(lr){14-16} \cmidrule(lr){17-18} \cmidrule(lr){19-19} \cmidrule(lr){20-20}
        Method $\downarrow$ & Rel. & Attr. & Ord. & Avg. & Avg. & Rel. & Attr. & Ord. & CU & AU & Avg. & Avg. & Rel. & Attr. & Ord. & CU & AU & Avg. & Avg. & Avg.\\
        \midrule
          \clip{}                         & 59.8 & 63.2 & 53.3 & 70.8 & 83.58 & 59.8 & 63.2 & 53.3 & 45.1 & 35.0 & 70.8 & 83.58 & 59.8 & 63.2 & 53.3 & 39.8 & 39.5 & 70.8 & 83.58 & 60.60\\[1pt]
          \negclip{}                      & 81.7 & 72.7 & 85.7 & 75.6 & 90.20 & 71.5 & 65.4 & 84.5 & 53.1 & 37.5 & 72.4 & 88.36 & 57.8 & 63.1 & 52.1 & 38.8 & 39.0 & 70.4 & 83.90 &  67.57\\[1pt]
          \midrule
          \methodcompbold{}            & \textbf{82.6} & \textbf{78.0} & 87.1 & \textbf{81.4} & \textbf{90.67} & \textbf{80.4} & \textbf{69.8} & \textbf{85.5} & \textbf{72.4} & \textbf{40.9} & 77.6 & 88.73 & \textbf{74.3} & 66.9 & \textbf{84.4} & \textbf{48.8} & \textbf{41.5} & \textbf{75.1} & \textbf{85.36} & \textbf{74.29}\\[1pt]
          \methodcompNoCurricbold{}    & 81.6 & 76.8 & \textbf{87.4} & \textbf{81.4} & 90.20 & 79.0 & 69.6 & 80.6 & 71.1 & 40.2 & \textbf{77.7} & \textbf{88.91} & 74.1 & \textbf{67.2} & 77.8 & 46.6 & 40.5 & 75.7 & 84.97 & 73.23\\[1pt]
          \methodcompwiseftbold{}      & 82.5 & 76.2 & 86.6 & 80.3 & 89.65 &  78.8 & 69.4 & 82.6 & 67.5 & 41.2 & 76.4 & 88.08 & 69.4 & 67.0 & 79.4 & 48.1 & 43.6 & 74.2 & 83.71 & 72.88\\[1pt]
        \bottomrule
      \end{tabular}
      }
    
      \caption{Ablating the effect of Curriculum learning and Robust fine-tuning. \methodcompNoCurricbold{} refers to the version of our model without any curriculum learning. \methodcompwiseftbold{} refers to the version where the image encoder of the final model (after fine-tuning) and before fine-tuning are weight-space ensembled. \clip{} and \negclip{} scores are also shown for reference. See Appendix Sec. \ref{curric_ft_effect}. }
      \label{clip_fine-tune_all_curric_rob_abl}
  \end{table*}

\begin{table}[h!]
  \fontsize{10}{10pt}\selectfont
      \centering
      \setlength{\tabcolsep}{2.5pt}
      {
        \begin{tabular}{lcc}
        \toprule
        Method & ZS(21) & Compositional Score\\
        & & (Meta Avg.)\\
        \midrule
        \clip{}   & \textbf{56.4} & 60.60\\
        \negclip{}    & \underline{56.8} & 67.57\\
        \methodcompNoCurric{}     & 55.8 & \underline{73.23}\\
        \methodcompwiseft{}   & \underline{56.8} & 72.88\\
        \methodcomp{}     & 55.7 & \textbf{74.29}\\
        \bottomrule
        \end{tabular}
    }
    \caption{Zero Shot accuracy on 21 multimodal datasets from ELEVATER and ImageNet. Results are average of the three fine-tuning datasets. \methodcomp{} has negligible drop in performance in general (compared to the gains on compositionality benchmarks), and one can boost performance by using \methodcompwiseft{} which has equal performance as compared to \negclip{} on 21 muldimodal datasets. Meta Avg. Compositional Score is taken from Table \ref{clip_fine-tune_all_curric_rob_abl}. Second best results are \underline{underlined}. \textbf{Conclusion:} One can use \methodcomp{} for getting the best compositional reasoning capabilities with minimal performance degradation on multimodal tasks, and use \methodcompwiseft{} for no degradation in performance on multimodal tasks, while still performing well on compositional reasoning.}
    \label{ft_zs_21_table}
\end{table}

\subsection{Effect of robust fine-tuning}
\label{robust_ft_effect}
Among many other techniques developed for mitigating forgetting in large models when they are fine-tuned, one prominent one is robust fin-tuning-WiSE-FT, \citep{wortsman2022robust}. Following \citet{wortsman2022robust} we perform weight-space ensembling on the image encoder before and after fine-tuning using our method and call this model \methodcompwiseft{}. The results on compositionality benchmarks can be seen in Table \ref{clip_fine-tune_all_curric_rob_abl} while results on 21 multimodal tasks from ELEVATER and ImageNet can be seen in Table \ref{ft_zs_21_table}. We find that \methodcompwiseft{} has a slight performance decrease on some compositonal benchmarks as compared to \methodcomp{}, however, it is significantly better than \negclip{} on most benchmarks. The real benefit of using \methodcompwiseft{} is that it leads to least forgetting, and there is little to no performance degradation on 21 tasks as showin in Table \ref{ft_zs_21_table}.

\subsection{Data efficiency}
\label{data_eff}
We find that our technique leads to significant data efficiency requiring about 0.3x-0.6x fo the total fine-tuning or pre-training data to match or exceed \negclip{} performance. Results are shown in Tables \ref{data_eff_pretr} and \ref{data_eff_ft}.
\begin{table}[h!]
\small
\centering
    \begin{tabular}{llcc|c}
    \toprule
    {Method} & Fraction of data& \multicolumn{2}{c|}{ARO} & \multicolumn{1}{c}{SVO} \\
    \cmidrule(lr){3-4} \cmidrule(lr){5-5}
    & & Rel. & Attr. & Avg. \\
    \midrule
    \negclip{} & Full & 73.6 & 58.9 & 76.10 \\
    \midrule
    \multirow{5}{*}{\methodcomp{}} & 0.3x & 71.6 & {\color{blue}60.6} & 70.82 \\
    & 0.5x & {\color{blue}74.3} & 60.8 & 74.04 \\
    & 0.6x & 74.7 & 63.8 & {\color{blue}75.76} \\
    & 0.8x & 77.0 & 66.3 & 77.22 \\
    & Full & 74.7 & 66.1 & 77.87 \\
    \bottomrule
    \end{tabular}

    \caption{Data efficiency of \methodcomp{} during pre-training. Numbers in blue are lowest numbers that are within $1\%$ or greater than \negclip{} performance. Pre-Training dataset: YFCC-15M.}
    \label{data_eff_pretr}
\end{table}

\begin{table}[h!]
\small
\centering
    \begin{tabular}{llcc|c}
    \toprule
    {Method} & Fraction of data& \multicolumn{2}{c|}{ARO} & \multicolumn{1}{c}{SVO} \\
    \cmidrule(lr){3-4} \cmidrule(lr){5-5}
    & & Rel. & Attr. & Avg. \\
    \midrule
    \negclip{} & Full & 71.5 & 65.4 & 88.36 \\
    \midrule
    \multirow{5}{*}{\methodcomp{}} & 0.3x & {\color{blue}70.8} & {\color{blue}67.7} & {\color{blue}88.70} \\
    & 0.5x & 74.5 & 68.6 & 88.80 \\
    & 0.6x & 75.3 & 69.3 & 88.76 \\
    & 0.8x & 78.2 & 69.8 & 88.98 \\
    & Full & 79.0 & 69.6 & 88.91 \\
    \bottomrule
    \end{tabular}

    \caption{Data efficiency of \methodcomp{} during fine-tuning. Numbers in blue are lowest numbers that are within $1\%$ or greater than \negclip{} performance. Fine-tuning dataset: CC-FT. Curriculum learning has not been used for these experiments.}
    \label{data_eff_ft}
\end{table}

\subsection{Computational cost}
\label{computational_cost}
Even though \methodcomp{} uses the same global batch size of image-text pairs, it requires more compute as compared to \negclip{} or \clip{} owing to the fact that decomposing sub-graph leads to a larger effective text-batch size and hence a larger contrastive learning matrix. It is a common practice in literature to trade-off larger compute for improving \clip{}'s compositionality, as also done by previous methods Syn-CLIP \citep{cascantebonilla2023going} that generate data using external graphics engines, and Teaching-SVLC \citep{doveh2023teaching} which use LLMs requiring massive compute even during inference. \\
\textbf{Providing \negclip{} with more compute:} One can argue that providing more compute to \negclip{} can lead to better performance, however, on the contrary we found that \negclip{}'s performance decreases as batch size is scaled (from 256 to 4096, much beyond \methodcomp{}'s text or image batch size), as shown in Table \ref{scale_negclip_batch_size}. \\
\textbf{Performance-Compute Tradeoff:} It is to be noted that \methodcomp{} performance continues to increase up to a threshold, as sub-graphs are increased as shown in Table \ref{fig:abalation_num_posneg_crepe} and \ref{fig:abalation_num_posneg_aro} hence this provides a clean tradeoff between number of sub-graphs and compute, and a practitioner can choose the number of sub-graphs their compute availablility. Along with this, in Appendix Sec. \ref{data_eff} we showed that we can achieve improved performance compared to \negclip{} with as low as 0.3x data closing the gap between \negclip{} and \methodcomp{} compute even more. \textit{It is to be noted that \methodcomp{} is a drop in replacement for \clip{} after training and requires the same inference cost as \clip{}.}

\begin{table}[h!]
\small
\centering
    \begin{tabular}{lccc}
    \toprule
    Batch Size (B) & \multicolumn{2}{c|}{ARO} & \multicolumn{1}{c}{SVO} \\
    \cmidrule(lr){2-3} \cmidrule(lr){4-4}
    & Rel. & Attr. & Avg. \\
    \midrule
    512 & 68.9 & 65.6 & 88.68 \\
    1024 & 67.6 & 65.1 & 88.93 \\
    2048 & 65.7 & 64.2 & 88.72 \\
    4096 & 62.5 & 63.7 & 88.11 \\
    \bottomrule
    \end{tabular}

    \caption{Performance of \negclip{} with increasing batch size. A batch size of B corresponds to an effective batch size of 8*B in \negclip{} after image and text negative mining. Fine-tuning dataset: CC-FT.}
    \label{scale_negclip_batch_size}
\end{table}

\section{Additional Results and Experiments}
\label{additional_expt_results}

\subsection{Comparison with recent baselines}
\label{comparison_other_baselines}
We compare with recently published and contemporary works \citep{cascantebonilla2023going, doveh2023teaching}. \citet{doveh2023teaching} show that one can create rule-based hard negative sentences and Large Language Models (LLMs) based hard negative sentences and use them when training CLIP style models to obtain an improved model that is better at handling tasks that require compositional reasoning.
We fine-tune on CC3M \citep{sharma-etal-2018-conceptual} for a fair comparison with \citet{doveh2023teaching}. Results are reported in Table \ref{tab:comparison_other_baselines}. A fair comparison with Syn-CLIP \citet{cascantebonilla2023going} is not possible since their synthetic dataset is not released. However in Table \ref{tab:comparison_other_baselines} we find that performance difference is large between \methodcomp{} and Syn-CLIP showing that our general coarse-to-fine grained approach is better than using targeted synthetic datasets for inducing compositional understanding in VLMs. Comparisons with \citet{doveh2023teaching} in Table show that our approach is competitve or better at attribute, relation and object understanding as measured by the VL-Checklist benchmark \citep{zhao2022vlchecklist}. Zero Shot performance on 21 datasets suffers minimally using our approach, and is even better than \citep{zhao2022vlchecklist}. It is to be noted that both approaches Syn-CLIP \citep{cascantebonilla2023going} and \citet{doveh2023teaching} are orthogonal to our approach and combining them with our coarse-to-fine understanding approach will likely result in much better performance overall, as compared to individual techniques. In particular, Syn-CLIP \citep{cascantebonilla2023going} faces the issue of having long captions for images, and they average out embeddings of parts of the caption before matching it to the image. This issue can be eaily resolved using our framework which can easily handle multiple positive captions for an image. Performing this ablation would be future work for us, once synthetic datasets like that used by \citet{cascantebonilla2023going} are open-sourced and gain more popularity. Our approach can similarly also include captions generated from LLMs, as explored by \citet{doveh2023teaching}.

\begin{table*}[h]
  \centering
  \begin{tabular}{l@{\hspace{0.75em}}ccc|ccc|c}
    \toprule
    Benchmark $\rightarrow$ & \multicolumn{3}{c|}{\textbf{VL-Checklist}} & \multicolumn{3}{c|}{\textbf{ARO}} & \multicolumn{1}{c}{\textbf{ZS(21)}} \\
    \cmidrule(lr){2-4} \cmidrule(lr){5-7} \cmidrule(lr){8-8}
    \multirow{1}{*}{Method} & Obj. & Attr. & Rel. & Rel. & Attr. & Ord. & Avg. \\
    \midrule
    \clip{} & 81.6	& 67.6	& 63.1	& 	59.9 & 63.6 & 53.3 & \textbf{56.4} \\[1pt]
    \clip{}-FT & 79.0 & 64.7 & 54.3 & 41.7 & 59.3 & 25.2 & 56.9 \\[1pt]
    Syn-CLIP\textsuperscript{$\dagger$} {\tiny \citep{cascantebonilla2023going}} & - - & 70.4 & 69.4 & 71.4 & 66.9 & 65.1 & 55.3 \\[1pt]
    Teaching SVLC\textsuperscript{$\ddagger$} {\tiny \citep{doveh2023teaching}} & 85.0 & 72.0 & 69.0 & - - & - - & - - & 54.8 \\[1pt]
    \midrule
    \rowcolor{cyan!12}
    \methodcompNoCurricbold{} & 86.4 &  \textbf{75.0} & 69.6  & 83.2 & \textbf{78.6} & 77.3 & 54.9 \\[1pt]
    \rowcolor{cyan!12}
    \methodcompwiseftbold{}    & \textbf{86.5} &  73.6 & \textbf{72.2}  &  82.6 & 77.0 & \textbf{79.9} & \underline{55.9} \\[1pt]
    \rowcolor{cyan!12}
    \methodcompbold{} & 86.4  &   73.7 & 71.9   & \textbf{83.7} & 78.0 & 79.4 &  53.5 \\
    \bottomrule
  \end{tabular}

  \caption{Comparison of \methodcomp{} with recently published and contemporary works Syn-CLIP \citep{cascantebonilla2023going} and Teaching SVLC \citet{doveh2023teaching}. Results are reported on {\color{blue} VL-Checklist}, {\color{blue} ARO} and Average Zero Shot results on 21 datasets from {\color{blue} ELEVATER} and {\color{blue} Imagenet}. Performance numbers of these models are reported from their respective papers (blank fields (---) are not reported in respective papers). \textsuperscript{$\dagger$}Uses million-scale synthetic data for fine-tuning. \textsuperscript{$\ddagger$}Uses external Large Language Models (LLMs) like BLOOM \citep{bloom} for text augmentation and hard negative text creation. See Sec. \ref{comparison_other_baselines} for more details.}
  \label{tab:comparison_other_baselines}
\end{table*}

\subsection{Standard deviations for fine-tuning results}
\label{std_dev_results}
Here we provide fine-tuning results on the CC-FT dataset \textit{with standard deviations} over 3 random seeds where OpenAI CLIP-ViT-B-32 is fine-tuned on CC-FT using \methodcomp{} and baseline techniques. See Table \ref{tab:std_dev_cc100k} for the results. The main paper Table \ref{clip_fine-tune_all} have average results for CC-FT while for COCO and YFCC-FT fine-tuning datasets, the results are for one seed.
We do-not run multiple pre-training experiments since they significantly more costly.
\begin{table*}[h!]
\small
\centering
    \begin{tabular}{lccc|ccc}
        \toprule
        Benchmark $\rightarrow$ & \multicolumn{3}{c|}{\textbf{ARO}} & \multicolumn{3}{c}{\textbf{SVO-Probes}} \\
        \cmidrule(lr){2-4} \cmidrule(lr){5-7}
        Method $\downarrow$ & Rel. & Attr. &  Ord. & Obj. & Subj. & Verb.\\
        \midrule
        \clip{}-FT & 58.1\scriptsize$\pm$0.63 & 63.3\scriptsize$\pm$0.28 & 42.7\scriptsize$\pm$0.18 & 93.17\scriptsize$\pm$0.11 & 88.64\scriptsize$\pm$0.17 & 83.87\scriptsize$\pm$0.03 \\
        \negclip{} & 71.5\scriptsize$\pm$0.40 & 65.4\scriptsize$\pm$0.58 & 84.5\scriptsize$\pm$0.11 & 92.90\scriptsize$\pm$0.09 & 88.16\scriptsize$\pm$0.11 & 84.02\scriptsize$\pm$0.02 \\
        \midrule
        \rowcolor{cyan!12}
        \methodcompNoCurricbold{} & 79.0\scriptsize$\pm$0.66 & 69.6\scriptsize$\pm$0.19 & 80.6\scriptsize$\pm$0.17 & 93.37\scriptsize$\pm$0.04 & 89.74\scriptsize$\pm$0.13 & 83.62\scriptsize$\pm$0.04 \\
        \rowcolor{cyan!12}
        \methodcompbold{} & 80.4\scriptsize$\pm$0.63 & 69.8\scriptsize$\pm$0.21 & 85.5\scriptsize$\pm$0.16 & 93.45\scriptsize$\pm$0.04 & 89.39\scriptsize$\pm$0.07 & 83.35\scriptsize$\pm$0.05 \\
        \bottomrule
    \end{tabular}

  \caption{Fine-Tuning Results on CC-FT dataset \textit{with standard deviations} across 3 random seeds. These results correspond to the CC-FT fine-tuning results in main paper Table \ref{clip_fine-tune_all}. Here the base model which is fine-tuned using different techniques is OpenAI-CLIP-ViT-B-32.}
  \label{tab:std_dev_cc100k}
\end{table*}

\subsection{Robustness to natural distribution shifts}
\label{robustness_detailed_results}
We find that pre-trained \methodcomp{} shows robustness to natural distribution shifts as measured by ImageNet natural distribution shifts benchmark. Results are presented in Table \ref{tab:rob_nat_shifts}. We believe that \methodcomp{} sees a larger variety of texts in the form of sub-graphs which can provide it with extra supervision for tackling natural distribution shifts. Intutively, sub-graphs can lead to diversity of texts being seen by the model during training and this might lead to broader coverage of concepts and concept combinations, resulting in improved robustness. Along with this a coarse to fine 
 hierarchical understanding of texts and thereby, of images should intuitively help in improving performance on robustness benchmarks given that the model will now be able to recognise details in images and texts more accuractely.

\begin{table*}[h!]
\small
  \centering
  \begin{tabular}{lllcccccccc}
      \toprule
      \multicolumn{3}{l}{} & \multicolumn{2}{c}{\textbf{ImageNet-A}} & \multicolumn{2}{c}{\textbf{ImageNet-R}} & \multicolumn{2}{c}{\textbf{ImageNet-S}} & \multicolumn{2}{c}{\textbf{ImageNet-V2}} \\
      \cmidrule(lr){4-5} \cmidrule(lr){6-7} \cmidrule(lr){8-9} \cmidrule(lr){10-11}
      {Arch.} & {Data} & {Method} & Top1 & Top5 & Top1 & Top5 & Top1 & Top5 & Top1 & Top5 \\
      \midrule
      & & \clip{} & 6.4 & 24.5 & 42.6 & 68.8 & 22.2 & 45.5 & 28.2 & 54.1 \\
      & & \negclip{} & 6.6 & 25.0 & 43.1 & 68.7 & 22.2 & 45.4 & 29.4 & 55.2 \\
      \rowcolor{cyan!12}
      \cellcolor{white} & \cellcolor{white} \multirow{-3}{*}{{CC-12M}} & \methodcompbold{} & \textbf{9.1} & \textbf{29.4} & \textbf{48.6} & \textbf{74.3} & \textbf{27.2} & \textbf{52.6} & \textbf{33.6} & \textbf{61.6} \\
      \cmidrule{2-11}
      & & \clip{} & 10.9 & 34.2 & 20.6 & 42.0 & 6.4 & 16.7 & 26.1 & 49.9 \\
      & & \negclip{} & 11.4 & 35.6 & 20.0 & 41.7 & 6.0 & 16.0 & 27.2 & 50.7 \\
      \rowcolor{cyan!12}
      \cellcolor{white} \multirow{-6}{*}{\rotatebox[origin=c]{90}{Swin-T}} & \cellcolor{white} \multirow{-3}{*}{{YFCC-15M}} & \methodcompbold{} & \textbf{14.6} & \textbf{40.2} & \textbf{22.3} & \textbf{44.9} & \textbf{6.8} & \textbf{17.7} & \textbf{32.0} & \textbf{57.2} \\
      \midrule
      & & \clip{} & 7.3 & 27.4 & 41.4 & 67.8 & 21.7 & 44.3 & 29.8 & 56.4 \\
      & & \negclip{} & 7.7 & 27.7 & 41.0 & 66.9 & 21.7 & 43.9 & 30.2 & 56.0 \\
      \rowcolor{cyan!12}
      \cellcolor{white} & \cellcolor{white} \multirow{-3}{*}{{CC-12M}} & \methodcompbold{} & \textbf{11.1} & \textbf{35.6} & \textbf{52.1} & \textbf{76.9} & \textbf{29.5} & \textbf{55.4} & \textbf{37.0} & \textbf{66.5} \\
      \cmidrule{2-11}
      & & \clip{} & 13.4 & 37.3 & 17.2 & 37.2 & 4.9 & 13.6 & 25.8 & 49.4 \\
      & & \negclip{} & 12.9 & 38.0 & 18.0 & 37.3 & 5.1 & 14.7 & 26.0 & 49.0 \\
      \rowcolor{cyan!12}
      \cellcolor{white} \multirow{-6}{*}{\rotatebox[origin=c]{90}{RN-50}} & \cellcolor{white} \multirow{-3}{*}{{YFCC-15M}} & \methodcompbold{} & \textbf{17.4} & \textbf{46.6} & \textbf{21.0} & \textbf{42.7} & \textbf{6.5} & \textbf{16.9} & \textbf{32.2} & \textbf{57.9} \\
      \bottomrule
  \end{tabular}

  \caption{Results on {\color{blue} ImageNet - Natural Distribution Shifts datasets}. \methodcompbold{} leads to improved robustness to natural distribution shifts. \negclip{} performs similarly as \clip{}. Models are zero-shot tested on ImageNet-A \citep{Hendrycks_2021_CVPR}, ImageNet-R \citep{hendrycks2021many}, ImageNet-S(ketch) \citep{wang2019learning} and ImageNet-V2 \citep{recht2019imagenet}.}
  \label{tab:rob_nat_shifts}
\end{table*}

\section{Dataset Details}
\label{data_details}
\begin{table*}[h!]
  \centering
  \begin{tabular}{lcc@{\hspace{0.75em}}ll}
    \toprule
    \textbf{Benchmark/Dataset} & \textbf{\#Examples} & \textbf{\#Subtasks} & \textbf{Subtask Examples} & \textbf{Datasets Used}\\
    & & & & \textbf{for Creation}\\
    \midrule
    \multicolumn{5}{c}{Compositional Reasoning (Evaluation)}\\
    \midrule
    \rowcolor{gray!12}
    ARO & 77K & 3 & Attribute, Relation, & Visual Genome,\\
    \rowcolor{gray!12}
    & & & Order understanding & COCO, Flickr \\
    CREPE-Systematicity & 642K & 2 & Systematic generalization & \\
    & & & generalization & Visual Genome\\
    \rowcolor{gray!12}
    VL-Checklist & 410K & 3 & Attribute, Relation & Visual Genome\\
    \rowcolor{gray!12}
    & & & Object understanding & HAKE, VAW, SWiG\\
    SVO-Probes & 48K & 3 & Verbs (Relations) & --\\
    & & & understanding & \\
    \rowcolor{gray!12}
    CREPE-Productivity & 183K & 9 & Productivity & Visual Genome\\
    \midrule
    \multicolumn{5}{c}{Fine-Tuning datasets} \\
    \midrule
    \rowcolor{gray!12}
    COCO & 109K & -- & -- & --\\
    CC-FT & 100K & -- & -- & --\\
    \rowcolor{gray!12}
    YFCC-FT & 100K & -- & -- & --\\
    CC-3M & 3.11M & -- & -- & --\\
    \midrule
    \multicolumn{5}{c}{Pre-Training datasets} \\
    \midrule
    \rowcolor{gray!12}
    CC-12M & 11.26M & -- & -- & --\\
    YFCC-15M & 14.20M & -- & -- & --\\
    \bottomrule
  \end{tabular}

  Citations: ARO\citep{yuksekgonul2022and}, CREPE\citep{ma2022crepe}, VL-Checklist\citep{zhao2022vlchecklist}, SVO\citep{hendricks-nematzadeh-2021-probing}, Visual Genome\citep{krishnavisualgenome}, COCO\citep{lin2014microsoft}, Flickr\citep{flickr}, HAKE\citep{hake}, VAW\citep{vaw}, SWiG\citep{swig} 
  \caption{Details of datasets used in this study for testing compositional reasoning, for fine-tuning and pre-training models. See Appendix Sec. \ref{data_details} for more details.}
  \label{data_details_table}
\end{table*}
Here we provide detailes about datasets used for fine-tuning, pre-training and evaluating models in this study. A summary is shown in Table \ref{data_details_table}
\subsection{Fine-tuning datasets}
Following \negclip{} \citep{yuksekgonul2022and} we use the COCO dataset released by \citep{yuksekgonul2022and} having 109k samples that had hard negative sentences that \citep{yuksekgonul2022and} create for training NegCLIP. As mentioned in the main paper, COCO dataset images are used for creating Visual Genome \citep{krishnavisualgenome}, and this is further used to create datasets such as CREPE \citep{ma2022crepe}, ARO \citep{yuksekgonul2022and} and a part of VL-Checklist \citep{zhao2022vlchecklist}. This can lead to confounding and potentially misleading results, since it is unclear if the performance increase using any method comes from the fine-tuning dataset (COCO) being close to the domain of test datasets, or if it's the fine-tuning methodology that leads to an increase in performance. Hence, for rigourous experimentation of the developed methods, one must use other datasets to fine-tune contrastively trained VLMs. We randomly sample similar sized (100k datapoints) from popular pre-training datasets CC-12M and YFCC-15M, and call these smaller datasets CC-FT and YFCC-FT. To train \negclip{}, hard negative sentences and images are required, for which we first use the code released by \citep{yuksekgonul2022and}\footnote{\url{https://github.com/mertyg/vision-language-models-are-bows}} to create hard negatives sentences as well as sample three hard negative images for each image based on OpenAI CLIP ViT-B/32 features, strictly following \citep{yuksekgonul2022and}.
For comparing with contemporary works \citep{doveh2023teaching}, \citep{cascantebonilla2023going} (as shown in Table \ref{tab:comparison_other_baselines_main}), we use CC3M \citep{sharma-etal-2018-conceptual} since it's used by these baselines, and makes a direct comparison possible with them.
\subsection{Pre-training datasets}
We use popular and standard large scale pre-training datasets CC-12M \citep{changpinyo2021cc12m} and YFCC-15M \citep{thomee2016yfcc100m} for pre-training all models in this study, including \clip{}, \negclip{} and \methodcomp{}.
\subsection{Evaluation datasets}
\label{eval_data_details}
Here we list the evaluation detailes used in this study and also provide a short description for each
\noindent \textbf{CREPE-Systematicity} \cite{ma2022crepe}: CREPE provides systematic generalization datasets to test models trained on popular pre-training datasets including CC-12M and YFCC-15M. While creating CREPE, \citet{ma2022crepe} make sure to split the dataset into seen and unseen parts, which correspond to weather the model has seen or not seen the combination of concepts, when pre-trained with popular pre-training datasets. We measure and report performance on both seen and unseen splits in our work.\\
\noindent \textbf{ARO} \cite{yuksekgonul2022and}: This benchmark consists of four datasets, including VG-Relation, VG-Attribution, COCO-Order, and Flickr-Order. The first two measure attribute and relation understanding of VL models, respectively, and the last two measure the word order understanding of VL models. VG-Relation and VG-Attribution consist of tuples having an image and two texts (one positive and one negative), and the model's task is to match the image with the correct text. order datasets have four negative texts and one positive text for each image, and the task is again to match the image with the correct text.\\
\noindent \textbf{SVO-Probes} \cite{hendricks-nematzadeh-2021-probing}: This dataset consists of tuples having two images and one text. All texts and images have a subject, verb, and object, and the images differ in only one of subject, verb, or object. This dataset helps in understanding if VL models can compositionally understand combinations of objects having a relation between them. The original dataset contains 48K examples.\footnote{Some image links provided by the the original repository(\url{https://github.com/deepmind/svo_probes}) were broken. In total, 36k data points were retrievd and used in this study.}\\ 
\noindent \textbf{CREPE-Productivity} \cite{ma2022crepe}: Productivity dataset tests the model's ability to generalize to longer and more complex sentences, with complexity ranging from 4 atoms to 12 atoms, where an atom can be an attribute, relation, or object. The CREPE-Productivity dataset has a number of test sets for each sentence complexity ranging from 4 atoms to 12 atoms.\\
\noindent \textbf{VL-Checklist} \cite{zhao2022vlchecklist}: This benchmark is created by combining annotations from datasets like Visual Genome \citep{krishnavisualgenome}, SWiG \citep{swig}, HAKE \citep{hake}, VAW \citep{vaw}. Each image in the resulting dataset has two captions, a positive and a negative. The positive caption is taken from the source dataset of the image, while the negative caption differs from the positive in only one word which makes it a hard negative and helps in testing compositional and fine-grained understanding of VLMs across various dimensions like attributes, relations, and size and locations of objects.\\

\section{Baselines:}
\label{baselines}
Here we list the baselines used in this study and also provide a short description for each.
\noindent \textbf{CLIP}\citep{radford2021learning}: Our first baseline is CLIP model released by OpenAI CLIP\citep{radford2021learning} and OpenCLIP \citep{Ilharco_OpenCLIP_2021}. In particular we use the ViT-B/32 model for fine-tuning results Table \ref{clip_fine-tune_all} of the main paper, except for CREPE dataset, which requires using models pre-traoined on specific datasets, for which we use ResNet-50 (RN-50) models pre-trained on CC-12M and YFCC-15M released by OpenCLIP repository\footnote{\url{https://github.com/mlfoundations/open_clip}} \citep{Ilharco_OpenCLIP_2021}.\\
\noindent \textbf{CLIP-FT}: For disentangling the effects of fine-tuning data, and fine-tuning methodology, we create a CLIP-FT baseline where we simply fine-tune the pre-trained CLIP model on the dataset at hand, by using the standard contrastive learning technique used by CLIP. \\
\noindent \textbf{NegCLIP}\citep{yuksekgonul2022and} [ICLR 2023]: NegCLIP is trained using negative mining of texts and images. \citet{yuksekgonul2022and} create sentence level hard negatives by swapping different linguistic elements. They also additionally include hard-negative images and their corresponding texts in the batch by fetching K nearest neighbours (K=3) for each image in the feature space constructed using a pretrained CLIP model.\\
\noindent \textbf{Teaching SVLC}\citep{doveh2023teaching} [CVPR 2023]: This method uses LLM's like BLOOM \citep{bloom} along with rules to create additional positive and negative sentences for each image while fine-tuning CLIP.\\
\noindent \textbf{Syn-CLIP}\citep{cascantebonilla2023going} [Arxiv 2023]: Syn-CLIP uses a million scale synthetic dataset to fine-tune CLIP and improve it's performance on compositional reasoning tasks. The synthetic data is created using a 3D physics-based simulation platform built on Unity3D, called ThreeDWorld \citep{gan2021threedworld}. This contemporary work is complementary to our data-centric approach and we believe our methods can help fine-tuning with synthetic datasets as well. \citet{cascantebonilla2023going} in their paper showed how dense and long captions can be obtained for synthetic images and which require splitting into sub-captons followed by averaging of features from all captions while fine-tuning CLIP. This is one avenue where we believe our method can be useful since our method inherently allows matching of images to multiple texts. This is part of future work, once such synthetic datasets are released and are easily available. \\

\section{Detailed Experimental Results}
\label{detailed_expt_results}
In the main paper Table \ref{clip_fine-tune_all} and Table \ref{pre-training_results_all} we had provided concise results for some datasets, based on lack of space due to extensive experimental results. Here we provide detailed results on these datasets:
\subsection{VL-Checklist: detailed results}
\label{vl_checklist_results}
Detailed Fine-tuning results on VL-Checklist dataset are provided in Table \ref{vl_checklist_results_finetune}. These are an extension to the VL-Checklist results provided in the main paper Table \ref{clip_fine-tune_all}. Detailed Pre-training results for VL-Checklist dataset are provided in Table \ref{vl_checklist_results_pretrain} which are an extension to the VL-Checklist results provided in the main paper Table \ref{pre-training_results_all}.

  \begin{table*}[h!]
      \centering
      \begin{tabular}{lr@{\hspace{0.75em}}r@{\hspace{0.75em}}r@{\hspace{0.75em}}|r@{\hspace{0.75em}}r@{\hspace{0.75em}}r@{\hspace{0.75em}}|r@{\hspace{0.75em}}r@{\hspace{0.75em}}r@{\hspace{0.75em}}}
        \toprule
        Benchmark $\rightarrow$ & \multicolumn{9}{c}{\textbf{VL-Checklist}}\\
        \cmidrule(lr){2-10}
        Fine-tuning data $\rightarrow$ & \multicolumn{3}{c|}{CC-100K} & \multicolumn{3}{c|}{YFCC-100K} & \multicolumn{3}{c}{COCO}\\
        \cmidrule(lr){2-4} \cmidrule(lr){5-7} \cmidrule(lr){8-10}
        Method & Obj. & Attr. & Rel.  & Obj. & Attr. & Rel.  & Obj. & Attr. & Rel. \\
        \midrule
          \clip{}                         &  81.6	& 67.6	& 63.1  &  81.6	& 67.6	& 63.1  &  81.6	& 67.6	& 63.1  \\[1pt]
          \clip{}-FT                      & 81.9 &  69.3 & 60.9  & 80.7 &  68.1 & 60.2  & 83.7 &  66.7 & 59.3  \\[1pt]
          \negclip{}                      & 82.1 &  71.4 & 70.3   & 81.0 &  68.1 & 67.1  & 85.2 &  67.2 & 63.0  \\[1pt]
          \midrule
          \rowcolor{cyan!12}
          \methodcompNoCurricbold{}     & \textbf{86.0} &  \textbf{77.2} & 77.7  & 84.0 &  \textbf{72.2} & \textbf{75.1}  & \textbf{89.2} &  70.4 & 72.6  \\[1pt]
          \rowcolor{cyan!12}
          \methodcompwiseftbold{}     & 85.3 &  71.4 & 72.4  & 83.6  &  69.5 & 69.6  & 88.5 &   \textbf{75.5} & \textbf{77.0}  \\[1pt]
          \rowcolor{cyan!12}
          \methodcompbold{}             & \textbf{86.0} &  76.8 & \textbf{78.4}  &  \textbf{84.1} &  72.1 & 74.8  & 89.0 &   70.1 & 71.3  \\[1pt]
        \bottomrule
      \end{tabular}
    
      \caption{Fine-tuning results on the {\color{blue} VL-Checklist} benchmark, for testing compositionality in terms of attribute, relation and object understanding. OpenAI CLIP VIT-B-32 pre-trained model is used as the base model for fine-tuning. See Sec. \ref{vl_checklist_results} for more details.}
      \label{vl_checklist_results_finetune}
  \end{table*}

  \begin{table}[h!]
  \fontsize{8.}{10pt}\selectfont
      \centering
      \begin{tabular}{p{0.4cm}lp{0.4cm}p{0.4cm}p{0.4cm}|p{0.4cm}p{0.4cm}p{0.4cm}}
          \toprule
          \multicolumn{2}{l}{Benchmark $\rightarrow$} & \multicolumn{6}{c}{\textbf{VL-Checklist}}\\
          \cmidrule(lr){3-8}
          \multicolumn{2}{l}{Pre-training data $\rightarrow$} & \multicolumn{3}{c|}{CC-12M} & \multicolumn{3}{c}{YFCC-15M}\\
          \cmidrule(lr){3-5} \cmidrule(lr){6-8}
          {Arch.} & Method & Obj. & Attr. & Rel.  & Obj. & Attr. & Rel.  \\
          \midrule
          & \clip{} & 75.2  &  61.1 & 60.6  & 73.6 &   63.0 & 62.0  \\[1pt]
          & \negclip{} & 75.0 &  67.7 & \textbf{67.4}  & 71.2 &   66.5 & 60.3  \\[1pt]
          \rowcolor{cyan!12}
          \cellcolor{white} \multirow{-3}{*}{\rotatebox[origin=c]{90}{Swin-T}} & \methodcompbold{} & \textbf{80.0}  &  \textbf{72.9} & 64.4  & \textbf{79.3} &   \textbf{71.3} & \textbf{64.8}  \\[1pt]
          \midrule
          & \clip{} & 75.5 & 62.7 & 60.5  & 73.2 &   62.8 & 58.3  \\[1pt]
          & \negclip{} & 75.4 & 67.6 & \textbf{65.5}  & 72.9 &   65.8 & 59.7  \\[1pt]
          \rowcolor{cyan!12}
          \cellcolor{white} \multirow{-3}{*}{\rotatebox[origin=c]{90}{RN-50}} & \methodcompbold{} & \textbf{79.2} & \textbf{73.2} & \textbf{65.3}  & \textbf{80.1} &   \textbf{71.6} & \textbf{65.1}  \\[1pt]
          \bottomrule
      \end{tabular}
    
      \caption{Pre-training results on {\color{blue} VL-Checklist} benchmark, for testing compositionality in terms of attribute, relation and object understanding. Results for both backbones Swin-Tiny and RN-50 are shown. See Sec. \ref{vl_checklist_results} for more details.}
      \label{vl_checklist_results_pretrain}
  \end{table}

\subsection{SVO-Probes: detailed results}
\label{svo_detailed_results}
Detailed Fine-tuning results on SVO-Probes dataset are provided in Table \ref{detailed_svo_fine_tune}. These are an extension to the SVO-Probes results provided in the main paper Table \ref{clip_fine-tune_all}. Detailed Pre-training results for SVO-Probes dataset are provided in Table \ref{detailed_svo_pre_train} which are an extension to the SVO-Probes results provided in the main paper Table \ref{pre-training_results_all}.

\begin{table}[h!]
\footnotesize
  \centering
  \begin{tabular}{p{0.15cm}p{2.2cm}p{0.7cm}p{0.7cm}p{0.7cm}p{0.7cm}}
      \toprule
      & \multicolumn{1}{l}{Benchmark $\rightarrow$} & \multicolumn{4}{c}{\textbf{SVO-Probes}} \\
      \cmidrule(lr){3-6}
      & Method & Obj & Subj & Verb & Avg \\ 
      \cmidrule{2-6}
       & \clip{} & 88.13 & 83.85 & 78.76 & 83.58 \\ 
       \midrule
       & \clip{}-FT & 93.17 & 88.64 & 83.87 & 88.56 \\ 
       & \negclip{} & 92.90 & 88.16 & \textbf{84.02} & 88.36 \\ 
       \rowcolor{cyan!12}
       \cellcolor{white}
       & \methodcompNoCurricbold{} & 93.37 & \textbf{89.74} & 83.62 & \textbf{88.91} \\ 
       \rowcolor{cyan!12}
       \cellcolor{white}
       & \methodcompwiseftbold{} & 92.65 & 88.69 & 82.90 & 88.08 \\ 
       \rowcolor{cyan!12}
       \cellcolor{white} \multirow{-5}{*}{\rotatebox[origin=c]{90}{CC-100K}}
       & \methodcompbold{} & \textbf{93.45} & 89.39 & 83.35 & 88.73 \\ 
      \midrule
       & \clip{}-FT & 89.63 & 85.83 & \textbf{80.36} & 85.27 \\ 
       & \negclip{} & 88.43 & 84.05 & 79.21 & 83.90 \\ 
       \rowcolor{cyan!12}
       \cellcolor{white}
       & \methodcompNoCurricbold{} & 89.49 & 85.59 & 79.83 & 84.97 \\ 
       \rowcolor{cyan!12}
       \cellcolor{white}
       & \methodcompwiseftbold{} & 87.86 & 84.97 & 78.30 & 83.71 \\
       \rowcolor{cyan!12}
       \cellcolor{white} \multirow{-5}{*}{\rotatebox[origin=c]{90}{YFCC-100K}}
       & \methodcompbold{} & \textbf{89.93} & \textbf{86.45} & 79.71 & \textbf{85.36} \\
      \midrule
       & \clip{}-FT & 93.60 & 91.37 & 85.48 & 90.15 \\ 
       & \negclip{} & 93.59 & 91.43 & \textbf{85.58} & 90.20 \\ 
       \rowcolor{cyan!12}
       \cellcolor{white}
       & \methodcompNoCurricbold{} & 94.14 & 92.22 & 84.23 & 90.20 \\ 
       \rowcolor{cyan!12}
       \cellcolor{white}
       & \methodcompwiseftbold{} & 93.13 & 92.07 & 83.75 & 89.65 \\
       \rowcolor{cyan!12}
       \cellcolor{white} \multirow{-5}{*}{\rotatebox[origin=c]{90}{COCO}}
       & \methodcompbold{} & \textbf{94.16} & \textbf{93.04} & 84.82 & \textbf{90.67} \\
      \bottomrule
  \end{tabular}

  \caption{Detailed Fine-tuning results on the {\color{blue} SVO-Probes} dataset. See Sec. \ref{svo_detailed_results} for more details.}
  \label{detailed_svo_fine_tune}
\end{table}

\begin{table}[h!]
\small
  \centering
  \begin{tabular}{l@{\hspace{0.8em}}l@{\hspace{0.8em}}l@{\hspace{0.8em}}c@{\hspace{0.8em}}c@{\hspace{0.8em}}c@{\hspace{0.8em}}c@{\hspace{0.8em}}}
      \toprule
      \multicolumn{3}{l}{Benchmark $\rightarrow$} & \multicolumn{4}{c}{\textbf{SVO-Probes}} \\
      \cmidrule{4-7}
      {Arch.} & {Data} & {Method} & Obj & Subj & Verb & Avg \\ 
      \midrule  
      & & \clip{} & 88.43 & 82.58 & 79.33 & 82.21 \\ 
      & & \negclip{} & 88.38 & 81.83 & 79.40 & 82.04 \\ 
      \rowcolor{cyan!12}
      \cellcolor{white} & \cellcolor{white} \multirow{-3}{*}{\rotatebox[origin=c]{90}{\tiny CC-12M}} & \methodcompbold{} & \textbf{91.89} & \textbf{87.11} & \textbf{82.20} & \textbf{85.62} \\
      \cmidrule{2-7}
      & & \clip{} & 83.38 & 77.09 & 72.80 & 76.27 \\ 
      & & \negclip{} & 84.07 & 76.87 & 72.28 & 76.10 \\ 
      \rowcolor{cyan!12}
      \cellcolor{white} \multirow{-6}{*}{\rotatebox[origin=c]{90}{Swin-T}} & \cellcolor{white} \multirow{-3}{*}{\rotatebox[origin=c]{90}{\tiny YFCC-15M}} & \methodcompbold{} & \textbf{86.20} & \textbf{79.24} & \textbf{73.61} & \textbf{77.87} \\ 
      \midrule
      & & \clip{} & 87.86 & 82.54 & 79.45 & 82.13 \\ 
      & & \negclip{} & 87.58 & 82.47 & 79.42 & 82.03 \\ 
      \rowcolor{cyan!12}
      \cellcolor{white} & \cellcolor{white} \multirow{-3}{*}{\rotatebox[origin=c]{90}{\tiny CC-12M}} & \methodcompbold{} & \textbf{90.18} & \textbf{85.22} & \textbf{80.48} & \textbf{83.86} \\ 
      \cmidrule{2-7}
      & & \clip{} & 82.61 & 76.21 & 72.27 & 75.60 \\ 
      & & \negclip{} & 81.40 & 76.05 & 72.06 & 75.18 \\ 
      \rowcolor{cyan!12}
      \cellcolor{white} \multirow{-6}{*}{\rotatebox[origin=c]{90}{RN-50}} & \cellcolor{white} \multirow{-3}{*}{\rotatebox[origin=c]{90}{\tiny YFCC-15M}} & \methodcompbold{} & \textbf{84.25} & \textbf{79.83} & \textbf{73.29} & \textbf{77.42} \\
      \bottomrule
  \end{tabular}

  \caption{Detailed Pre-training results on the {\color{blue} SVO-Probes} dataset. See Sec. \ref{svo_detailed_results} for more details.}
  \label{detailed_svo_pre_train}
\end{table}

\subsection{CREPE-Systematicity: detailed results}
\label{crepe_detailed_results}
Here we provide detailed results on CREPE-Systematicity dataset used for measuring systematic generalization. In the main paper we had only provided the results related to systematic generalization (i.e., the unseen split), but here we provide results on both the seen and unseen split, for both hard negative retrieval sets (Comp and Atom) that are used when evaluating performance on CREPE by \citet{ma2022crepe}.
Detailed Fine-tuning results on CREPE-Systematicity dataset on both the seen and unseen splits are provided in Table \ref{clip_fine-tune_crepe}. These are an extension to the CREPE-Systematicity results provided in the main paper Table \ref{clip_fine-tune_all}. Detailed Pre-training results for CREPE-Systematicity dataset are provided in Table \ref{pre-training_results_crepe} which are an extension to the CREPE-Systematicity results provided in the main paper Table \ref{pre-training_results_all}.

\begin{table*}[h!]
  \small
      \centering
      \begin{tabular}{lrrrr|rrrr}
        \toprule
        \multicolumn{1}{r}{(Pre-training, Fine-tuning) data $\rightarrow$} & \multicolumn{4}{c|}{CC-12M, CC-100K} & \multicolumn{4}{c}{YFCC-15M, YFCC-100K}\\
        \cmidrule(lr){2-5} \cmidrule(lr){6-9}
        \multicolumn{1}{r}{Retrieval Set $\rightarrow$} & \multicolumn{2}{c}{Comp} & \multicolumn{2}{c|}{Atom} & \multicolumn{2}{c}{Comp} & \multicolumn{2}{c}{Atom} \\
        \cmidrule(lr){2-3} \cmidrule(lr){4-5} \cmidrule(lr){6-7} \cmidrule(lr){8-9}
        \multicolumn{1}{r}{\begin{tabular}{@{}l@{\hspace{7em}}r@{}}Method & Split $\rightarrow$\end{tabular}} & Seen & Unseen & Seen & Unseen & Seen & Unseen & Seen & Unseen \\
  
        \midrule 
          \clip{}                         & 48.3 & 45.1 & 39.2 & 35.0 & 42.0 & 39.8 & 43.4 & 39.5\\[1pt]
          \clip{}-FT                      & 48.5 & 45.8 & 40.0 & 35.6 & 39.1 & 36.4 & 42.4 & 38.3\\[1pt]
          \negclip{}                      & 55.1 & 53.1 & 41.5 & 37.5 & 41.9 & 38.8 & 42.8 & 39.0\\[1pt]
          \midrule
          \rowcolor{cyan!12}
          \methodcompNoCurricbold{}     & 71.4 & 71.1 & 45.3 & 40.2 & 50.1 & 46.6 & 44.9 & 40.5\\[1pt]
          \rowcolor{cyan!12}
          \methodcompwiseftbold{}     &  68.4 & 67.5 & 46.1 & \textbf{41.2}  &  48.9 & 48.1 & \textbf{46.2} & \textbf{43.6} \\[1pt]
          \rowcolor{cyan!12}
          \methodcompbold{}             & \textbf{73.1} & \textbf{72.4} & \textbf{46.2} & \textbf{40.9} & \textbf{52.3} & \textbf{48.8} & \textbf{45.7} & \textbf{41.5}\\[1pt]
        \bottomrule
      \end{tabular}
    
      \caption{Fine-tuning results on the {\color{blue} CREPE - Systematicity} datasets. We take OpenCLIP models pre-trained on CC-12M, and YFCC-15M, fine-tune them on CC-100K, and YFCC-100K, respectively, and test them on CC-12M, YFCC-15M split of CREPE dataset, respectively. See Sec. \ref{results} for more details. We recalculate CLIP results since \citet{ma2022crepe} do not normalize \clip{} embeddings before taking the dot product for text and image embeddings, resulting in an incorrect score.}
      \label{clip_fine-tune_crepe}
  \end{table*}
  
\begin{table*}[h!]
  \small
  \centering
  \begin{tabular}{llrrrr|rrrr}
  \toprule
  & \multicolumn{1}{r}{Pre-training data $\rightarrow$} & \multicolumn{4}{c|}{CC-12M} & \multicolumn{4}{c}{YFCC-15M}\\
  \cmidrule(lr){3-6} \cmidrule(lr){7-10}
  & \multicolumn{1}{r}{Retrieval Set $\rightarrow$} & \multicolumn{2}{c}{Comp} & \multicolumn{2}{c|}{Atom} & \multicolumn{2}{c}{Comp} & \multicolumn{2}{c}{Atom}\\
  \cmidrule(lr){3-4} \cmidrule(lr){5-6} \cmidrule(lr){7-8} \cmidrule(lr){9-10}
  {Arch.} & \multicolumn{1}{r}{\begin{tabular}{@{}l@{\hspace{5em}}r@{}}Method $\downarrow$ & Split $\rightarrow$\end{tabular}} & Seen & Unseen & Seen & Unseen & Seen & Unseen & Seen & Unseen \\
    \midrule 
    & \clip{} & 45.9 &	44.1 &	41.7 &	37.3 &  40.2 &	39.6 &	42.9 &	41.7\\[1pt]
    & \negclip{} & 76.4 &	80.3 &	45.1 &	39.6 &  47.3 &	47.1 &	43.2 &	41.5 \\[1pt]
    \rowcolor{cyan!12}
    \cellcolor{white} \multirow{-3}{*}{\rotatebox[origin=c]{90}{Swin-T}} & \methodcompbold{}  & \textbf{85.3} & \textbf{92.1} & \textbf{49.3} &	\textbf{44.5} &  \textbf{80.7} & \textbf{89.6} & \textbf{48.2} & \textbf{45.3} \\[1pt]
    \midrule
      
    & \clip{} & 44.9	& 42.9	& 40.9	& 36.7 & 38.7	& 38.9	& 40.6	& 38.9 \\[1pt]
    & \negclip{} & 78.6 & 82.0 & 46.8 & 41.4 & 61.5	& 67.2	& 43.5	& 41.5 \\[1pt]
    \rowcolor{cyan!12}
    \cellcolor{white} \multirow{-3}{*}{\rotatebox[origin=c]{90}{RN-50}} & \methodcompbold{}  & \textbf{85.3} & \textbf{92.6} & \textbf{47.8} & \textbf{44.4} & \textbf{80.1} & \textbf{90.2} & \textbf{46.6} & \textbf{45.0} \\[1pt]
    \bottomrule
  \end{tabular}

  \caption{Pre-training results on {\color{blue} CREPE - Systematicity} datasets. Models are pre-trained using CC-12M and YFCC-15M datasets and tested on the corresponding CC-12M and YFCC-15M split of the CREPE dataset. Results for both backbones Swin-Tiny and RN-50 are shown. See Sec. \ref{results} for more details.}
  \label{pre-training_results_crepe}
  \end{table*}

\section{Reproducibility}
\label{reproducibility}
Here we provide necessary details to reproduce our work, that might not have been included in the main paper.
\subsection{Training and hyperparameter details}
\label{hyperparams}
\underline{Fine-tuning:} For all fine-tuning experiments, we follow \citet{yuksekgonul2022and} for hyperparameters. In particular, all models are fine-tuned for 5 epochs, with a batch size of 256, using a cosine learning rate schedule with 50 steps of warmup and random-crop augmentation during training. AdamW is used for optimization. $1e-5$ is used as the initial learning rate. Training is performed using 4 NVIDIA A100 GPUs for all models. From the ARO dataset, $10\%$ examples from attribute and relation splits are used as validation examples, and the rest are used as the test set for all models. On all other datasets, we evaluate zero-shot performance. For \methodcomp{}, we find that sampling a maximum of 3 positive and 6 negative sub-graphs per image during fine-tuning gives the best result on the ARO validation set and hence is used in all our experiments (including pre-training experiments). 
For \methodcomp{}, we keep sub-graph sampling probabilities as $p_2=p_3$. We vary $p_1$ in $\{0, 0.08, 0.15\}$ while fine-tuning on the randomly chosen YFCC dataset. We choose the best model according to the ARO val-set and keep the hyperparameters the same for all other fine-tuning datasets.

\noindent\underline{Pre-training:} For pre-training experiments, we follow the training protocol used in \citet{yang2022unified, radford2021learning}. In particular, all models are trained for 32 epochs, with a batch size of 4096, using a cosine learning rate schedule with 5000 steps of warmup and random-crop augmentation during training. AdamW is used for optimization. The initial learning rate is $1e-3$, and weight decay is set to $0.1$. Training is performed using 64 NVIDIA A100 GPUs.
\negclip{}'s hard negative text creation method often results in no negative text for some texts in the pre-training dataset. Removing all such image-text pairs with no possible hard negative text results in poor performance for \negclip{} (due to fewer data to pre-train on). If we include these image-text pairs, the text batch size might differ for different GPUs since some image-text pairs are without hard negative texts and this causes instabilities. We hence keep a cache of sentences from previous batches and add it to the batch as negative examples so that all GPUs have the same text batch size during training. The same is done for \methodcomp{} since not all images might have the same number of unique positive and negative sub-graphs available. For \negclip{} we create hard negative sentences using code released by \citep{yuksekgonul2022and}. For \methodcomp{} training, for each image, we always use one hard negative text createdusing \negclip{}'s swapping technique, followed by positive and negative subgraphs created using our method. Sub-graph sampling probabilities are kept as $p_2=p_3$, $p_1=0.15$.
\subsection{Tree-Score details:} \citet{murty2022characterizing} devised a method to calculate the tree-score of a transformer over a given dataset of sentences $\mathbb{D}$. This tree-score measures the functional tree-structuredness of a given transformer encoder. See \citet{murty2022characterizing} for exact details for the algorithm to calculate the tree-scores. We use the code released by the authors\footnote{https://github.com/MurtyShikhar/TreeProjections} for the purpose of calculating tree-scores for CLIP's language encoder. In practice we use 5K sentences from the COCO-validation set as the held ouot test set $\mathbb{D}$ over which we calculate the tree-scores.

\subsection{Computing Infrastructure and Run-Time:} We use NVIDIA A100 GPUs for all our experiments. Pre-training experiments took about 1.5 days per model while using 64 GPUs. Fine-tuning experiments on CC-FT, YFCC-FT and COCO took about 45 mins each and experiments on CC3M took 5 hours per model, while using 4 GPUs.

\subsection{Model Parameters:} We use standrad CLIP models and as part of all models, is a transformer language encoder having 12 layers, 8 attention heads and 512 as it's width. For vision encoders we use 1. ResNet-50 hvaing 23M trainable parameters and 2. Transformer vision encoders a) Swin-Tiny with patch-size 4 and window size 7 following \citep{yang2022unified} and b) ViT-B-32 which has patch size 32, 12 layers and 12 attention heads.

\subsection{Evaluation Metrics:} Strictly following the respective papers and released code\footnote{ARO: \url{https://github.com/mertyg/vision-language-models-are-bows}, SVO-Probes \url{https://github.com/deepmind/svo_probes}, VL-Checklist: \url{https://github.com/om-ai-lab/VL-CheckList}}, for ARO, VL-Checklist, SVO we use accuracy as the metric as defined by the respecitve papers. And for CREPE-Productivty, and CREPE-Systematicity \footnote{CREPE Code: \url{https://github.com/RAIVNLab/CREPE}} we use Recall@1 as our metric of evaluation.

\subsection{Summary Statistics of results:} We provide standard deviation results using 3 random seeds in Appendix Section \ref{std_dev_results} for Fine-tuning experiments on the CC-FT dataset. For all other datasets, including the expensive pre-training runs we use a single seed for our experiments.

\newpage
\begin{figure}[h!]
    \centering
    \subcaptionbox{Finetuning data: CC-100k}{\includegraphics[width=0.45\columnwidth]{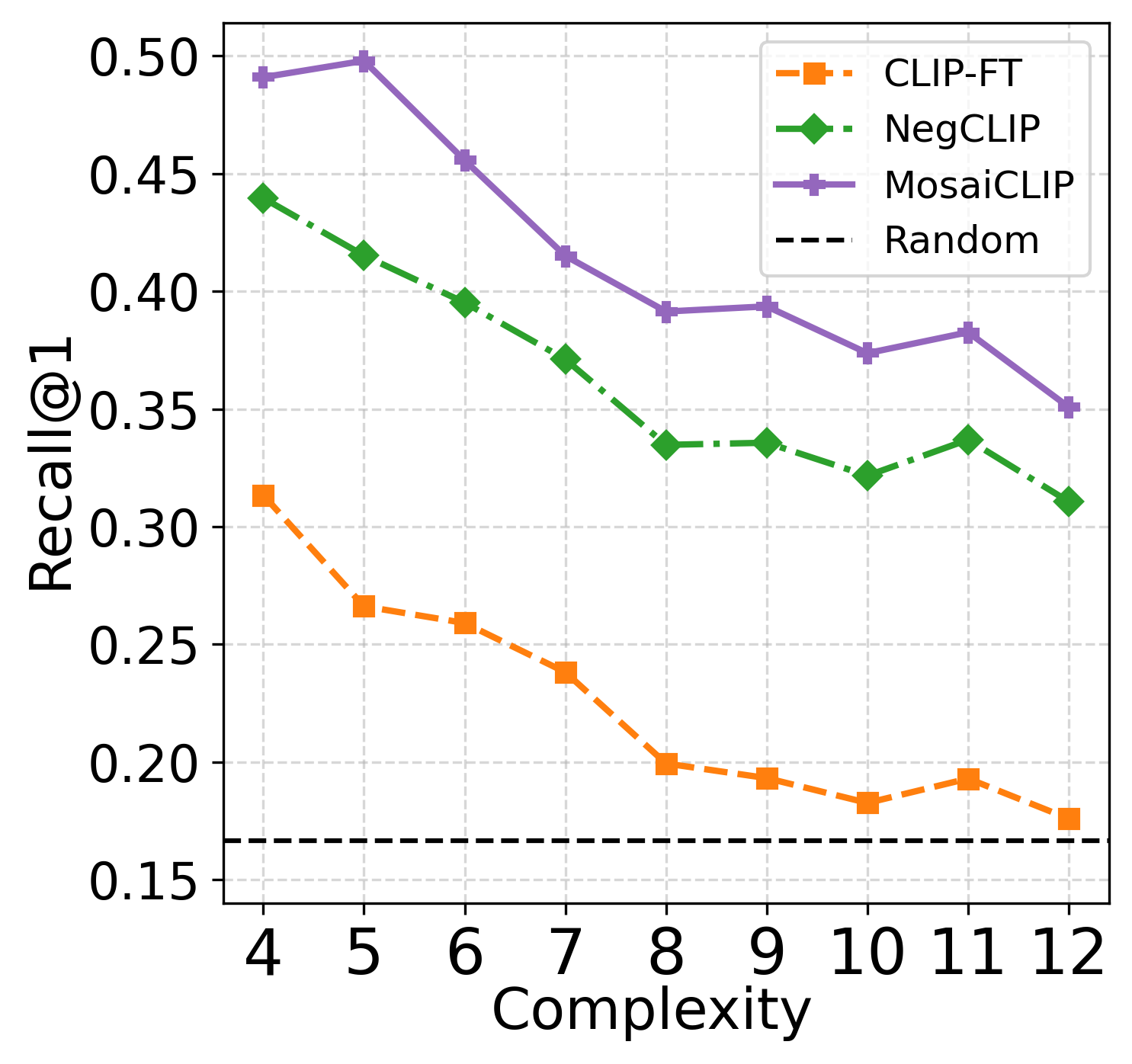}}
    \hfill
    \subcaptionbox{Finetuning data: COCO}{\includegraphics[width=0.45\columnwidth]{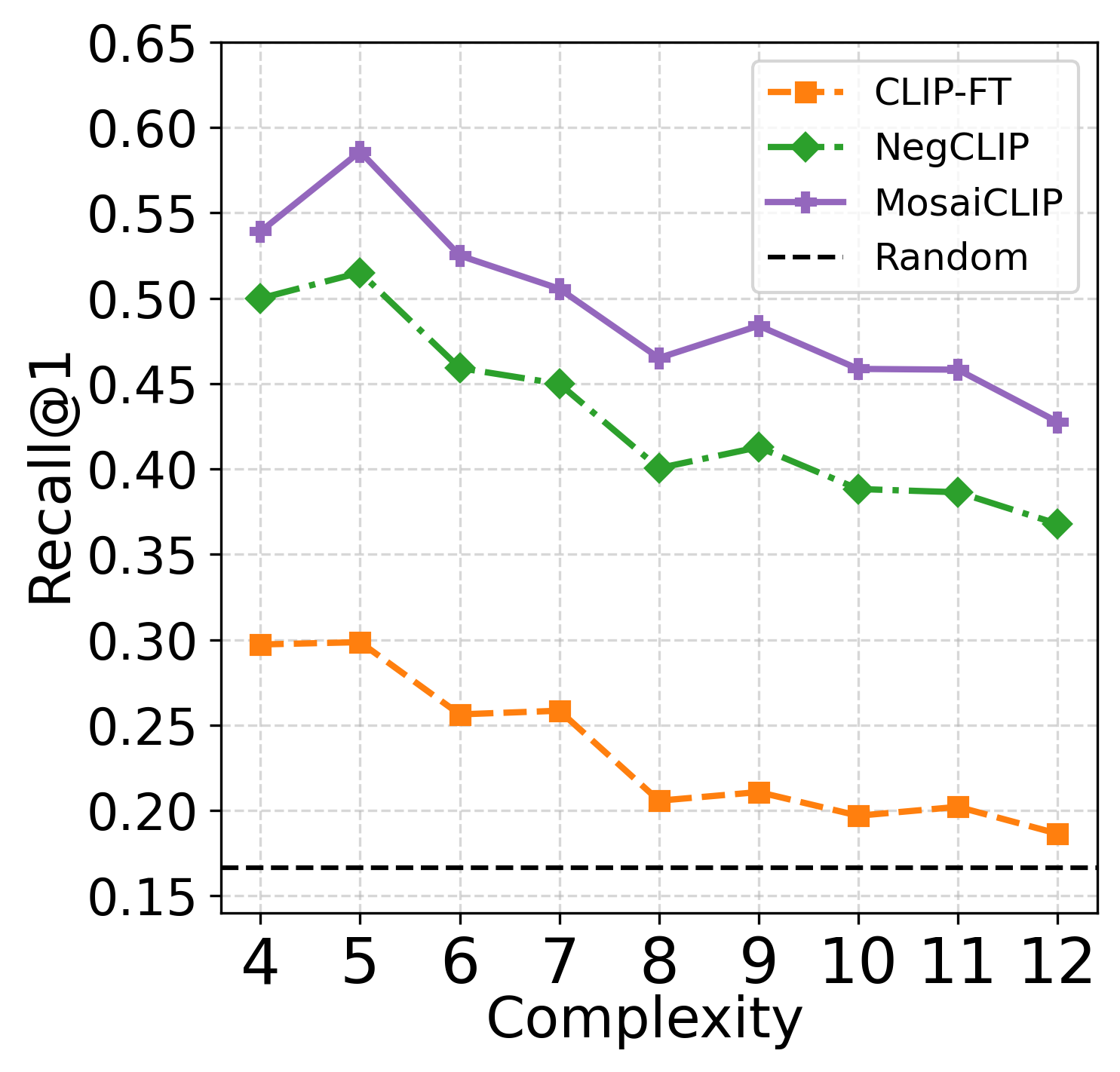}}
    \caption{Fine-tuninig Results on {\color{blue} CREPE - Productivity} (generalization to longer and more complex sentences). Fine-tuning datasets are mentioned below each figure.}
    \label{fig:Productivity_clip_ft_cc_coo}
\end{figure}

\begin{figure}[h!]
    \centering
    \subcaptionbox{Swin-Tiny, CC-12M}{\includegraphics[width=0.45\columnwidth]{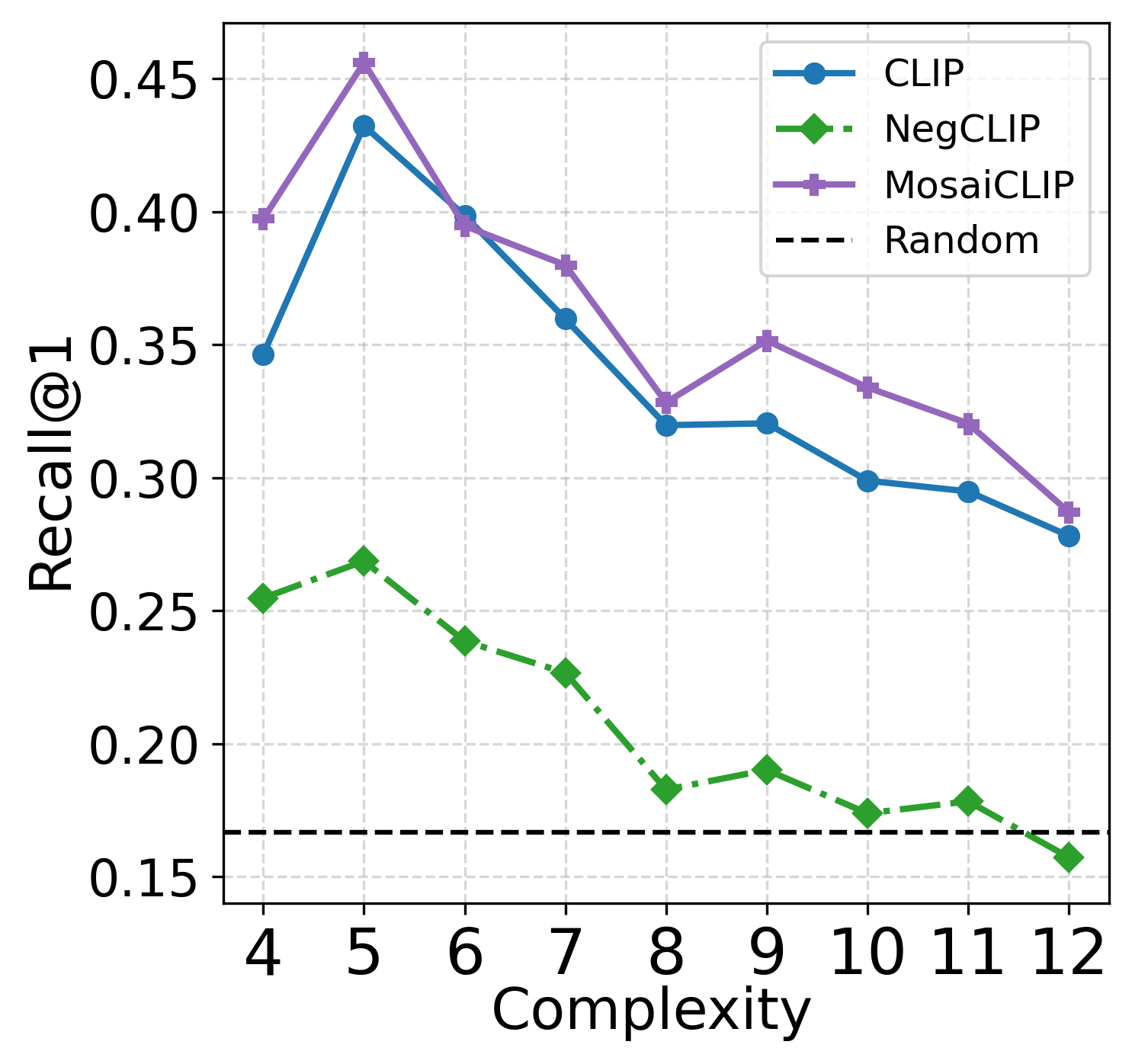}}
    \hfill
    \subcaptionbox{Swin-Tiny, YFCC-15M}{\includegraphics[width=0.45\columnwidth]{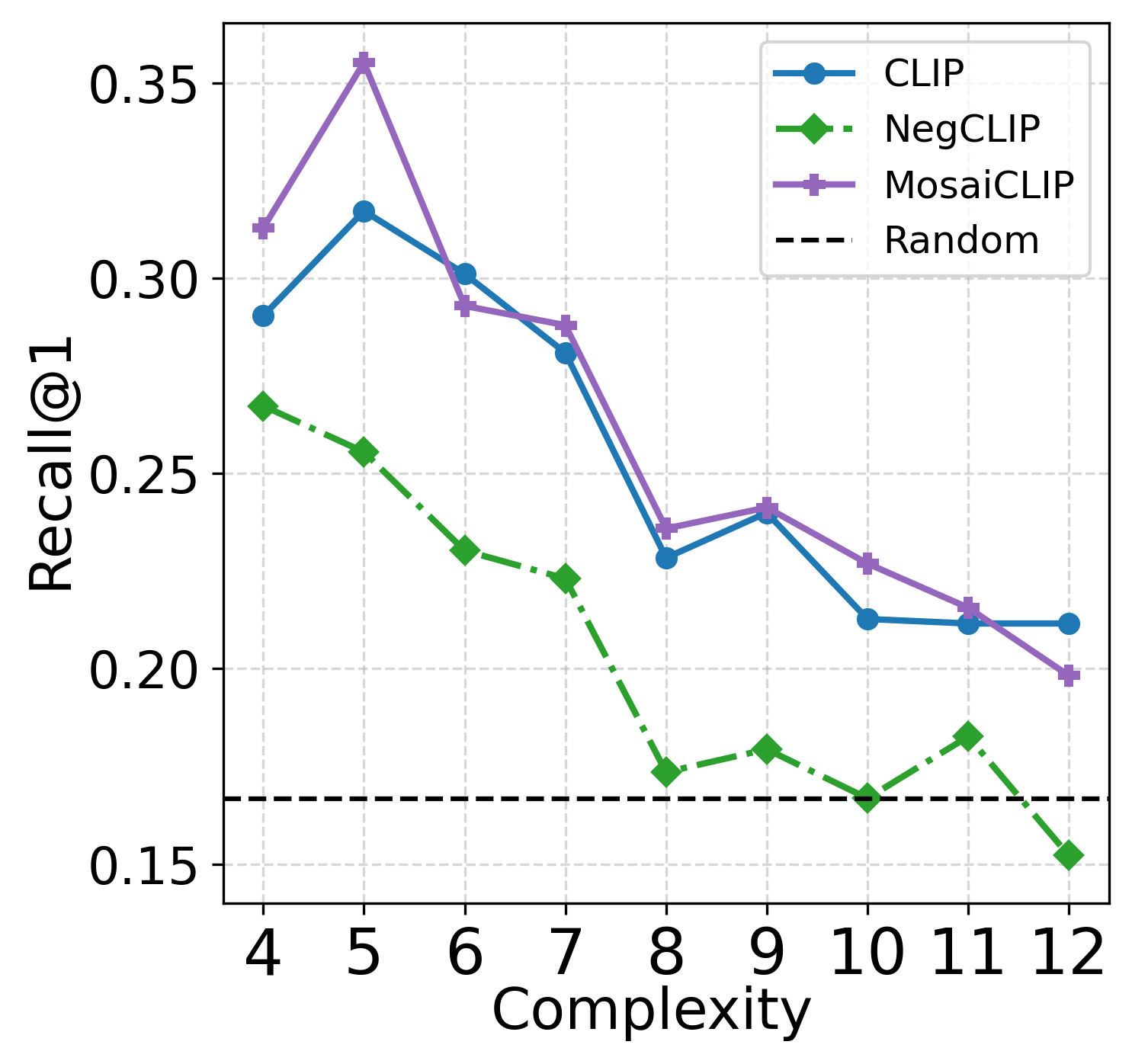}}
    \hfill
    \subcaptionbox{RN-50, CC-12M}{\includegraphics[width=0.45\columnwidth]{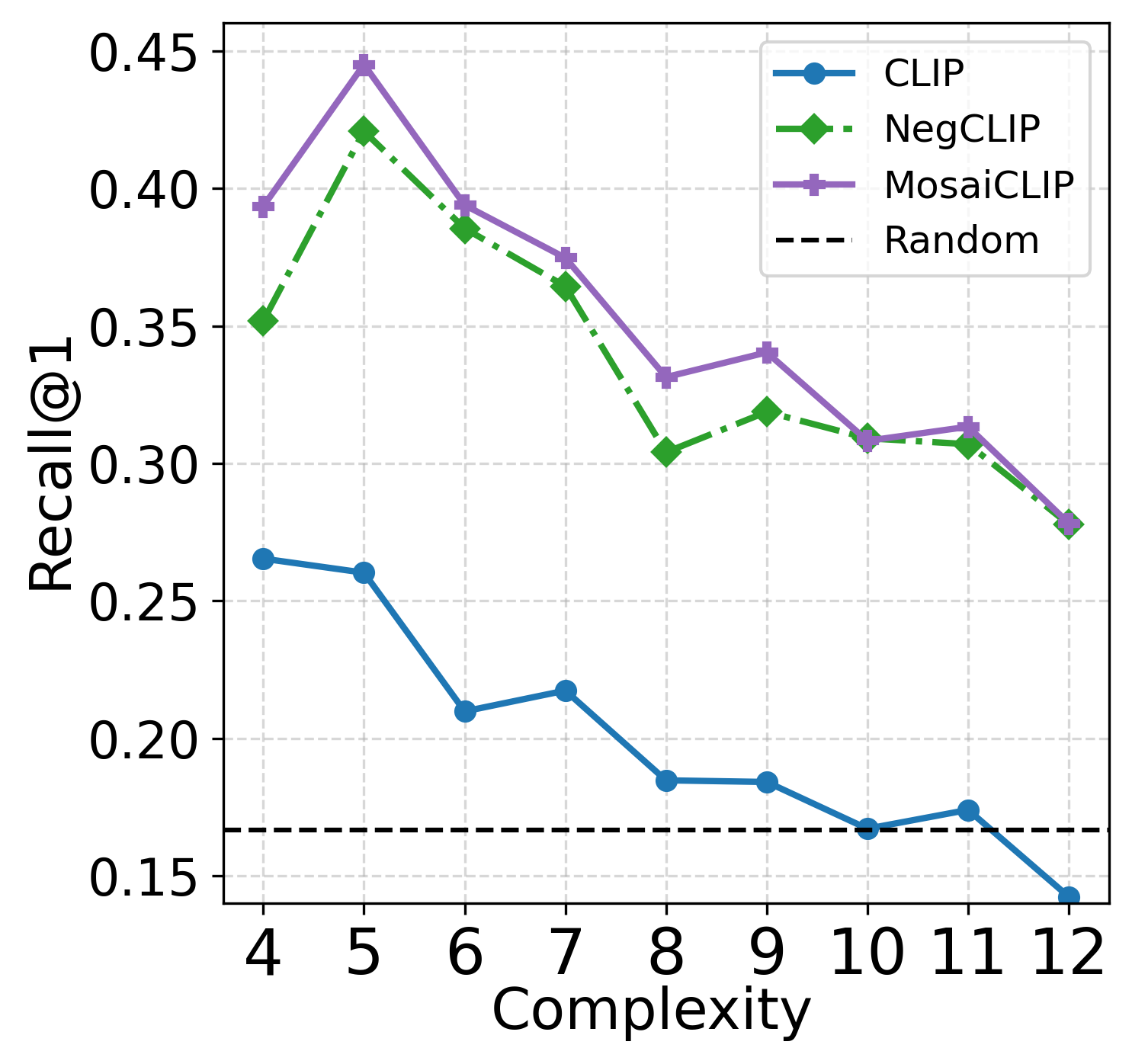}}
    \hfill
    \subcaptionbox{RN-50, YFCC-15M}{\includegraphics[width=0.45\columnwidth]{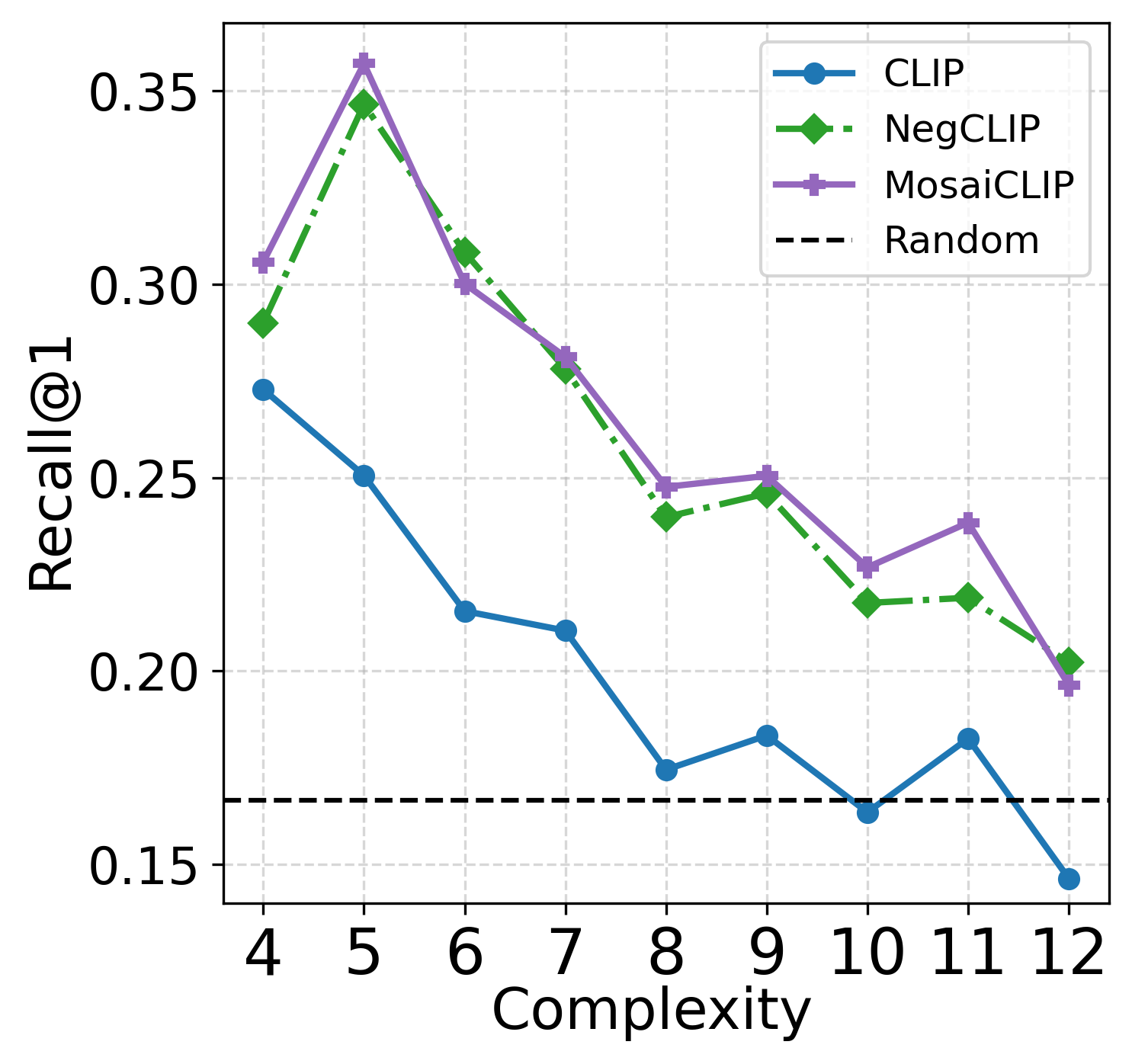}}
    \caption{Pre-Training Results on {\color{blue} CREPE - Productivity} (generalization to longer and more complex sentences). Pre-Training model and datasets are mentioned below each figure.}
    \label{fig:Productivity_clip_pretr}
\end{figure}

\begin{figure}[h!]
    \centering
    \centering
    \subcaptionbox{SVO}{\includegraphics[width=0.7\columnwidth]{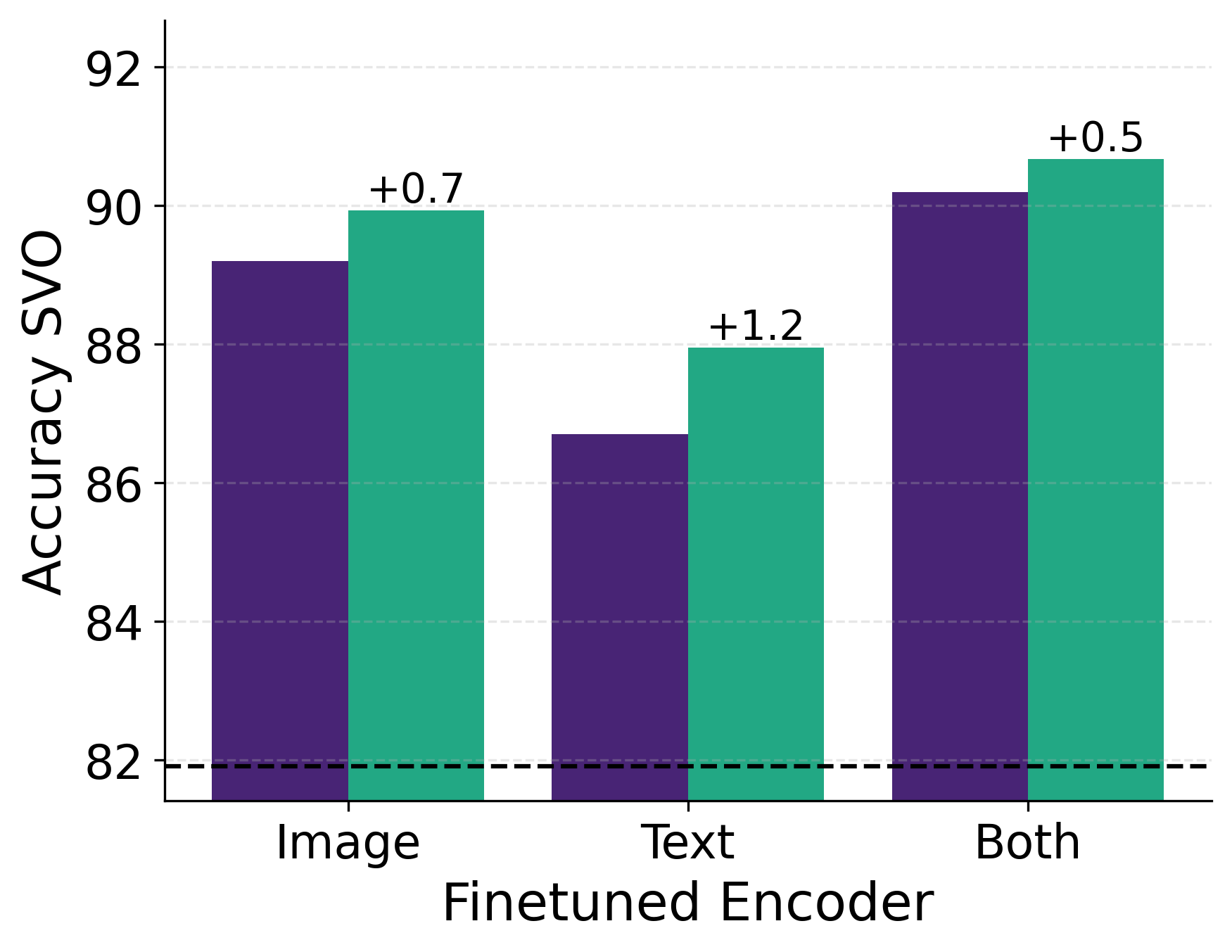}}
    \caption{Extension of Figure \ref{fig:freezing_expts_ourmethod_and_tree_score_clip_pretr} c), d). Selectively fine-tuning of image, text encoders and measure performance on {\color{blue} SVO-Probes} dataset.}
    \label{fig:freezing_expts__SVO}
\end{figure}

\begin{figure*}[h!]
    \centering
    \includegraphics[width=\linewidth]{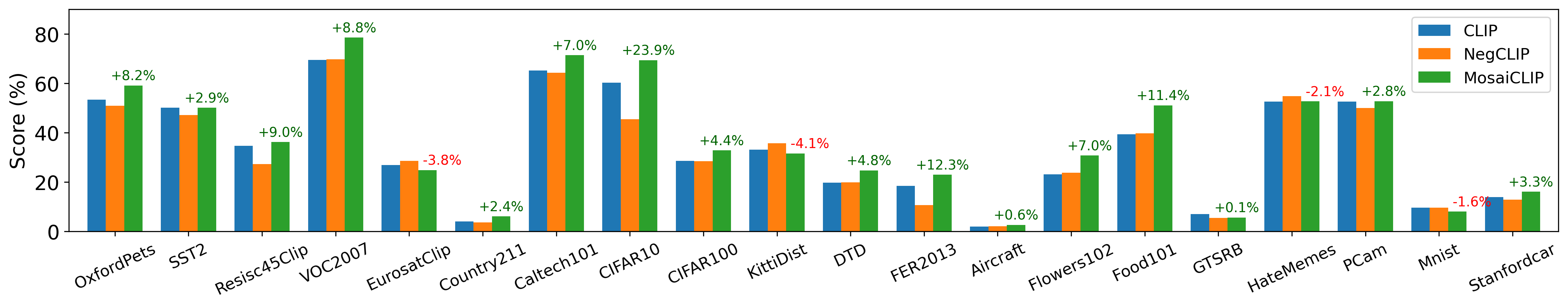}
    \caption{Comparing of \clip{}, \negclip{} and \methodcomp{} on 20 datasets of from the {\color{blue} ELEVATER} \citep{li2022elevater} benchmark. Models in this graph are pretrained with CC-12M data and have Swin-Tiny as the vision backbone. See Sec. \ref{results} for more details.}
    \label{fig:20_datasets_swin_CC}
\end{figure*}

\begin{figure*}[ht]
  \centering
  \includegraphics[width=\linewidth]{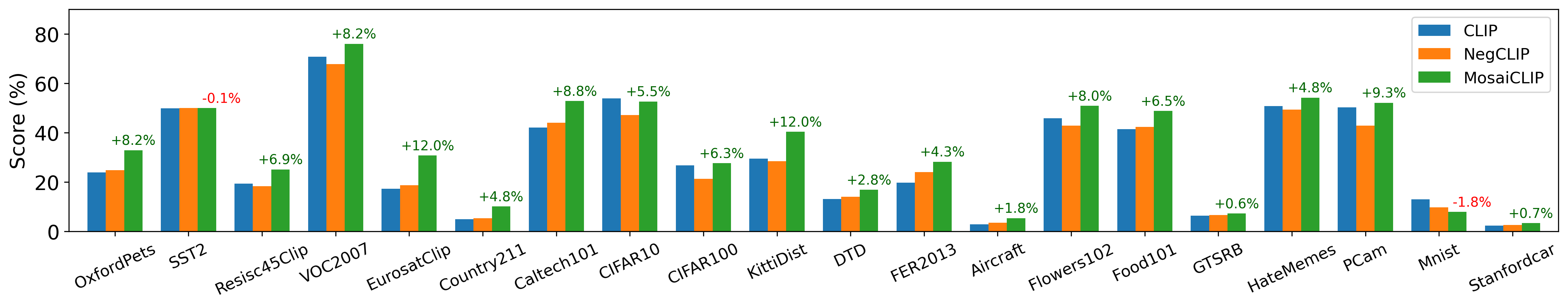}
  \caption{Comparing of \clip{}, \negclip{} and \methodcomp{} on 20 datasets of from the {\color{blue} ELEVATER} \citep{li2022elevater} benchmark. Models in this graph are pretrained with YFCC-15M data and have Swin-Tiny as the vision backbone. See Sec. \ref{results} for more details.}
  \label{fig:20_datasets_swin_YFCC}
\end{figure*}

\begin{figure*}[ht]
  \centering
  \includegraphics[width=\linewidth]{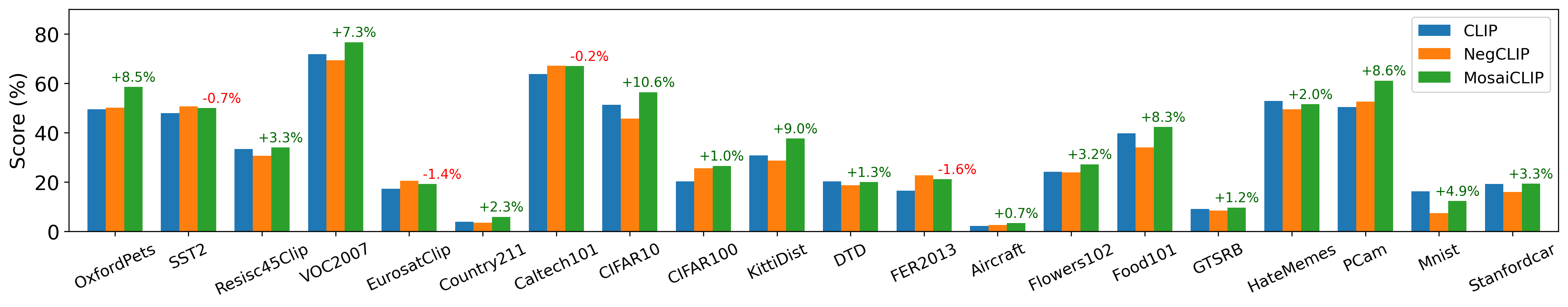}
  \caption{Comparing of \clip{}, \negclip{} and \methodcomp{} on 20 datasets of from the {\color{blue} ELEVATER} \citep{li2022elevater} benchmark. Models in this graph are pretrained with CC-12M data and have ResNet-50 as the vision backbone. See Sec. \ref{results} for more details.}
  \label{fig:20_datasets_rn50_CC}
\end{figure*}

\begin{figure*}[ht]
  \centering
  \includegraphics[width=\linewidth]{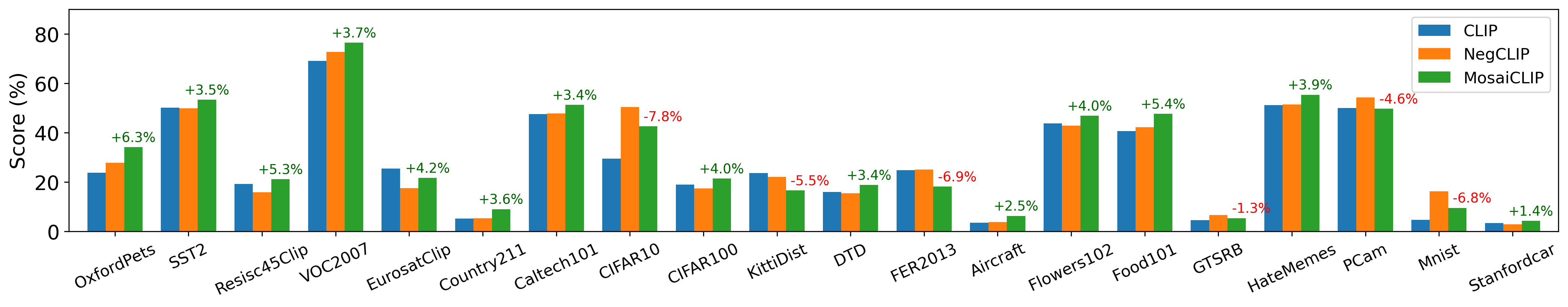}
  \caption{Comparing of \clip{}, \negclip{} and \methodcomp{} on 20 datasets of from the {\color{blue} ELEVATER} \citep{li2022elevater} benchmark. Models in this graph are pretrained with YFCC-15M data and have ResNet-50 as the vision backbone. See Sec. \ref{results} for more details.}
  \label{fig:20_datasets_rn50_YFCC}
\end{figure*}

%% file: main.bbl
\begin{thebibliography}{51}
\expandafter\ifx\csname natexlab\endcsname\relax\def\natexlab#1{#1}\fi

\bibitem[{Cascante-Bonilla et~al.(2023)Cascante-Bonilla, Shehada, Smith, Doveh,
  Kim, Panda, Varol, Oliva, Ordonez, Feris, and
  Karlinsky}]{cascantebonilla2023going}
Paola Cascante-Bonilla, Khaled Shehada, James~Seale Smith, Sivan Doveh,
  Donghyun Kim, Rameswar Panda, Gül Varol, Aude Oliva, Vicente Ordonez,
  Rogerio Feris, and Leonid Karlinsky. 2023.
\newblock \href {http://arxiv.org/abs/2303.17590} {Going beyond nouns with
  vision \& language models using synthetic data}.

\bibitem[{Changpinyo et~al.(2021)Changpinyo, Sharma, Ding, and
  Soricut}]{changpinyo2021cc12m}
Soravit Changpinyo, Piyush Sharma, Nan Ding, and Radu Soricut. 2021.
\newblock {Conceptual 12M}: Pushing web-scale image-text pre-training to
  recognize long-tail visual concepts.
\newblock In \emph{CVPR}.

\bibitem[{Crain and Nakayama(1987)}]{crain}
Stephen Crain and Mineharu Nakayama. 1987.
\newblock \href {http://www.jstor.org/stable/415004} {Structure dependence in
  grammar formation}.
\newblock \emph{Language}, 63(3):522--543.

\bibitem[{Deng et~al.(2009)Deng, Dong, Socher, Li, Li, and
  Fei-Fei}]{deng2009ImageNet}
Jia Deng, Wei Dong, Richard Socher, Li-Jia Li, Kai Li, and Li~Fei-Fei. 2009.
\newblock Imagenet: A large-scale hierarchical image database.
\newblock In \emph{2009 IEEE conference on computer vision and pattern
  recognition}, pages 248--255. Ieee.

\bibitem[{Diwan et~al.(2022)Diwan, Berry, Choi, Harwath, and
  Mahowald}]{diwan-etal-2022-winoground}
Anuj Diwan, Layne Berry, Eunsol Choi, David Harwath, and Kyle Mahowald. 2022.
\newblock \href {https://aclanthology.org/2022.emnlp-main.143} {Why is
  winoground hard? investigating failures in visuolinguistic compositionality}.
\newblock In \emph{Proceedings of the 2022 Conference on Empirical Methods in
  Natural Language Processing}, pages 2236--2250, Abu Dhabi, United Arab
  Emirates. Association for Computational Linguistics.

\bibitem[{Doveh et~al.(2023)Doveh, Arbelle, Harary, Schwartz, Herzig, Giryes,
  Feris, Panda, Ullman, and Karlinsky}]{doveh2023teaching}
Sivan Doveh, Assaf Arbelle, Sivan Harary, Eli Schwartz, Roei Herzig, Raja
  Giryes, Rogerio Feris, Rameswar Panda, Shimon Ullman, and Leonid Karlinsky.
  2023.
\newblock Teaching structured vision \& language concepts to vision \& language
  models.
\newblock In \emph{Proceedings of the IEEE/CVF Conference on Computer Vision
  and Pattern Recognition}, pages 2657--2668.

\bibitem[{Gan et~al.(2021)Gan, Schwartz, Alter, Mrowca, Schrimpf, Traer,
  Freitas, Kubilius, Bhandwaldar, Haber, Sano, Kim, Wang, Lingelbach, Curtis,
  Feigelis, Bear, Gutfreund, Cox, Torralba, DiCarlo, Tenenbaum, Mcdermott, and
  Yamins}]{gan2021threedworld}
Chuang Gan, Jeremy Schwartz, Seth Alter, Damian Mrowca, Martin Schrimpf, James
  Traer, Julian~De Freitas, Jonas Kubilius, Abhishek Bhandwaldar, Nick Haber,
  Megumi Sano, Kuno Kim, Elias Wang, Michael Lingelbach, Aidan Curtis,
  Kevin~Tyler Feigelis, Daniel Bear, Dan Gutfreund, David~Daniel Cox, Antonio
  Torralba, James~J. DiCarlo, Joshua~B. Tenenbaum, Josh Mcdermott, and
  Daniel~LK Yamins. 2021.
\newblock \href {https://openreview.net/forum?id=db1InWAwW2T} {Three{DW}orld: A
  platform for interactive multi-modal physical simulation}.
\newblock In \emph{Thirty-fifth Conference on Neural Information Processing
  Systems Datasets and Benchmarks Track (Round 1)}.

\bibitem[{Goyal et~al.(2022)Goyal, Kumar, Garg, Kolter, and
  Raghunathan}]{goyal2022finetune}
Sachin Goyal, Ananya Kumar, Sankalp Garg, Zico Kolter, and Aditi Raghunathan.
  2022.
\newblock \href {http://arxiv.org/abs/2212.00638} {Finetune like you pretrain:
  Improved finetuning of zero-shot vision models}.

\bibitem[{Hale et~al.(2018)Hale, Dyer, Kuncoro, and
  Brennan}]{hale-etal-2018-finding}
John Hale, Chris Dyer, Adhiguna Kuncoro, and Jonathan Brennan. 2018.
\newblock \href {https://doi.org/10.18653/v1/P18-1254} {Finding syntax in human
  encephalography with beam search}.
\newblock In \emph{Proceedings of the 56th Annual Meeting of the Association
  for Computational Linguistics (Volume 1: Long Papers)}, pages 2727--2736,
  Melbourne, Australia. Association for Computational Linguistics.

\bibitem[{Hendricks and Nematzadeh(2021)}]{hendricks-nematzadeh-2021-probing}
Lisa~Anne Hendricks and Aida Nematzadeh. 2021.
\newblock \href {https://doi.org/10.18653/v1/2021.findings-acl.318} {Probing
  image-language transformers for verb understanding}.
\newblock In \emph{Findings of the Association for Computational Linguistics:
  ACL-IJCNLP 2021}, pages 3635--3644, Online. Association for Computational
  Linguistics.

\bibitem[{Hendrycks et~al.(2021{\natexlab{a}})Hendrycks, Basart, Mu, Kadavath,
  Wang, Dorundo, Desai, Zhu, Parajuli, Guo, Song, Steinhardt, and
  Gilmer}]{hendrycks2021many}
Dan Hendrycks, Steven Basart, Norman Mu, Saurav Kadavath, Frank Wang, Evan
  Dorundo, Rahul Desai, Tyler Zhu, Samyak Parajuli, Mike Guo, Dawn Song, Jacob
  Steinhardt, and Justin Gilmer. 2021{\natexlab{a}}.
\newblock The many faces of robustness: A critical analysis of
  out-of-distribution generalization.
\newblock \emph{ICCV}.

\bibitem[{Hendrycks et~al.(2021{\natexlab{b}})Hendrycks, Zhao, Basart,
  Steinhardt, and Song}]{Hendrycks_2021_CVPR}
Dan Hendrycks, Kevin Zhao, Steven Basart, Jacob Steinhardt, and Dawn Song.
  2021{\natexlab{b}}.
\newblock Natural adversarial examples.
\newblock In \emph{Proceedings of the IEEE/CVF Conference on Computer Vision
  and Pattern Recognition (CVPR)}, pages 15262--15271.

\bibitem[{Ilharco et~al.(2021)Ilharco, Wortsman, Wightman, Gordon, Carlini,
  Taori, Dave, Shankar, Namkoong, Miller, Hajishirzi, Farhadi, and
  Schmidt}]{Ilharco_OpenCLIP_2021}
Gabriel Ilharco, Mitchell Wortsman, Ross Wightman, Cade Gordon, Nicholas
  Carlini, Rohan Taori, Achal Dave, Vaishaal Shankar, Hongseok Namkoong, John
  Miller, Hannaneh Hajishirzi, Ali Farhadi, and Ludwig Schmidt. 2021.
\newblock \href {https://doi.org/10.5281/zenodo.5143773} {{OpenCLIP}}.

\bibitem[{Jia et~al.(2021)Jia, Yang, Xia, Chen, Parekh, Pham, Le, Sung, Li, and
  Duerig}]{jia2021scaling}
Chao Jia, Yinfei Yang, Ye~Xia, Yi-Ting Chen, Zarana Parekh, Hieu Pham, Quoc Le,
  Yun-Hsuan Sung, Zhen Li, and Tom Duerig. 2021.
\newblock Scaling up visual and vision-language representation learning with
  noisy text supervision.
\newblock In \emph{International Conference on Machine Learning}, pages
  4904--4916. PMLR.

\bibitem[{Johnson et~al.(2018)Johnson, Gupta, and Fei-Fei}]{johnson2018image}
Justin Johnson, Agrim Gupta, and Li~Fei-Fei. 2018.
\newblock Image generation from scene graphs.
\newblock In \emph{Proceedings of the IEEE conference on computer vision and
  pattern recognition}, pages 1219--1228.

\bibitem[{Johnson et~al.(2015)Johnson, Krishna, Stark, Li, Shamma, Bernstein,
  and Fei-Fei}]{johnson2015image}
Justin Johnson, Ranjay Krishna, Michael Stark, Li-Jia Li, David Shamma, Michael
  Bernstein, and Li~Fei-Fei. 2015.
\newblock Image retrieval using scene graphs.
\newblock In \emph{Proceedings of the IEEE conference on computer vision and
  pattern recognition}, pages 3668--3678.

\bibitem[{Krishna et~al.(2016)Krishna, Zhu, Groth, Johnson, Hata, Kravitz,
  Chen, Kalantidis, Li, Shamma, Bernstein, and Fei-Fei}]{krishnavisualgenome}
Ranjay Krishna, Yuke Zhu, Oliver Groth, Justin Johnson, Kenji Hata, Joshua
  Kravitz, Stephanie Chen, Yannis Kalantidis, Li-Jia Li, David~A Shamma,
  Michael Bernstein, and Li~Fei-Fei. 2016.
\newblock \href {https://arxiv.org/abs/1602.07332} {Visual genome: Connecting
  language and vision using crowdsourced dense image annotations}.

\bibitem[{Kumar et~al.(2022)Kumar, Raghunathan, Jones, Ma, and
  Liang}]{kumar2022fine}
Ananya Kumar, Aditi Raghunathan, Robbie Jones, Tengyu Ma, and Percy Liang.
  2022.
\newblock Fine-tuning can distort pretrained features and underperform
  out-of-distribution.
\newblock \emph{arXiv preprint arXiv:2202.10054}.

\bibitem[{Li et~al.(2022{\natexlab{a}})Li, Liu, Li, Zhang, Aneja, Yang, Jin,
  Hu, Liu, Lee et~al.}]{li2022elevater}
Chunyuan Li, Haotian Liu, Liunian Li, Pengchuan Zhang, Jyoti Aneja, Jianwei
  Yang, Ping Jin, Houdong Hu, Zicheng Liu, Yong~Jae Lee, et~al.
  2022{\natexlab{a}}.
\newblock Elevater: A benchmark and toolkit for evaluating language-augmented
  visual models.
\newblock \emph{Advances in Neural Information Processing Systems},
  35:9287--9301.

\bibitem[{Li et~al.(2022{\natexlab{b}})Li, Li, Xiong, and Hoi}]{li2022blip}
Junnan Li, Dongxu Li, Caiming Xiong, and Steven Hoi. 2022{\natexlab{b}}.
\newblock Blip: Bootstrapping language-image pre-training for unified
  vision-language understanding and generation.
\newblock In \emph{International Conference on Machine Learning}, pages
  12888--12900. PMLR.

\bibitem[{Li et~al.(2022{\natexlab{c}})Li, Zhang, Zhang, Yang, Li, Zhong, Wang,
  Yuan, Zhang, Hwang et~al.}]{li2022grounded}
Liunian~Harold Li, Pengchuan Zhang, Haotian Zhang, Jianwei Yang, Chunyuan Li,
  Yiwu Zhong, Lijuan Wang, Lu~Yuan, Lei Zhang, Jenq-Neng Hwang, et~al.
  2022{\natexlab{c}}.
\newblock Grounded language-image pre-training.
\newblock In \emph{Proceedings of the IEEE/CVF Conference on Computer Vision
  and Pattern Recognition}, pages 10965--10975.

\bibitem[{Li et~al.(2017)Li, Ouyang, Zhou, Wang, and Wang}]{li2017scene}
Yikang Li, Wanli Ouyang, Bolei Zhou, Kun Wang, and Xiaogang Wang. 2017.
\newblock Scene graph generation from objects, phrases and region captions.
\newblock In \emph{Proceedings of the IEEE international conference on computer
  vision}, pages 1261--1270.

\bibitem[{Li et~al.(2019)Li, Xu, Liu, Huang, Xu, Chen, Ma, Wang, Fang, and
  Lu}]{hake}
Yong-Lu Li, Liang Xu, Xinpeng Liu, Xijie Huang, Yue Xu, Mingyang Chen, Ze~Ma,
  Shiyi Wang, Hao-Shu Fang, and Cewu Lu. 2019.
\newblock Hake: Human activity knowledge engine.
\newblock \emph{arXiv preprint arXiv:1904.06539}.

\bibitem[{Lin et~al.(2014)Lin, Maire, Belongie, Hays, Perona, Ramanan,
  Doll{\'a}r, and Zitnick}]{lin2014microsoft}
Tsung-Yi Lin, Michael Maire, Serge Belongie, James Hays, Pietro Perona, Deva
  Ramanan, Piotr Doll{\'a}r, and C~Lawrence Zitnick. 2014.
\newblock Microsoft coco: Common objects in context.
\newblock In \emph{Computer Vision--ECCV 2014: 13th European Conference,
  Zurich, Switzerland, September 6-12, 2014, Proceedings, Part V 13}, pages
  740--755. Springer.

\bibitem[{Ma et~al.(2022)Ma, Hong, Gul, Gandhi, Gao, and Krishna}]{ma2022crepe}
Zixian Ma, Jerry Hong, Mustafa~Omer Gul, Mona Gandhi, Irena Gao, and Ranjay
  Krishna. 2022.
\newblock Crepe: Can vision-language foundation models reason compositionally?
\newblock \emph{arXiv preprint arXiv:2212.07796}.

\bibitem[{Mitchell et~al.(May 2021-May 2022)Mitchell, Pistilli, Jernite,
  Ozoani, Gerchick, Rajani, Luccioni, Solaiman, Masoud, Nikpoor,
  Ferrandis~Carlos, Bekman, Akiki, Contractor, Lansky, McMillan-Major, Thrush,
  Ilić, Dupont, Longpre, Dey, Biderman, Kiela, Baylor, Le~Scao, Gokaslan,
  Launay, and Muennighoff}]{bloom}
Margaret Mitchell, Giada Pistilli, Yacine Jernite, Ezinwanne Ozoani, Marissa
  Gerchick, Nazneen Rajani, Sasha Luccioni, Irene Solaiman, Maraim Masoud,
  Somaieh Nikpoor, Muñoz Ferrandis~Carlos, Stas Bekman, Christopher Akiki,
  Danish Contractor, David Lansky, Angelina McMillan-Major, Tristan Thrush,
  Suzana Ilić, Gérard Dupont, Shayne Longpre, Manan Dey, Stella Biderman,
  Douwe Kiela, Emi Baylor, Teven Le~Scao, Aaron Gokaslan, Julien Launay, and
  Niklas Muennighoff. May 2021-May 2022.
\newblock Bigscience, bigscience language open-science open-access multilingual
  (bloom) language model.
\newblock \emph{International}.

\bibitem[{Mokady et~al.(2021)Mokady, Hertz, and Bermano}]{mokady2021clipcap}
Ron Mokady, Amir Hertz, and Amit~H Bermano. 2021.
\newblock Clipcap: Clip prefix for image captioning.
\newblock \emph{arXiv preprint arXiv:2111.09734}.

\bibitem[{Murty et~al.(2022)Murty, Sharma, Andreas, and
  Manning}]{murty2022characterizing}
Shikhar Murty, Pratyusha Sharma, Jacob Andreas, and Christopher~D Manning.
  2022.
\newblock Characterizing intrinsic compositionality in transformers with tree
  projections.
\newblock \emph{arXiv preprint arXiv:2211.01288}.

\bibitem[{OpenAI(2023)}]{openai2023gpt4}
OpenAI. 2023.
\newblock \href {http://arxiv.org/abs/2303.08774} {Gpt-4 technical report}.

\bibitem[{Pallier et~al.(2011)Pallier, Devauchelle, and
  Dehaene}]{pallier2011cortical}
Christophe Pallier, Anne-Dominique Devauchelle, and Stanislas Dehaene. 2011.
\newblock Cortical representation of the constituent structure of sentences.
\newblock \emph{Proceedings of the National Academy of Sciences},
  108(6):2522--2527.

\bibitem[{Pham et~al.(2021)Pham, Kafle, Lin, Ding, Cohen, Tran, and
  Shrivastava}]{vaw}
Khoi Pham, Kushal Kafle, Zhe Lin, Zhihong Ding, Scott Cohen, Quan Tran, and
  Abhinav Shrivastava. 2021.
\newblock Learning to predict visual attributes in the wild.
\newblock In \emph{Proceedings of the IEEE/CVF Conference on Computer Vision
  and Pattern Recognition (CVPR)}, pages 13018--13028.

\bibitem[{Pratt et~al.(2020)Pratt, Yatskar, Weihs, Farhadi, and
  Kembhavi}]{swig}
Sarah Pratt, Mark Yatskar, Luca Weihs, Ali Farhadi, and Aniruddha Kembhavi.
  2020.
\newblock Grounded situation recognition.
\newblock In \emph{Computer Vision--ECCV 2020: 16th European Conference,
  Glasgow, UK, August 23--28, 2020, Proceedings, Part IV 16}, pages 314--332.
  Springer.

\bibitem[{Radford et~al.(2021)Radford, Kim, Hallacy, Ramesh, Goh, Agarwal,
  Sastry, Askell, Mishkin, Clark et~al.}]{radford2021learning}
Alec Radford, Jong~Wook Kim, Chris Hallacy, Aditya Ramesh, Gabriel Goh,
  Sandhini Agarwal, Girish Sastry, Amanda Askell, Pamela Mishkin, Jack Clark,
  et~al. 2021.
\newblock Learning transferable visual models from natural language
  supervision.
\newblock In \emph{International Conference on Machine Learning}, pages
  8748--8763. PMLR.

\bibitem[{Recht et~al.(2019)Recht, Roelofs, Schmidt, and
  Shankar}]{recht2019imagenet}
Benjamin Recht, Rebecca Roelofs, Ludwig Schmidt, and Vaishaal Shankar. 2019.
\newblock Do imagenet classifiers generalize to imagenet?
\newblock In \emph{International conference on machine learning}, pages
  5389--5400. PMLR.

\bibitem[{Sharma et~al.(2018)Sharma, Ding, Goodman, and
  Soricut}]{sharma-etal-2018-conceptual}
Piyush Sharma, Nan Ding, Sebastian Goodman, and Radu Soricut. 2018.
\newblock \href {https://doi.org/10.18653/v1/P18-1238} {Conceptual captions: A
  cleaned, hypernymed, image alt-text dataset for automatic image captioning}.
\newblock In \emph{Proceedings of the 56th Annual Meeting of the Association
  for Computational Linguistics (Volume 1: Long Papers)}, pages 2556--2565,
  Melbourne, Australia. Association for Computational Linguistics.

\bibitem[{Tan and Bansal(2019)}]{tan-bansal-2019-lxmert}
Hao Tan and Mohit Bansal. 2019.
\newblock \href {https://doi.org/10.18653/v1/D19-1514} {{LXMERT}: Learning
  cross-modality encoder representations from transformers}.
\newblock In \emph{Proceedings of the 2019 Conference on Empirical Methods in
  Natural Language Processing and the 9th International Joint Conference on
  Natural Language Processing (EMNLP-IJCNLP)}, pages 5100--5111, Hong Kong,
  China. Association for Computational Linguistics.

\bibitem[{Thomee et~al.(2016)Thomee, Shamma, Friedland, Elizalde, Ni, Poland,
  Borth, and Li}]{thomee2016yfcc100m}
Bart Thomee, David~A Shamma, Gerald Friedland, Benjamin Elizalde, Karl Ni,
  Douglas Poland, Damian Borth, and Li-Jia Li. 2016.
\newblock Yfcc100m: The new data in multimedia research.
\newblock \emph{Communications of the ACM}, 59(2):64--73.

\bibitem[{Thrush et~al.(2022)Thrush, Jiang, Bartolo, Singh, Williams, Kiela,
  and Ross}]{thrush2022winoground}
Tristan Thrush, Ryan Jiang, Max Bartolo, Amanpreet Singh, Adina Williams, Douwe
  Kiela, and Candace Ross. 2022.
\newblock Winoground: Probing vision and language models for visio-linguistic
  compositionality.
\newblock In \emph{Proceedings of the IEEE/CVF Conference on Computer Vision
  and Pattern Recognition}, pages 5238--5248.

\bibitem[{Wang et~al.(2019)Wang, Ge, Lipton, and Xing}]{wang2019learning}
Haohan Wang, Songwei Ge, Zachary Lipton, and Eric~P Xing. 2019.
\newblock Learning robust global representations by penalizing local predictive
  power.
\newblock \emph{Advances in Neural Information Processing Systems}, 32.

\bibitem[{Wortsman et~al.(2022)Wortsman, Ilharco, Kim, Li, Kornblith, Roelofs,
  Lopes, Hajishirzi, Farhadi, Namkoong et~al.}]{wortsman2022robust}
Mitchell Wortsman, Gabriel Ilharco, Jong~Wook Kim, Mike Li, Simon Kornblith,
  Rebecca Roelofs, Raphael~Gontijo Lopes, Hannaneh Hajishirzi, Ali Farhadi,
  Hongseok Namkoong, et~al. 2022.
\newblock Robust fine-tuning of zero-shot models.
\newblock In \emph{Proceedings of the IEEE/CVF Conference on Computer Vision
  and Pattern Recognition}, pages 7959--7971.

\bibitem[{Wu et~al.(2019)Wu, Mao, Zhang, Jiang, Li, Sun, and
  Ma}]{wu2019unified}
Hao Wu, Jiayuan Mao, Yufeng Zhang, Yuning Jiang, Lei Li, Weiwei Sun, and
  Wei-Ying Ma. 2019.
\newblock Unified visual-semantic embeddings: Bridging vision and language with
  structured meaning representations.
\newblock In \emph{Proceedings of the IEEE/CVF Conference on Computer Vision
  and Pattern Recognition}, pages 6609--6618.

\bibitem[{Xu et~al.(2017)Xu, Zhu, Choy, and Fei-Fei}]{xu2017scene}
Danfei Xu, Yuke Zhu, Christopher~B Choy, and Li~Fei-Fei. 2017.
\newblock Scene graph generation by iterative message passing.
\newblock In \emph{Proceedings of the IEEE conference on computer vision and
  pattern recognition}, pages 5410--5419.

\bibitem[{Yang et~al.(2022)Yang, Li, Zhang, Xiao, Liu, Yuan, and
  Gao}]{yang2022unified}
Jianwei Yang, Chunyuan Li, Pengchuan Zhang, Bin Xiao, Ce~Liu, Lu~Yuan, and
  Jianfeng Gao. 2022.
\newblock Unified contrastive learning in image-text-label space.
\newblock In \emph{Proceedings of the IEEE/CVF Conference on Computer Vision
  and Pattern Recognition}, pages 19163--19173.

\bibitem[{Yang et~al.(2019)Yang, Tang, Zhang, and Cai}]{yang2019auto}
Xu~Yang, Kaihua Tang, Hanwang Zhang, and Jianfei Cai. 2019.
\newblock Auto-encoding scene graphs for image captioning.
\newblock In \emph{Proceedings of the IEEE/CVF conference on computer vision
  and pattern recognition}, pages 10685--10694.

\bibitem[{Young et~al.(2014)Young, Lai, Hodosh, and Hockenmaier}]{flickr}
Peter Young, Alice Lai, Micah Hodosh, and Julia Hockenmaier. 2014.
\newblock \href {https://doi.org/10.1162/tacl_a_00166} {{From image
  descriptions to visual denotations: New similarity metrics for semantic
  inference over event descriptions}}.
\newblock \emph{Transactions of the Association for Computational Linguistics},
  2:67--78.

\bibitem[{Yuksekgonul et~al.(2022)Yuksekgonul, Bianchi, Kalluri, Jurafsky, and
  Zou}]{yuksekgonul2022and}
Mert Yuksekgonul, Federico Bianchi, Pratyusha Kalluri, Dan Jurafsky, and James
  Zou. 2022.
\newblock When and why vision-language models behave like bag-of-words models,
  and what to do about it?
\newblock \emph{arXiv preprint arXiv:2210.01936}.

\bibitem[{Zeng et~al.(2021)Zeng, Zhang, and Li}]{zeng2021multi}
Yan Zeng, Xinsong Zhang, and Hang Li. 2021.
\newblock Multi-grained vision language pre-training: Aligning texts with
  visual concepts.
\newblock \emph{arXiv preprint arXiv:2111.08276}.

\bibitem[{Zhang et~al.(2019)Zhang, Shih, Elgammal, Tao, and
  Catanzaro}]{zhang2019graphical}
Ji~Zhang, Kevin~J Shih, Ahmed Elgammal, Andrew Tao, and Bryan Catanzaro. 2019.
\newblock Graphical contrastive losses for scene graph parsing.
\newblock In \emph{Proceedings of the IEEE/CVF Conference on Computer Vision
  and Pattern Recognition}, pages 11535--11543.

\bibitem[{Zhang et~al.(2021)Zhang, Li, Hu, Yang, Zhang, Wang, Choi, and
  Gao}]{Zhang_2021_CVPR}
Pengchuan Zhang, Xiujun Li, Xiaowei Hu, Jianwei Yang, Lei Zhang, Lijuan Wang,
  Yejin Choi, and Jianfeng Gao. 2021.
\newblock Vinvl: Revisiting visual representations in vision-language models.
\newblock In \emph{Proceedings of the IEEE/CVF Conference on Computer Vision
  and Pattern Recognition (CVPR)}, pages 5579--5588.

\bibitem[{Zhao et~al.(2022)Zhao, Zhang, Zhu, Shen, Lee, Lu, and
  Yin}]{zhao2022vlchecklist}
Tiancheng Zhao, Tianqi Zhang, Mingwei Zhu, Haozhan Shen, Kyusong Lee, Xiaopeng
  Lu, and Jianwei Yin. 2022.
\newblock \href {http://arxiv.org/abs/2207.00221} {Vl-checklist: Evaluating
  pre-trained vision-language models with objects, attributes and relations}.

\bibitem[{Zhong et~al.(2022)Zhong, Yang, Zhang, Li, Codella, Li, Zhou, Dai,
  Yuan, Li et~al.}]{zhong2022regionclip}
Yiwu Zhong, Jianwei Yang, Pengchuan Zhang, Chunyuan Li, Noel Codella,
  Liunian~Harold Li, Luowei Zhou, Xiyang Dai, Lu~Yuan, Yin Li, et~al. 2022.
\newblock Regionclip: Region-based language-image pretraining.
\newblock In \emph{Proceedings of the IEEE/CVF Conference on Computer Vision
  and Pattern Recognition}, pages 16793--16803.

\end{thebibliography}
